\documentclass[twocolumn,10pt]{article}
\usepackage[top=1in, bottom=1in, left=.65in, right=.65in]{geometry}
\usepackage[utf8]{inputenc}
\usepackage{setspace}
\usepackage{etoolbox}
\usepackage{xcolor}
\usepackage{graphicx}
\usepackage{cprotect}
\usepackage{xspace}
\usepackage[affil-it]{authblk} 
\usepackage[symbol]{footmisc}
\graphicspath{ {./figures/} }
\usepackage{palatino} % needs 10
\usepackage[numbers,sort&compress]{natbib}

\usepackage[title]{appendix}

\title{\huge{\textbf{SINDy-RL: Interpretable and Efficient\\ Model-Based Reinforcement Learning}}}

\author[1,5]{Nicholas Zolman\thanks{Corresponding author (nzolman@uw.edu)}}
\author[1]{Christian Lagemann}
\author[2]{Urban Fasel}
\author[3,4]{J. Nathan Kutz}
\author[1]{Steven L. Brunton}
{\small
\affil[1]{\small Department of Mechanical Engineering, University of Washington, Seattle, WA 98195, USA}
\affil[2]{\small Department of Aeronautics, Imperial College, London SW7 2AZ, United Kingdom}
\affil[3]{\small Department of Applied Mathematics, University of Washington, Seattle, WA 98195}
\affil[4]{\small Department of Electrical and Computer Engineering, University of Washington, Seattle, WA 98195}
\affil[5]{\small Data Science and Artificial Intelligence Department, The Aerospace Corporation, El Segundo, CA 90245}
}
\date{}

\usepackage{url}

\usepackage{microtype}
\usepackage{graphicx}
\usepackage{booktabs} %
\usepackage{hyperref}

\usepackage{caption}
\usepackage{natbib}

\usepackage{amsmath,amssymb,amsfonts,amsthm}

\usepackage{mathrsfs}
\usepackage{xcolor}
\usepackage{mathtools}
\usepackage{dsfont}
\usepackage{hyperref}
\usepackage{bm}
\usepackage{nicefrac}
\usepackage{wrapfig}
\usepackage{lipsum}
\usepackage{enumerate}
\usepackage{outlines}
\usepackage{amsmath,algorithm,float,tabularx,amsfonts}

\usepackage[utf8]{inputenc}
\usepackage{algorithm,algcompatible,amssymb,amsmath}
\renewcommand{\COMMENT}[2][.65\linewidth]{%
  \leavevmode\hfill\makebox[#1][l]{$\triangleright$\ ~#2}}

\algnewcommand\algorithmicto{\textbf{to}}
\algnewcommand\RETURN{\State \textbf{return} }

\usepackage{algpseudocode}
\makeatletter
\newcommand{\multiline}[1]{%
    \begin{tabularx}{\dimexpr\linewidth-\ALG@thistlm}[t]{@{}X@{}}
        #1
    \end{tabularx}
}
\newcommand{\Input}[1]{\algrenewcommand{\alglinenumber}[1]{Input: \ \setcounter{ALG@line}{\numexpr##1-1}} #1}
\newcommand{\Step}[1]{\algrenewcommand{\alglinenumber}[1]{Step ##1: } #1}
\newcommand{\NoNumber}{\algrenewcommand{\alglinenumber}[1]{\setcounter{ALG@line}{\numexpr##1-1} \ \ \ \ \ \ \ \ \ \ }}
\newcommand{\Output}[1]{\algrenewcommand{\alglinenumber}[1]{Output:\setcounter{ALG@line}{\numexpr##1-1}} #1}

\newcounter{assumption}%
\renewcommand{\theassumption}{\arabic{assumption}}

\def\vx{{\bf{x}}}
\def\vX{{\bf X}}
\def\vy{{\bf y}}
\def\vY{{\bf Y}}

\def\vu{{\bf u}}

\def\vU{{\bf U}}

\def\vPSI{{\bf \Psi}}

\newcommand*\E[1]{\mathbb{E}\left[#1\right]}

\def\vTheta{{\mathbf{\Theta}}}

\def\vXi{{\mathbf{\Xi}}}
\def\vxi{{\mathbf{\xi}}}

\numberwithin{lemma}{section} % important bit
\numberwithin{theorem}{section} % important bit
\numberwithin{corollary}{section} % important bit
\numberwithin{proposition}{section} % important bit
\numberwithin{definition}{section} % important bit
\numberwithin{example}{section} % important bit
\numberwithin{question}{section} % important bit

\DeclareMathOperator*{\argmax}{argmax}
\DeclareMathOperator*{\argmin}{argmin}

\usepackage{multirow}
\usepackage{caption}
\usepackage{subcaption} 
\usepackage[font={scriptsize},labelfont={bf}]{caption}

\usepackage{xspace}
\usepackage{cprotect}
\newcommand{\env}{\mathcal{E}}
\newcommand{\RLstate}{\vx}

\newcommand{\RLact}{\vu}
\newcommand{\RLpolicy}{\pi}
\newcommand{\RLrew}{r}

\newcommand{\policyParams}{\phi}
\newcommand{\swingup}{\texttt{swing-up}\xspace}
\newcommand{\swimmer}{\texttt{Swimmer-v4}\xspace}
\newcommand{\dmcontrol}{\texttt{dm\_control}\xspace}
\newcommand{\gym}{\texttt{{gymnasium}}\xspace}
\newcommand{\hydrogym}{HydroGym\xspace}
\newcommand{\airfoil}{3D Airfoil\xspace}
\newcommand{\hydrogymGPU}{HydroGym-GPU\xspace}
\newcommand{\pysindy}{\texttt{PySINDy}\xspace}

\newcommand{\TheAppendix}{Supplementary\xspace}

\newcommand{\CustomReference}[1]
    {#1\xspace}

\newcommand{\CustomSection}{§}

\newcommand{\appendixAlgoRef}
    {
        \CustomReference{Algorithm 1}%
    }

\newcommand{\appendixBackground}
    {
    \CustomReference{%
        \TheAppendix \CustomSection2}%
    }

\newcommand{\appendixEnvs}
    {
    \CustomReference{%
        \TheAppendix \CustomSection3}%
    }

\newcommand{\appendixBenchmarking}
    {
    \CustomReference{%
        \TheAppendix \CustomSection4}%
    }

\newcommand{\appendixBenchmarkingMBRL}
    {
    \CustomReference{%
        \TheAppendix \CustomSection4A}%
    }

\newcommand{\appendixBenchmarkingPBT}
    {
    \CustomReference{%
        \TheAppendix \CustomSection4B}%
    }

\newcommand{\appendixDynaTable}
    {
        \TheAppendix \CustomReference{Table 2}%
    }

\newcommand{\appendixClockTable}
    {
        \TheAppendix \CustomReference{Table 4}%
    }

\newcommand{\appendixDynamics}
    {
    \CustomReference{%
        \TheAppendix \CustomSection5}%
    }

\newcommand{\appendixRewards}
    {
    \CustomReference{%
        \TheAppendix \CustomSection6}%
    }

\newcommand{\appendixPolicy}
    {
    \CustomReference{%
        \TheAppendix \CustomSection7}%
    }

\newcommand{\appendixPolicySwingUp}
    {
    \CustomReference{%
        \TheAppendix \CustomSection7B}%
    }

\newcommand{\appendixFigDynPhase}
    {
        \CustomReference{\TheAppendix Fig. 13}% MAKE SURE THIS IS RIGHT
    }

\newcommand{\keywords}[1]
{\centerline{
  \small	
  \textbf{\textit{Keywords: }} #1
}
}

\usepackage{bibunits}
\begin{document}

\begin{bibunit}[unsrt]

% \bibliographyunit %[\chapter]

\twocolumn[
    \maketitle
    \begin{@twocolumnfalse}
        
        \vspace{-30pt}
        % ------------------------------------------------
        % Abstract
        % ------------------------------------------------
        \begin{abstract}
            Deep reinforcement learning (DRL) has shown significant promise for uncovering sophisticated control policies that interact in complex environments, such as stabilizing a tokamak fusion reactor or minimizing the drag force on an object in a fluid flow. However, DRL requires an abundance of training examples and may become prohibitively expensive for many applications. In addition, the reliance on deep neural networks often results in an uninterpretable, black-box policy that may be too computationally expensive to use with certain embedded systems. Recent advances in sparse dictionary learning, such as the sparse identification of nonlinear dynamics (SINDy), have shown promise for creating efficient and interpretable data-driven models in the low-data regime. 
            In this work we introduce SINDy-RL, a unifying framework for combining SINDy and DRL to create efficient, interpretable, and trustworthy representations of the dynamics model, reward function, and control policy. We demonstrate the effectiveness of our approaches on benchmark control environments and flow control problems, including gust mitigation on a 3D NACA 0012 airfoil at $Re=1000$. 
            SINDy-RL achieves comparable performance to modern DRL algorithms using significantly fewer interactions in the environment and results in an interpretable control policy orders of magnitude smaller than a DRL policy. \\
            % \noindent\textbf{Keywords:} Reinforcement learning, sparse identification of nonlinear dynamics, machine learning, model-based RL, deep reinforcement learning
        \end{abstract}
        \vspace{20pt}
    \end{@twocolumnfalse}
]{
  \renewcommand{\thefootnote}%
    {\fnsymbol{footnote}}
  \footnotetext[1]{\thanks{Corresponding author (nzolman@uw.edu)}}
}

% \nopagebreak[4] \twocolumn 

% \saythanks
% ------------------------------------------------
% Intro
% ------------------------------------------------
% \section{Introduction}
\label{intro}
Much of the success of modern technology can be attributed to our ability to control dynamical systems: designing safe biomedical implants for homeostatic regulation, gimbling rocket boosters for reusable launch vehicles, operating power plants and power grids, industrial manufacturing, among many other examples. Recently, advances in machine learning and optimization have rapidly accelerated our ability to tackle complicated data-driven tasks---particularly in the fields of computer vision~\cite{szeliski2022computer} and natural language processing~\cite{khurana2023natural}. Reinforcement learning (RL) is at the intersection of both machine learning and optimal control, and the core ideas of RL date back to the infancy of both fields. An RL agent iteratively improves its control policy by interacting with an environment and receiving feedback about its performance on a task through a reward. Deep reinforcement learning (DRL) has shown particular promise for uncovering control policies in complex, high-dimensional spaces
    \cite {Kober2012book,dutta2018reinforcement,recht2019tour,agarwal2019reinforcement,
    % Brunton2022book,
    van2016deep,wang2016dueling,qureshi2018adversarial,cheng2018fast}.
DRL has been used to achieve super-human performance in games 
   ~\cite{mnih2015human,silver2017mastering, silver2018general, berner2019dota, vinyals2019grandmaster} 
and drone racing 
   ~\cite{kaufmann2023champion}, 
to control the plasma dynamics in a tokamak fusion reactor
   ~\cite{degrave2022magnetic}, 
to discover novel drugs 
   ~\cite{popova2018deep}, 
and for many applications in fluid mechanics
    \cite {gazzola2014reinforcement,colabrese2017flow,Pivot2017aiaa,verma2018efficient,biferale2019zermelo,novati2019controlled,fan2020reinforcement,rabault2020deep,beintema2020controlling,novati2021automating, bae2022scientific}.
However, these methods rely on neural networks and typically suffer from three major drawbacks: (1) they are infeasible to train for many applications because they require millions---or even billions~\cite{vinyals2019grandmaster}---of interactions with the environment; (2) they are challenging to deploy in resource-constrained environments (such as embedded devices and micro-robotic systems) due to the size of the networks and need for specialized software; and (3) they are ``black-box'' models that lack interpretability, making them untrustworthy to operate in safety-critical systems or high-consequence environments. 
In this work we improve the sample efficiency of reinforcement learning algorithms, even for high-dimensional problems, by leveraging sparse dictionary learning. Specifically, we build small, interpretable surrogate models for the environment dynamics, reward, and policy.
% In this work, we use sparse dictionary learning to build small, interpretable models for developing sample efficient model-based reinforcement learning algorithms. 

There has been significant research into reducing the amount of experience needed to train RL policies, such as offline RL~\cite{levine2020offline}, experience replay methods ~\cite{schaul2016prioritized,rolnick2019experience,andrychowicz2017hindsight}, transfer learning  ~\cite{zhu2023transfer}, and meta-learning~\cite{wang2016learning, finn2017model}. Training in a low-fidelity representation of the  environment is perhaps the most common way to reduce the number of interactions in a full-order environment. 
However, there are many cases where an analytic reduced-order model does not exist and the dynamics must be learned from data to create a \textit{surrogate} representation of the environment. Dyna-style model-based reinforcement learning (MBRL) algorithms iteratively switch between learning and improving a surrogate model of the environment and training model-free policies inside the surrogate environment by generating ``imaginary'' experience~\cite{sutton1990integrated}. Deep MBRL algorithms have shown significant promise for reducing sample complexity on benchmark environments~\cite{wang2019benchmarking} by simultaneously training neural network models of the environment. Although neural network models have recently become popular and are gaining wide adoption over traditional modeling methods, they are still overparameterized, data-inefficient, and uninterpretable. 

\begin{figure*}[t]
    \centering
    \includegraphics[width=0.88\textwidth]{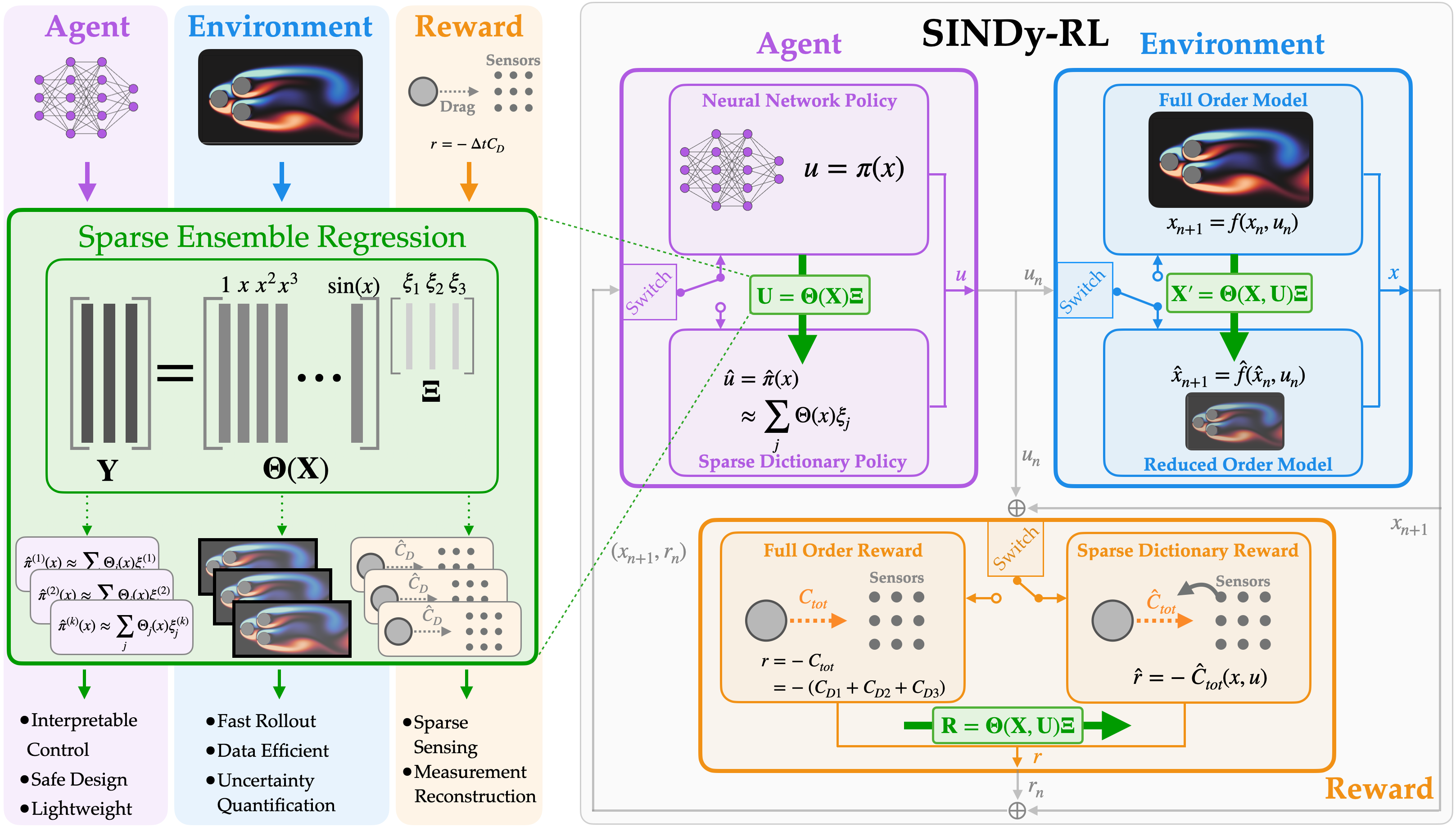}
    \caption{\textit{Left}: SINDy-RL provides a unifying set of methods for creating efficient, interpretable models of (1) the environment dynamics, (2) the reward function, and (3) the control policy through sparse ensemble dictionary learning. Each SINDy-RL method can be used together or independently depending on the context.
    \textit{Right}: Schematic of SINDy-RL logical flow. Switches indicate which model should be used; the depicted configuration is used for training a neural network policy using a surrogate environment and reward.   
    } \label{fig:overview}
    \vspace{-0.1in}
\end{figure*}

In contrast, sparse dictionary learning provides an efficient and interpretable alternative to learn models from data, as in the sparse identification of nonlinear dynamics (SINDy)~\cite{brunton2016discovering}.  
Sparse dictionary learning is a type of symbolic regression that learns a representation of a function as a sparse linear combination of pre-chosen candidate dictionary (or ``library") functions.  
A sparse, symbolic model lends itself naturally to interpretation and analysis---especially for physical systems where the dictionary terms have physical meaning.

Importantly, SINDy has been extended to systems with control and used to design model predictive control (MPC) laws~\cite{kaiser2018sparse, lore2023time}. SINDy methods are incredibly efficient---both for model creation and deployment---making them promising for both online learning and resource-constrained control. In particular, SINDy has been used for online simultaneous dynamics discovery and optimal control~\cite{farsi2020structured}.  Recent work used a single SINDy model to accelerate DRL and demonstrated its use on simple DRL benchmarks~\cite{arora2022model}. In this work, we generalize this DRL framework to include \textit{ensembles} of dictionary models of both the dynamics \textit{and reward} to accelerate learning in the low-data limit and quantify uncertainty. 

\paragraph{Our Contributions.}
% \textbf{Our Contributions:} 
In this work, we develop methods at the intersection of sparse dictionary learning and DRL for creating trustworthy, interpretable, efficient, and generalizable models that operate in the low-data limit. Building on recent advances in ensemble dictionary learning~\cite{fasel2022ensemble}, we first introduce a Dyna-style MBRL algorithm that fits an ensemble of SINDy models to approximate an environment's dynamics and uses modern model-free reinforcement learning to train agents in the surrogate environments. In systems where the reward is difficult to measure directly from the observed state (e.g. limited sensor information for flow control on an aircraft), we augment the Dyna-style algorithm by learning an ensemble of sparse dictionary models to form a surrogate reward function. Finally, after training a DRL policy, we use an ensemble of dictionary models to learn a lightweight, symbolic policy, which can be readily transferred to an embedded system. Figure \ref{fig:overview} provides a schematic of the SINDy-RL framework. We evaluate our methods on benchmark environments for continuous control from mechanical systems using the \dmcontrol~\cite{tunyasuvunakool2020dm_control} and \gym suites~\cite{brockman2016openai} as well as fluid systems from \hydrogym~\cite{HydroGym} and \hydrogymGPU \cite{HydroGym-GPU}. We demonstrate that our methods can:
\begin{itemize}  \setlength\itemsep{0em}
    \item Improve sample efficiency by orders of magnitude for training a control policy by leveraging surrogate experience in an E-SINDy model of the environment.
    \item Leverage the efficiency of the surrogate models to accelerate expensive hyperparameter tuning.
    \item Learn a surrogate reward when the reward is not directly measurable from observations.
    \item Reduce the complexity of a neural network policy by learning a sparse, symbolic surrogate policy, with comparable performance and smoother control. % and improved consistency.
    \item Quantify the uncertainty of models and provide insight into the quality of the learned models.
\end{itemize}

By leveraging sparse dictionary learning in combination with deep reinforcement learning, it can become feasible to rapidly train control policies in expensive, data-constrained environments while simultaneously obtaining interpretable representations of the dynamics, reward, and control policy.

% ------------------------------------------------
% Background
% ------------------------------------------------
\section{Background}
\label{background}
Here we provide a brief summary of relevant background in reinforcement learning and sparse nonlinear modeling; a more detailed discussion can  be found in \appendixBackground. 
\paragraph{Reinforcement Learning.} Reinforcement learning (RL) comprises a family of methods where an agent learns a policy, $\RLpolicy$, to perform a task through repeated interaction with an environment, $\env$.
Explicitly, at each environment state $\RLstate$, the agent samples an action $\RLact_n \sim \pi(\RLstate_n)$ and executes it in the environment, producing a new state $\RLstate_{n+1}$ and reward $\RLrew_n$---an indication of how well the agent performed at that time. 
We define the value function to be the expected future return for taking actions from the policy,  
$
V_\RLpolicy(\RLstate) 
    = \mathbb{E}\left(
        \sum_{k=0}^\infty \gamma^k \RLrew_k
        \bigg \vert \RLstate_0 = \RLstate
        \right)
$ 
 where $0 < \gamma \leq 1$ is the \textit{discount factor}. RL methods seek a policy that maximizes this quantity, which can be a very challenging optimization problem, especially in the case of high-dimensional state-spaces, continuous action spaces, and nonlinear dynamics. Deep reinforcement learning (DRL) has made significant progress in addressing these problems by parameterizing functions as deep neural networks (DNNs), such as the policy $\RLpolicy(\RLstate) \approx \RLpolicy_\policyParams(\RLstate)$, and training on collected experiences.

\paragraph{Dictionary Learning.}
Dictionary learning is an efficient form of symbolic regression that models a function $\vy(\vx) = f(\vx)$ as a linear combination of $d$ dictionary (or ``library'') functions (such as polynomials, sines, cosines, etc.): 
$
    \vTheta (\vx) = (\theta_{1} (\vx), \theta_2 (\vx), \dots \theta_{d} (\vx) )
$. 
To learn a model from $N$ data samples of inputs 
$
    \vX 
        = [\vx_1, \vx_2, \dots, \vx_N]^T \in \mathbb{R}^{N \times m}  
$, and their associated labels
$
    \vY      
        = [\vy_1, \vy_2, \dots, \vy_N]^T \in \mathbb{R}^{N \times n}
$, we  evaluate the dictionary at the data, $\vX$:
$$
\qquad \vTheta(\vX)
                = [\theta_1 (\vX), \theta_2 (\vX), \dots, \theta_d (\vX)] \in \mathbb{R}^{N \times d}
$$
to form the linear model
$
    \vY = \vTheta(\vX) \vXi
$, 
where $\vXi \in \mathbb{R}^{d\times n}$ are the coefficients to be fit. 
Sparse dictionary learning assumes that the desired function can be well approximated by a small subset of terms in the library, i.e. $\vXi$ is a sparse matrix.
For dynamics discovery, such as SINDy \cite{brunton2016discovering,Rudy2017sciadv} (where $\vy=\frac{d}{dt}\vx$), a sparse library is physically motivated by the observation that the governing equations for most physical systems have relatively few terms. To achieve this parsimony, a sparse optimization problem is formulated:
\begin{equation} \label{eq:sparse-opt}
\vXi = \argmin_{\hat\vXi} 
    \Vert \vY -  \vTheta(\mathbf{X}) \hat{\vXi}\Vert_F^2 + \mathcal R(\hat{\vXi})
\end{equation}
where $||\cdot||_F$ is the Frobenius norm and  $\mathcal R(\vXi)$ is a sparsity-promoting regularization.

Ensemble-SINDy (E-SINDy)~\cite{fasel2022ensemble} introduced a way to select a model from an ensemble of SINDy models and is more robust to noise than SINDy, particularly in the low-data limit.
E-SINDy can be generalized to arbitrary dictionary models by considering ensembles: 
$
    \vY^{(k)} 
        = \vTheta^{(k)}  (\vX^{(k)}) \vXi^{(k)}
$, for $k = 1 \dots N_e$. 
Treating the coefficients, $\vXi$, as random variables, the ensemble acts as an empirical 
approximation for the distribution of likely $\vXi$ values. From this, one can derive an efficient framework for analytically approximating a model's pointwise variance. 
Explicitly, for $\vXi = [\vxi_1, \dots \vxi_n]$, and letting $\text{Cov}(\vxi_i)$ denote the sample covariance matrix of $\vxi_i$ computed from the ensemble, the uncertainty of $\vy$ given $\vx$ is: 
\begin{equation}\label{eq:multi-var}
    \text{Tr} \left(\text{Var}_\vXi[\vy(\vx) | \vx] \right)
        = \sum_{i=1}^n \vTheta(\vx)\text{Cov}(\vxi_i) \vTheta(\vx)^T.
\end{equation}
It is important to note that there have been many proposed variants of SINDy to make the algorithm more robust~\cite{schaeffer2017sparse, reinbold2020using, messenger2021weak, messenger2021weakGalerikin, kaptanoglu2021promoting, forootani2023robust}. These methods are generally compatible---if not synergetic---with E-SINDy; we therefore only present the simplest formulation in this work, though a practitioner may seek to amend our framework with a more specialized variant to best suit their purpose.

% ------------------------------------------------
% Methods
% ------------------------------------------------
\section{SINDy-RL: Sparse Dictionary Learning for RL} 
\label{methods}
In this work, we introduce SINDy-RL, a unifying perspective for applying sparse dictionary learning to DRL control tasks. We separately approximate the environment dynamics, reward function, and the final learned neural network policy using ensemble sparse dictionary learning. 
We frequently use the ``hat'' notation to indicate a surrogate approximation of a function, i.e. $\hat f(x) \approx f(x)$. 

\begin{algorithm}
\renewcommand{\thealgorithm}{1}
    \caption{(Dyna-style SINDy-RL)}
    \begin{algorithmic}[1]
        \Input 
            \State 
                $\env$
                    \COMMENT{full-order environment}
            
                $N_{\text{off}}, N_{\text{collect}}$
                    \COMMENT{\# of off- and on-line policy steps}

                $n_{\text{batch}}$
                   \COMMENT{\# of policy iters / SINDy update}
                
                $\Theta$
                    \COMMENT{dictionary functions}

                $\mathcal A$
                    \COMMENT{policy optimization algorithm}

                $\pi_0$
                    \COMMENT{default policy}

                \medskip
                
        \Step 
            \State \textbf{Initialize Surrogate Environment}
            \NoNumber \State $\mathcal{D}_{\text{off}}$ = $\text{CollectData}(\env$, 
                                                $\pi_0$, 
                                                $N_{\text{off}})$  
            \NoNumber\State $\mathcal{D}$ = $\text{InitializeDatastore}(
                                        \mathcal{D}_{\text{off}})$                                            
            \State $\Xi$ = 
                                $\text{ESINDy}(\mathcal{D}$, $\Theta)$
            \State $\hat{\env}$ = Surrogate($\Xi$)
        \medskip
        \Step 
            \State \textbf{Model Improvement}
            \NoNumber
            \State $\pi$ = InitializePolicy()
            \While{not done:}
                \State $\pi$ = $\mathcal{A}(\hat{\env}, \pi, n_{\text{batch}})$
                    
                \State $\mathcal{D}_{\text{on}}$ = $\text{CollectData}(\env, \pi, N_{\text{collect}})$
                \State $\mathcal{D}$ = UpdateStore$(\mathcal{D}$, $\mathcal{D}_{\text{on}})$
                \State $\Xi$ = ESINDy($\mathcal{D}$, $\Theta$)
                \State $\hat{\env}$ = Surrogate($\Xi$)
            \EndWhile
        \medskip
        \Output \State Optimized policy, $\pi$, and SINDy environment, $ \hat\env$. 
    \end{algorithmic}
    \label{algo:DynaSINDy}
\end{algorithm}

\paragraph{Approximating Dynamics.}
We propose a Dyna-style MBRL algorithm where we iteratively improve a dynamics model and learned policy. First, we collect offline data samples from the full-order environment by deploying a default policy (such as random input, an untrained neural network, Schroeder sweep~\cite{schroeder1970synthesis}, etc.). We use the collected data to fit an ensemble of SINDy models to initialize a surrogate environment. Next we iteratively train a policy for a fixed number of policy updates using the surrogate environment with a policy optimization algorithm, such as proximal policy optimization, (PPO) \cite{schulman2017proximal}. By forcing the agent to only interact in the surrogate environment, we offload the majority of the expensive sample collection to the lightweight E-SINDy surrogate. Finally, we deploy the trained policy to the full-order environment and collect data for evaluation and use the newly collected data to update the E-SINDy models. 
We repeat this process, iteratively updating and improving both the E-SINDy model and policy. In this work, we specify a fixed number of policy iterations before evaluating the agent, though a practitioner may consider an adaptive number of updates based on training metrics in the surrogate environment. A full summary of this process can be found in Algorithm \ref{algo:DynaSINDy}.

Each dynamics model in the ensemble is fit using SINDy with control (SINDy-C)~\cite{kaiser2018sparse} and STRidge \cite{Rudy2017sciadv}. 
Either a continuous or discrete model can be fit; however, we learn discrete-time models 
where next-step updates are predicted explicitly, rather than integrating a continuous-time model forward in time, for ease of deploying in the environment. It is common for nonlinear dynamics models---especially learned models---to grow unbounded over long time horizons unless stability guarantees are enforced \cite{kaptanoglu2021promoting}. To accommodate this limitation, we bound the state space and reset the surrogate environment during training if a trajectory exits the bounding box.

\paragraph{Approximating Rewards.}
For Dyna-style MBRL, it is assumed that the reward function  can be directly evaluated from observations of the environment. However, there are cases in which the reward function cannot be readily evaluated because the system may only be partially observable due to missing sensor information---as is the case in many fluids systems. There has been significant work on creating proxy rewards for controls tasks using \textit{reward shaping}~\cite{ng1999policy} and learning an objective function through inverse reinforcement learning~\cite{arora2021survey}. We propose learning a proxy reward, $\hat{R}(\vx_{k+1}, \vu_{k})$, with supervised sparse dictionary learning when there are offline evaluations of the reward available: $r_k = R(\vx_{k+1}, \vu_k)$. Our implementation uses sparse ensemble dictionary learning under the assumption that there is a sparse deterministic relationship between the reward function and observations from the environment. We incorporate this into Algorithm \ref{algo:DynaSINDy} by learning the reward function alongside the dynamics to create the surrogate environment.

% ------------------------------------------------
% Results FIGURE
% ------------------------------------------------
\begin{figure*}[t!]
    \centering
    \includegraphics[width=.95\textwidth]{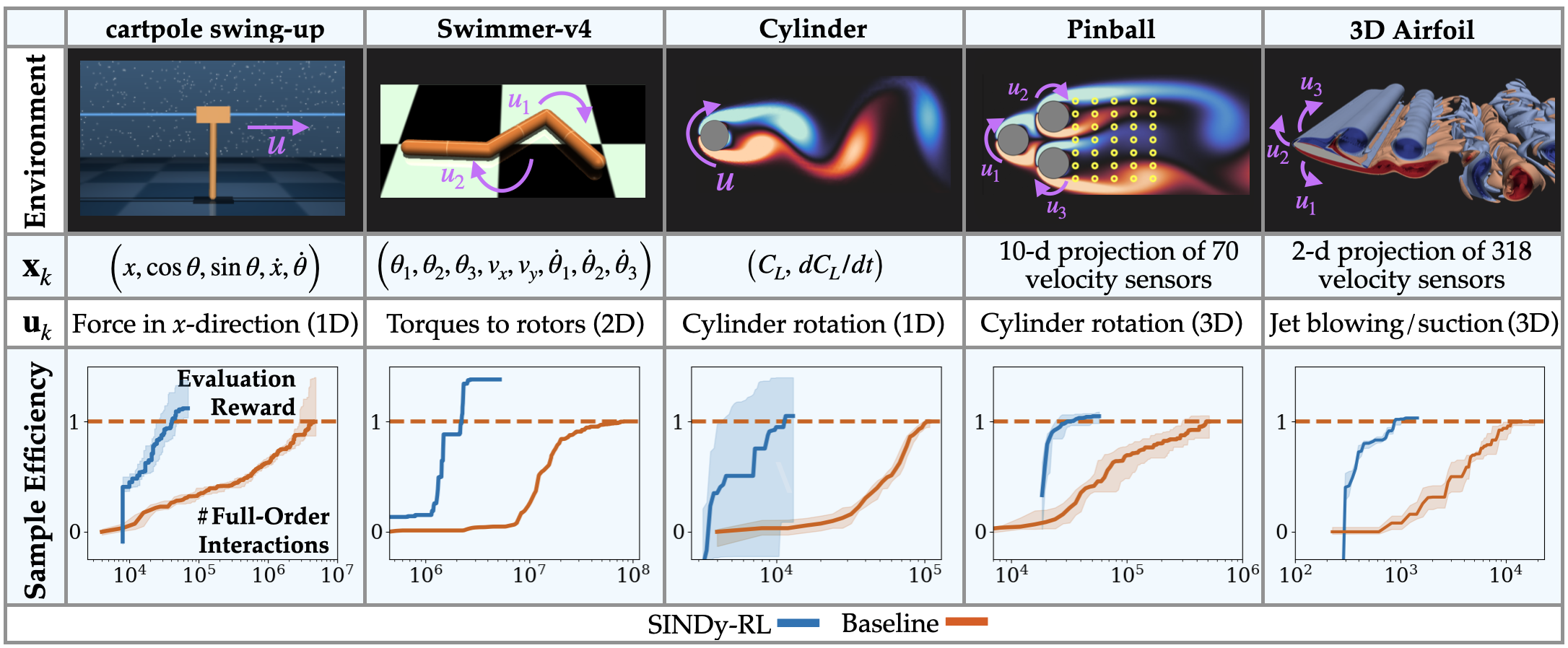}
    \caption{\textbf{Overview of environments and results.} 
    \textit{Environment.} The environments used and their corresponding observations, $\RLstate_k$, and actions $\RLact_k$. For Pinball, the yellow circles indicate the location of the sensors. A detailed schematic of the \airfoil can be found in Figure \ref{fig:naca}.
    \textit{Sample Efficiency:} SINDy-RL sample efficiency comparison with a baseline DRL approach (including initial offline data collection detailed in \appendixDynaTable). Shaded regions indicate the performance for the 25th to 75th quantiles among independent agents. Evaluation rewards are scaled based on the median Baseline DRL performance.  Swimmer comparison uses PBT with top-performing agent.   
    }
     \label{fig:summary}   
     \vspace{-.1in}
\end{figure*}

\paragraph{Approximating Policies.}
Taking inspiration from behavior cloning algorithms in imitation learning~\cite{hussein2017imitation}, we fit a sparse dictionary model approximation of the \textit{final} learned policy, $\pi_\policyParams$. While imitation learning attempts to mimic the policy of an expert actor with a neural network student, we instead use the learned neural network as our expert and the dictionary model as our student, resulting in a lightweight, symbolic approximation. A dictionary model is not as expressive as a neural network; however, there has been previous work indicating that even limiting to purely linear policies can be a competitive alternative to deep policy networks~\cite{mania2018simple, rajeswaran2017towards}, showing that even complicated control tasks may have a simple controller. Likewise, there has been recent investigation into approximating neural networks with polynomials through Taylor expansion~\cite{zhu2022nn}, which  have shown to provide sufficiently robust approximations. These reduced representations can be orders of magnitude smaller than a fully-connected neural network and they can be efficiently implemented and deployed to resource-constrained environments, such as embedded systems.

Many model-free algorithms assume a stochastic policy for optimally interacting with a stochastic environment and encouraging exploration; we proceed with a sparse ensemble fit of the expectation:  $\hat \pi(\vx) \approx \mathbb{E}[\pi_\policyParams(\vx)]$. Because there is no temporal dependence on $\pi_\policyParams$, we can assemble our data and label pairs $(\vx, \mathbb{E}[\pi_\policyParams(\vx)])$ by evaluating $\pi_\policyParams$ for \textit{any} $\vx$. 
To build such a dataset, we sample points from trajectories of the agent interacting in the environment, ensuring that the states are directly relevant to the controls task. To avoid the cost of collecting more data from the full-order environment, $\env$, we propose \textit{sampling new trajectories from a learned ensemble of dynamics models}. Explicitly, we sample $N_\tau$ trajectories by propagating the E-SINDy model
$$
    \tau^{(i)} = \{(\vx_k^{(i)}, \vu_k^{(i)}, \vx_{k+1}^{(i)}, r_k^{(i)})\}_{k=1}^{T_i} , \qquad i = 1, \dots N_\tau, 
$$
$$
    \vu_k = \mathbb{E}[\pi_\policyParams(\vx_k)], \qquad \vx_{k+1} = \hat{f}(\vx_k, \vu_k)
$$
 and use the collected data $(\vx_k, \vu_k)$ from each $\tau$ to fit the dynamics model. During the DRL training, the neural network policy may overfit to specific regions of the space and bias our data collection; thus, we draw inspiration from tube MPC~\cite{lopez2019dynamic}, where an MPC controller attempts to stay within some bounded region of a nominal trajectory. 
Instead of only sampling points from the surrogate trajectories, we can sample plausible points in a neighborhood of the trajectories to create a more accurate approximation while still avoiding regions of the space that can no longer be trusted.

% ------------------------------------------------
% Results
% ------------------------------------------------
\section{Results}
\label{results}
We evaluate our methods on five environments depicted in Figure \ref{fig:summary}: 
(1) \dmcontrol \swingup \cite{tunyasuvunakool2020dm_control} balances a pole on a cart in the unstable upright position starting from the stable down position at rest,
(2) \gym \swimmer\cite{brockman2016openai} controls a 3-segment robot to travel as far as possible in the horizontal direction (along the $x$-axis) for a fixed time, 
(3) \hydrogym Cylinder \cite{HydroGym} reduces the drag force, $C_D$, of a rotating cylinder in an unsteady fluid flow at $Re=100$ with measurements of the lift force, $C_L$, (4) \hydrogym Pinball reduces the net drag force $C_{D1} + C_{D2} + C_{D3}$  on the system of cylinders to stabilize the quasi-periodic flow at $Re=100$,
and (5) \hydrogymGPU \airfoil \cite{HydroGym-GPU} mitigates the effect of a large incoming gust in an unsteady flow at $Re=1000$ by minimizing $|C_L - C^{ref}_L| + \frac{1}{4}|C_D- C^{ref}_D|$, i.e. deviations of $C_L$ and $C_D$ from reference values.
For Pinball and \airfoil, state information is obtained from a mesh of velocity probes in the flow, forming a 70- and 318-dimensional measurement space, respectively. This data is projected onto the leading singular value decomposition modes (SVD) and use the coefficients, $a_i = \mathbf v_i^T \RLstate$, to form a 10- and 2-dimensional observation space for the respective systems.
Details for each environment can be found in \appendixEnvs.
Whereas the \swingup reward is analytically expressible in terms of the observed variables, all other rewards are computed from environment information that cannot retrieved analytically from the observation. We treat the rewards as functions of the observations and actions by approximating them with an ensemble of sparse models.

In Section \ref{results:benchmarking}, we use our method to train SINDy-RL agents using Dyna-style MBRL with surrogate dynamics and reward functions, and baseline sample efficiency against other approaches by comparing the amount data collected from the full-order environment.  Sections \ref{results:dynamics} and \ref{results:rewards} explore the evolution of the learned dynamics and rewards during training. In Section \ref{results:policy}, we use the learned neural network policies from Algorithm \ref{algo:DynaSINDy} to obtain surrogate dictionary policies. Finally, in Section \ref{results:uq}, we examine how to quantify the variance of dictionary models and provide additional insight into learned models. 

\subsection{Benchmarking Sample Efficiency} \label{results:benchmarking}

\begin{figure}[t]
    \centering
    \includegraphics[width=.95\linewidth]{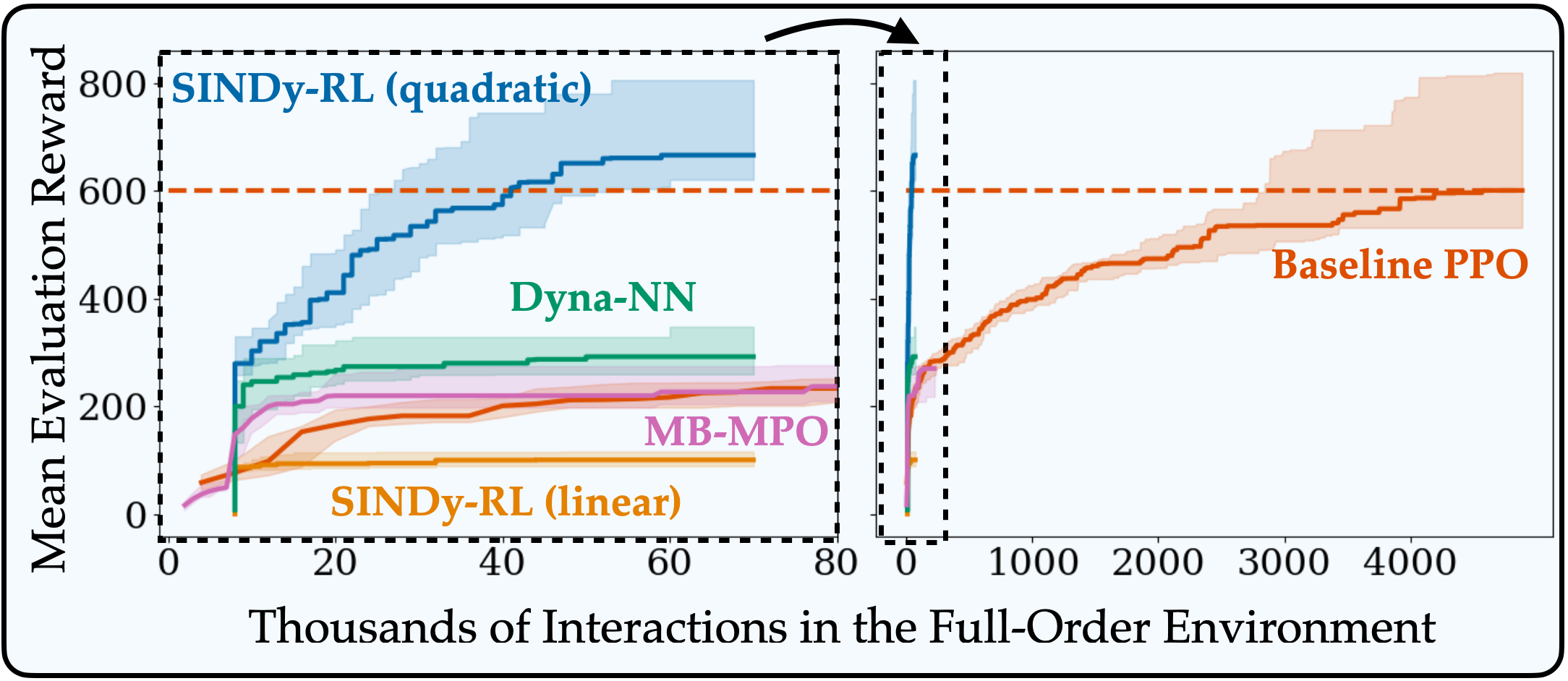}
    \cprotect 
    \caption{\textbf{Swing-up Sample Efficiency}.
    Comparison of the number of interactions in the full-order \swingup environment for different algorithms.
    For each algorithm, twenty independent seeds were run and periodically evaluated on 5 randomly generated episodes. Each agent's best average performance was tracked, updated, and recorded. The shaded area corresponds to the 25th and 75th quantiles, and the bold lines are the median. The dashed red line corresponds to the median best performance of the baseline PPO agent before training was stopped.
    }
     \label{fig:cartpole-baseline-best}   
\end{figure}

For every algorithm and experiment, multiple independent instantiations are run to provide a distribution of performance across random seeds.\footnote{Twenty instantiations were used for all environments, except \airfoil where we are limited to four instantiations due to the computational demand of the environment.} Specific details about the training, hyperparameters, and more can be found in \appendixBenchmarking. 

% ----------------------
% Swing-up benchmark
% ----------------------
\paragraph{Accelerating Training.} 
In Figure \ref{fig:cartpole-baseline-best}, we compare our proposed SINDy-RL method from Algorithm \ref{algo:DynaSINDy} with a quadratic dynamics library
to four different baseline experiments: (1)  model-free proximal policy optimization (PPO), (2) Algorithm \ref{algo:DynaSINDy} with a linear SINDy library, (3) Algorithm \ref{algo:DynaSINDy} where SINDy models are replaced with neural network dynamics models (i.e. ``Dyna-NN''), and (4) RLlib's implementation of Model-Based Meta-Policy Optimization (MB-MPO)~\cite{clavera2018model}. 
PPO is used as the policy optimization algorithm, $\mathcal{A}$, for all Dyna-style experiments with identical hyperparameters between comparisons. 
As shown in Figure \ref{fig:cartpole-baseline-best}, SINDy-RL with a quadratic library learns a control policy with $\mathbf{100 \times}$ \textbf{fewer interactions in the full-order environment} compared to model-free RL and greatly outperforms other model-based approaches. 
The linear models are incapable of approximating the global dynamics since there are multiple fixed points.
In contrast, while neural network models should be more expressive, it appears that both the MB-MPO and Dyna-NN baselines fail to accurately capture the dynamics sufficient to achieve the task. 
For both comparisons, it is suspected that the dynamics may be captured with significantly larger data collections---as  shown in previous work comparing SINDy with neural network models for MPC~\cite{kaiser2018sparse}. 
An investigation of the $N_{\text{collect}}$ and $n_{\text{batch}}$ hyperparameters from our method can be found in \appendixBenchmarkingMBRL.

\begin{figure}[t]
    \centering
    \includegraphics[width=.95\linewidth]{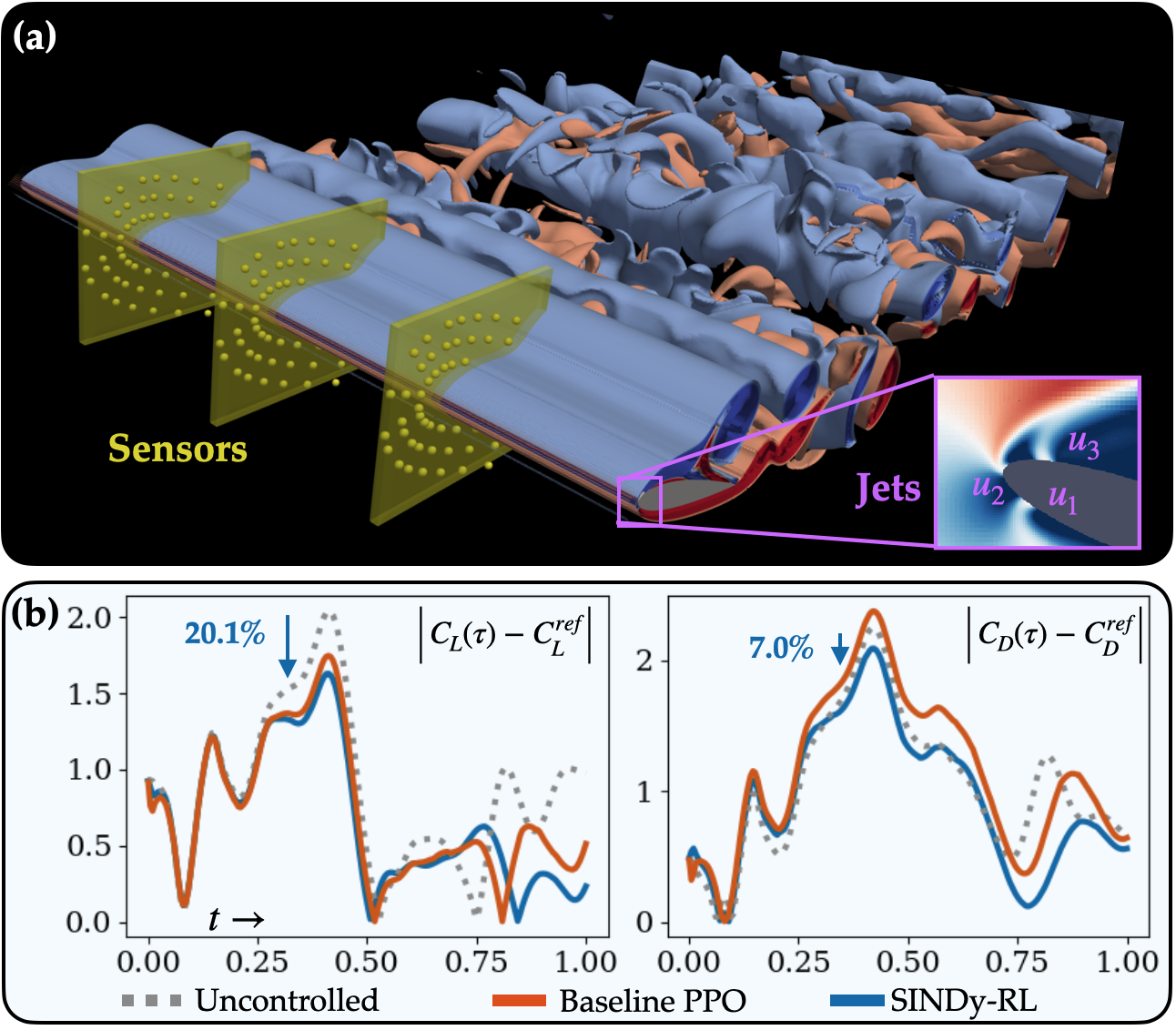}
    \cprotect 
    \caption{\textbf{3D Airfoil}. 
    (a) Schematic of the 3D Airfoil environment at $Re=1000$, $20^\circ$ angle-of-attack, and $M=0.2$. The gust profile follows a 1-cosine approach with a gust factor $G = 2.0$. Q-criterion isosurfaces of the vorticity demonstrate the presence of 3D instabilities. Sensors are restricted to three planes, each consisting of 53 2-d velocity probes. Actuation is performed using three independent jets along the spanwise direction. 
    (b) Effect of SINDy-RL policy on the normalized lift and drag forces used in the reward function. Compared to the uncontrolled environment, SINDy-RL reduces the peak error from the reference $C_L$ and $C_D$ values by 20.1\% and 7.0\% respectively. The best trained baseline PPO agent was able to reduce the peak $C_L$ error by 14.3\%, but rose the peak $C_D$ error by 5.7\%.
    }
     \label{fig:naca}   
\end{figure}

The increased sample efficiency is particularly important for applications interacting with physical systems that require humans in-the-loop to monitor and reset environments or computationally expensive simulations---such as CFD solvers---that require significant time and resources to run. The benefits of reducing the sample efficiency can be seen using the fluid flow control environments. The \airfoil environment uses the lattice Boltzmann method with over 72M cells, requiring 75GB of VRAM across four A100 GPUs. Even when using the GPUs, a single step in the full-order model takes on the order of 45s to update the environment. In comparison, the learned E-SINDy polynomials for the \airfoil require less than 10kB of RAM on CPU, and a single step takes approximately 1-4 milliseconds using NumPy ~\cite{harris2020array}---i.e. \textbf{SINDy-RL experience collection is $\mathbf{10^4 \times}$ faster than the full-order CFD models}. 
The SINDy-RL agent was $\mathbf{14.47 \times}$ more sample efficient for the \airfoil environment; with only 25 dynamics updates (i.e. iterations of Step 2 from Algorithm \ref{algo:DynaSINDy}), the median SINDy-RL agent reached the final median baseline agent performance. Physically, this was the difference between \textbf{14 hours of training using the SINDy-RL environment and 185 hours using the full-order environment}, greatly reducing the total amount of time needed to learn capable policies. 
A comparison of clock times for different aspects of training the \hydrogym environments can be found in \appendixClockTable. 

It is important to note the large variance in performance among the 20 trained SINDy-RL Cylinder agents depicted in the shaded region of Figure \ref{fig:summary}. After interacting with the full-order Cylinder environment 13,000 times, the top 50\% performing SINDy-RL agents were able to surpass the best baseline performance---some achieving 11\% drag reduction, comparable with optimized solutions from previous literature \cite{choi2002characteristics, rabault2019artificial}. However, the bottom 25\% of agents performed poorly. Surrogate dynamics models can quickly diverge---especially early in training---which can provide misleading rewards. When these effects are further coupled with unfavorable policy network initializations, agent performance can severely degrade. The overall success of the majority of the agents indicates that methods like \textit{population-based training} may be extremely effective at pruning these bad combinations of surrogate dynamics, rewards, and policies to quickly find exceptional policies using very few interactions in the full-order environment.

% ----------------------
% Swimmer benchmark (PBT)
% ----------------------
\paragraph{Accelerating Hyperparameter Tuning.}  The success of DRL to find effective control policies can often depend on the neural network initialization and choice of hyperparameters, otherwise training might lead to a suboptimal policy. To address this, it is common to use different random initializations for the network and hyperparameter tuning, and software packages have been created to facilitate these searches, such as Ray Tune~\cite{liaw2018tune}. While these tuning algorithms can be effective for uncovering sophisticated policies, they generally rely on parallelizing training and significant compute usage. Previous benchmarks ~\cite{weng2022tianshou, huang2022cleanrl} have shown that the \swimmer environment is particularly challenging to train with PPO and it has been suggested~\cite{franceschetti2022making} that learned policies for \swimmer are especially sensitive to certain hyperparameters and tuning can achieve superior performance. 
We pursue this idea by comparing population-based training (PBT)~\cite{jaderberg2017population} to improve policies for agents interacting in the full-order \swimmer environment and a SINDy-RL environment with dictionary surrogates for the dynamics and reward. Both experiments used a population of 20 policies and periodically evaluated the policies using the rollouts from the full-order model. As shown in Figure \ref{fig:summary}, the SINDy-RL training is able to achieve nearly 30\% better maximal performance than the baseline using \textbf{$\mathbf{50\times}$ fewer samples from the full-order environment}. This indicates that SINDy-RL can provide a convenient way to accelerate hyperparameter tuning in more sophisticated environments. Additional training details for PBT can be found in \appendixBenchmarkingPBT.

\begin{figure*}[t]
    \centering
    \includegraphics[width=0.99\textwidth]{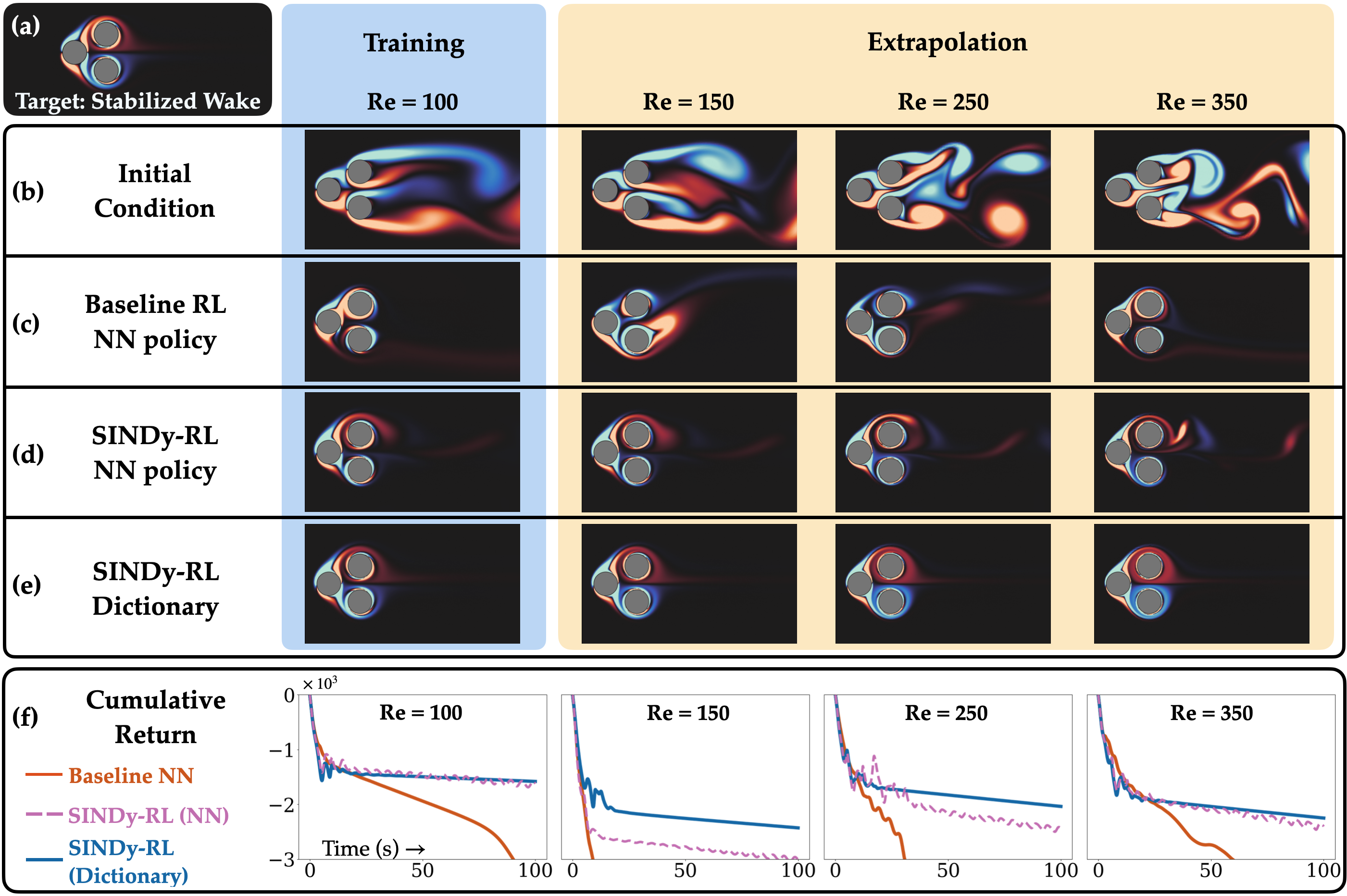}
    \caption{\textbf{Evaluation of Pinball polices for various $Re$.} 
    (a) Vorticity snapshot of the target state with the wake stabilized. 
    (b) Vorticity snapshots of initial conditions for the system at each $Re$. 
    (c)--(e) evaluation of the baseline PPO, SINDy-RL (with NN policy), and dictionary policy distilled from  from the SINDy-RL NN. Policies were only trained using $Re=100$. Each snapshot is taken after 100s of feedback control. 
    (f) Comparison of the cumulative return over time for the policies across different $Re$; accumulation of negative rewards leads to a decreasing curve.
    } \label{fig:pinball_grid}
    \vspace{-0.1in}
\end{figure*}

\paragraph{Generalization.} 
It has been well-documented that while DNNs are excellent at performing on unseen data sampled from the same data distribution used for training, i.e. \textit{interpolation}, they often fail at \textit{extrapolation} to data beyond the convex hull of the training set. In Figure \ref{fig:pinball_grid}, we investigate the ability of top-performing Pinball agents trained at $Re=100$ for $20$-second segments to extrapolate to unseen dynamics at $Re=150,250,$ and $350$ for a $100$-second evaluation. It is well-known that the dynamics of the Pinball system undergo a bifurcation at around $Re=115$ where the dynamics transition from being quasi-periodic to chaotic~\cite{deng2020low}; thus, the dynamics are fundamentally different at these $Re$ values, which can be qualitatively seen---especially at $Re=350$---in the initial conditions shown in Figure \ref{fig:pinball_grid}(b). Despite these fundamental differences, the SINDy-RL policy is able to reasonably extrapolate. While performance does degrade compared to $Re=100$, the agent is able to obtain a much better cumulative return of reward over the long trajectory compared to the baseline neural network counterpart.

% --------------------
% Dynamics
% --------------------
\subsection{Surrogate Dictionary Dynamics} \label{results:dynamics}
 We now examine the SINDy dynamics learned from agents during the training outlined above. We utilize a polynomial library for each dynamics model.  Despite none of the environments having a polynomial representation of the dynamics, the surrogate dynamics provide a sufficient representation of the environment to learn a control policy through repeated interaction. 
 A thorough investigation of the learned dynamics from each environment can be found in\appendixDynamics. 

We find that for most systems, it is imperative to periodically refit the dynamics while the agent explores control strategies. For example, with the \swingup dynamics, the agent has no information about the goal state at the unstable equilibrium early in training. In \appendixFigDynPhase, we show that the first dynamics model provides a reasonable representation of the phase portrait, but the learned unstable fixed point is offset from the ground-truth position. Without refitting the dynamics, the agent would learn to drive the system to the wrong point and never stabilize the true system. However, by deploying the learned policy on the full-order system and gathering new data, the quality of the dynamics model improves and the agent is able to complete the task both using the surrogate and full-order dynamics. Finally, unlike a neural network representation of the dynamics, our approach provides a symbolic representation of the learned dynamics. In \appendixDynamics, we provide the learned coefficients $\vXi$ and show that the \swingup dynamics model is well-represented by an Eulearian integration scheme: $\mathbf{x}_{k+1} = \mathbf{x}_k + \Delta t f(\mathbf{x}_k)$. 

It is important to note a limitation of using dictionary dynamics; for large observation spaces, the size of a polynomial dictionary increases combinatorially. For example, a quadratic dynamics library with a 318-dimensional state would have over 16M parameters\xspace
\footnote{A library of monomials with degree $\leq d$ in $m$ variables has $\binom{m+d}{d}$ terms; dynamics regression consists of $m$ equations, for a total of $m \cdot \binom{m+d}{d}$ parameters.}. Therefore for large spaces, dimensionality reduction is critical. 
We demonstrate the viability of this approach for both the Pinball and \airfoil environments by linearly projecting the 70- and 318-dimensinoal observations onto the dominant SVD modes of the system and found this to be sufficient for purposes of training. 
For more complicated dynamics, autoencoders have been found to be useful for finding nonlinear projections \cite{Champion2019pnas}.
 
\subsection{Surrogate Dictionary Rewards}
\label{results:rewards}

\begin{figure}[t]
    \centering
    \includegraphics[width=0.95\linewidth]{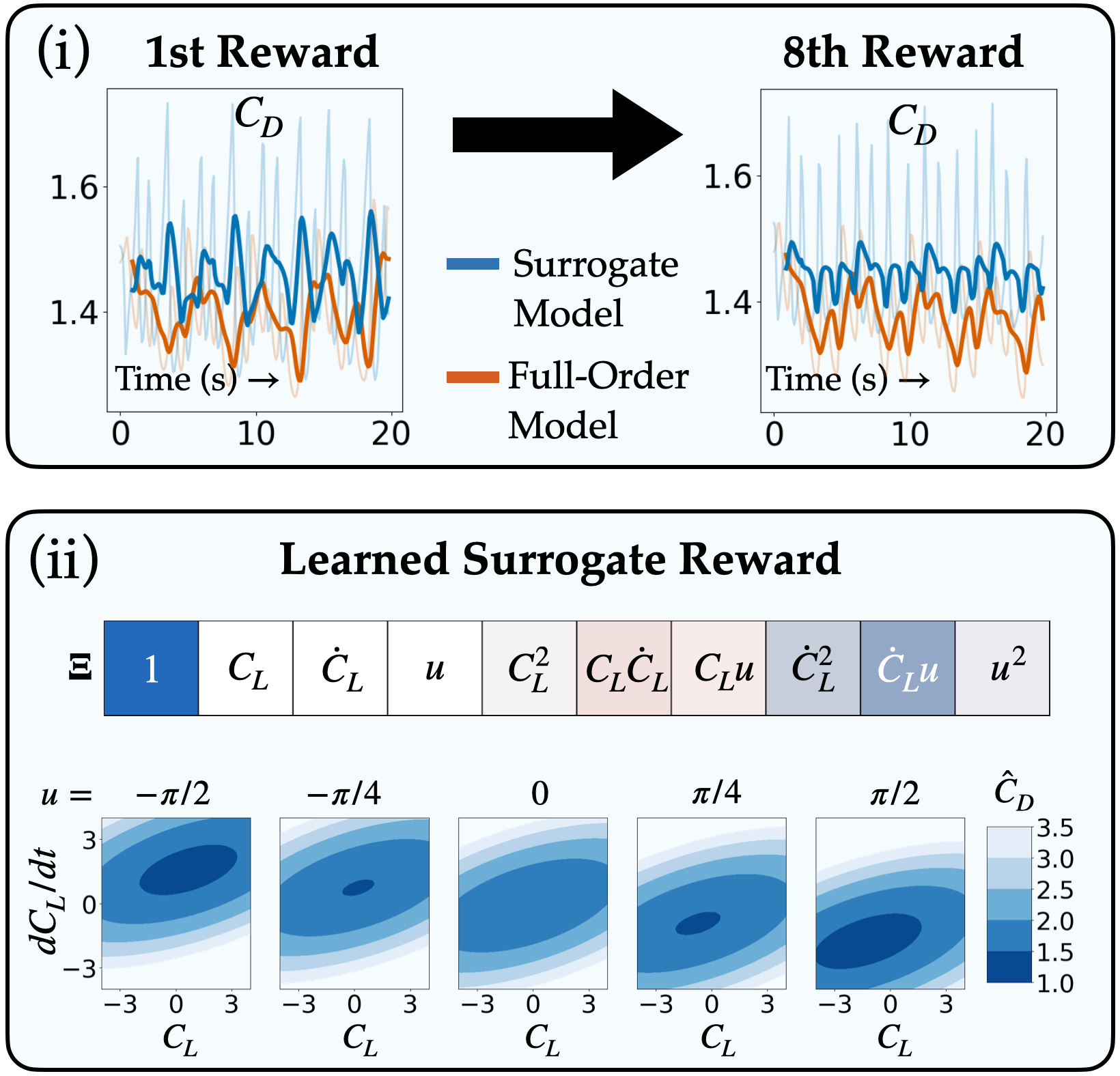}
    \cprotect \caption{\textbf{Cylinder Reward.} (\textit{i}): Comparison of the learned surrogate reward $\hat{r}(C_L, dC_L/dt, u) = -\hat{C}_D \Delta t$ for the Cylinder environment at the beginning of training and after 8 dynamics and reward updates.
    (\textit{ii, Top}): a heatmap of the median dictionary coefficients $\vXi$,
    (\textit{ii, Bottom}): a series of contour plots show the effect of control on $\hat{C}_D = -\hat{r}/\Delta t$

    }
     \label{fig:cylinder-rew}   
\end{figure}

We now consider the case where the exact rewards are not analytically expressible from the observations. In this setting, we periodically fit the surrogate dictionary reward $\hat{r} = \hat R(\vx,\vu)$ alongside and independent of the SINDy dynamics. For \swimmer, the rewards in the full-order environment are given by the agent's body-centered horizontal velocity. However, this quantity is not provided by the observation space. It has been well-documented that---even with access to the exact rewards---solving the \swimmer task is a considerable challenge because of the placement of the velocity sensors; researchers have even created their own modified versions of the environment when performing benchmarks~\cite{wang2019benchmarking}. Our method identified that a sparse reward of $\hat{r} \approx v_x$---the horizontal velocity of the \textit{leading segment} (not the body)---was a suitable surrogate and remained stable throughout the entirety of the population-based training described in Section \ref{results:benchmarking}. 

In contrast to the \swimmer environment, the \hydrogym environments are governed by a nonlinear PDE, where the rewards are scalar measurements evolving with the dynamics on the domain and are not analytically expressible in terms of the provided observations. The lack of available information makes modeling the environment and the reward a very challenging task. Just as in the discussion in Section \ref{results:dynamics}, Figure \ref{fig:cylinder-rew}(i) demonstrates the necessity of periodically updating the surrogate models. At the beginning of training, the learned reward is actually \textit{anti-correlated} with the full-order reward; however, the learned reward becomes well-correlated after subsequent updates. Despite the inability to learn the exact reward in the fluid environments, the surrogate reward provides a sufficient enough learning signal for training a comparable control policy when having access to the exact rewards.
We also highlight the interpretability of our method in Figure \ref{fig:cylinder-rew}(ii); the learned reward function is a quadratic polynomial of the observations, and we can analyze the explicit influence of the control. The bowl-shaped function (given by the quadratic) gradually shifts in response to the increasingly positive control. Furthermore, it is clear that the drag is minimized (i.e. reward is maximized) for large values of $|u|$, indicating that an optimal control strategy would apply maximal control input to stay in the bowl's minimum.
For the Pinball and \airfoil environments, the sparse sensors in the flow are not sufficient to describe the net forces acting on surfaces; indeed, the discovered reward functions are dominated primarily by terms coupling the actuation (local information) to the sensors. 
The learned rewards and a detailed investigation for each environment can be found in \appendixRewards.

\subsection{Surrogate Dictionary Policy}
\label{results:policy}

While DNN policies can find solutions to complicated control problems, they are typically over-parameterized, black-box models that lack interpretability. There have been several approaches to combine symbolic regression and DNNs to improve interpretability~\cite{cranmer2020discovering, udrescu2020ai, kim2020integration, pmlr-v80-sahoo18a, both2021deepmod, forootani2023robust}, but they tend to focus on discovering the dynamics rather than a symbolic form for a controller. 
Here, we discover a lightweight model of the control with sparse dictionary learning using the behavioral cloning method described in Section \ref{methods}; for each environment, we leverage the E-SINDy dynamics and the neural network policy obtained from Algorithm 1 to collect data for fitting the dictionary model.

Figure \ref{fig:pinball_grid} compares the performance of the SINDy-RL neural network policy to sparse dictionary policy distilled from it in the Pinball environment. As shown in Figure \ref{fig:pinball_grid}(e), the dictionary approach is able to consistently stabilize the wake across various values of $Re$ compared to its neural network counterpart. This is shown quantitatively in Figure \ref{fig:pinball_grid}(f) where the dictionary policy consistently receives higher cumulative returns during long evaluations of the policy. The final dictionary control policy is represented as a quadratic polynomial library containing 66 elements, nearly two orders of magnitude fewer parameters than the corresponding neural network policy with over 5,000 parameters. This is also considered a relatively small neural network model; for deeper or wider network, one can reasonably expect even larger gains. In \appendixPolicySwingUp, we provide further evidence that a dictionary policy can produce smoother and more consistent control inputs---outperforming the neural network when it has overfit to a particular trajectory in the \swingup environment.

However, the Cylinder environment provides an example where this method may struggle; the surrogate policy only reduces drag by about 3.7\% compared to the neural network policy's 8.7\% reduction. The original neural network agent learned a bang-bang control policy; this is very difficult to approximate with a smooth polynomial due to the bounded derivatives. One of the key challenges that supervised imitation learning faces (e.g. behavior cloning) is the issue of compounding errors during policy deployment. Because our policy approximation is a type of behavior cloning, our method inherits this challenge; the approximate policy slowly drifts away from the state-action pairs it was trained on and ultimately performs suboptimally.
A detailed investigation of the learned dictionary policies for each environment (including a comparison of different data sampling strategies used for the behavior cloning) can be found in \appendixPolicy.
%

% ------------------------------------------------
% Uncertainty Quantification
% ------------------------------------------------
\subsection{Uncertainty Quantification}
\begin{figure}[t]
    \centering
    \includegraphics[width=0.95\linewidth]{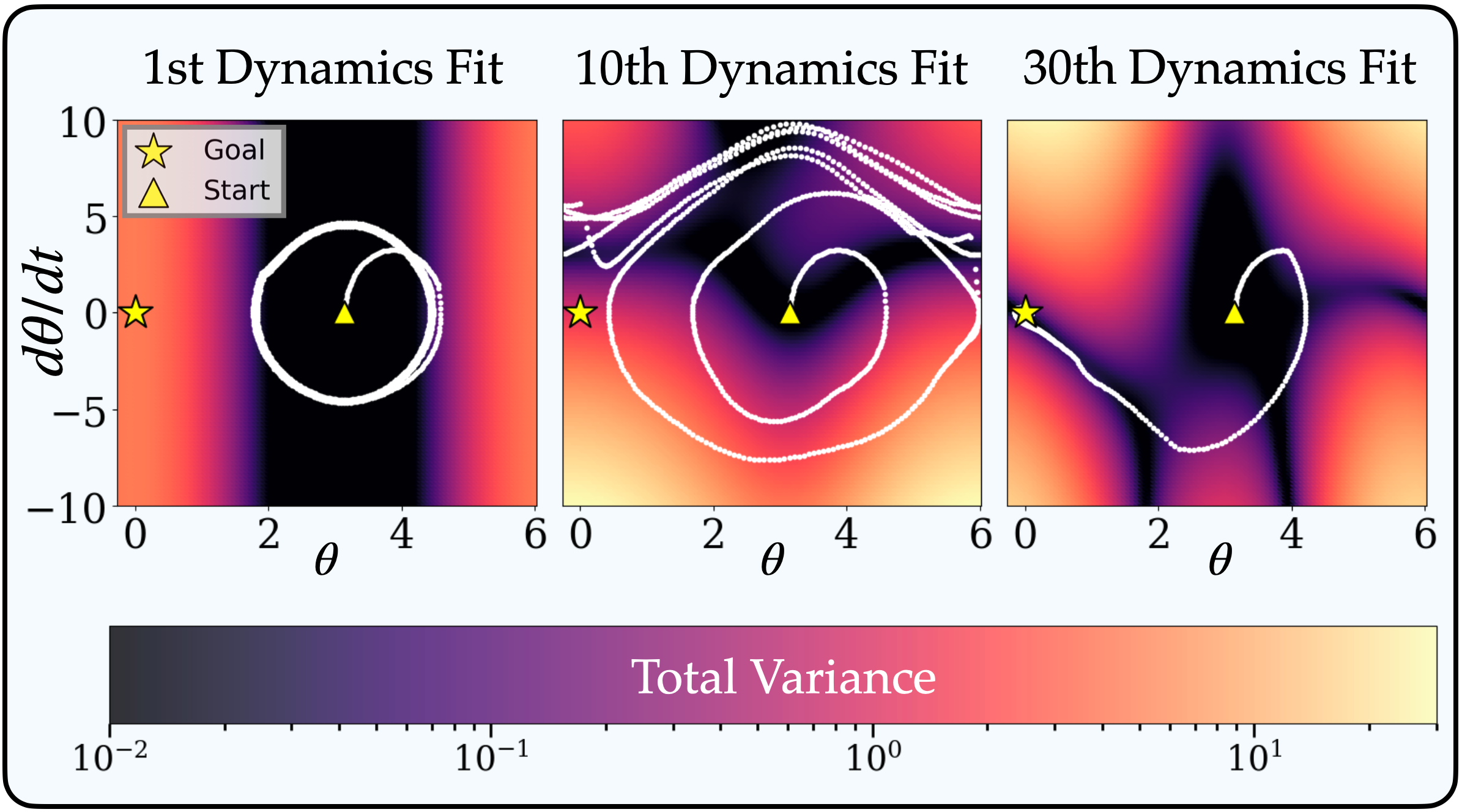}
    \cprotect \caption{\textbf{Swing-up Dynamics Variance.}
    The variance of learned dynamics across snapshots during SINDy-RL training. For ease of visualization, the landscape is evaluated at $x=dx/dt=u=0$.
    }
     \label{fig:cart-dyn-uq}   
\end{figure}

\label{results:uq}
 From Equation \ref{eq:multi-var}, we have a way of using the structure of the dictionary model to efficiently compute the pointwise variance from an ensemble of trained models. We now demonstrate how we can use this as a tool to investigate our learned functions.
In Figure \ref{fig:cart-dyn-uq}, we visualize the evolution of the uncertainty landscape for the \swingup dynamics  when training the dynamics and policy with SINDy-RL. At the beginning of training, the dynamics are confined to a narrow band of low-variance states where data has been previously collected. During training, the variance landscape changes in response to new, on-policy data. By the time the agent has learned to solve the \swingup task, the learned dynamics have become more certain and the agent takes trajectories over regions that have less variance.  

In this work we only use variance to inspect the learned dictionary functions, although there are further opportunities to \textit{exploit} the variance. There is a trade-off in sample efficiency between refining the learned dynamics and improving the policy early in training. With a poor representation of the surrogate dynamics, we risk overfitting to the surrogate and discovering policies that do not generalize well to the real environment. Analogous to curiosity-driven learning~\cite{pathak2017curiosity}, the estimated uncertainty of the dynamics and reward models may strategically guide the exploration of the environment and rapidly improve them---making it less necessary to query the environment later. Likewise, we may encourage ``risk-averse'' agents by penalizing regions with high-uncertainty---encouraging agents to take trajectories where the dynamics are trusted and agree with the full-order model.

% ------------------------------------------------
% Discussion
% ------------------------------------------------
\section{Discussion}
\label{discussion}
This work developed a unifying framework for combining SINDy (i.e., sparse dictionary learning) with deep reinforcement learning to learn efficient, interpretable, and trustworthy representations of the environment dynamics, the reward function, and the control policy, using significantly fewer interactions with the full environment. 
We demonstrate the effectiveness of SINDy-RL on several challenging benchmark control environments, including performing gust mitigation of NACA 0012 airfoil at $Re=1000$ in a 3D, unsteady environment.  

By learning a sparse representation of the dynamics, we developed a Dyna-style MBRL algorithm that could be $10-100\times$ more sample efficient than a model-free approach, while maintaining a significantly smaller model representation than a black-box neural network model. When the reward function for an objective is not easily measurable from the observations---e.g. with only access to sparse sensor data---SINDy-RL can simultaneously learn dictionary models of the reward and dynamics from the environment for sample-efficient DRL. We also demonstrated that SINDy-RL can learn a sparse dictionary representation of the control policy for certain tasks; the resulting sparse policy is 
(1) orders of magnitude smaller than the original neural network policy, 
(2) has smoother structure, 
and 
(3) is inherently more interpretable. With a lightweight polynomial representation of the control policy, it becomes more feasible to transition to embedded systems and resource-limited applications. 
The interpretable representation of the dictionary policy also facilitates classical sensitivity analysis of the dynamics and control, such as quantifying stability regions and providing robust bounds---an especially important quality in high-consequence and safety critical environments. Finally, we have shown that for dictionary models, it is possible to analytically compute the point-wise variance of the model to efficiently estimate the uncertainty from an ensemble, which can provide insight into the trustworthiness of the model and possibly be used for active learning by intelligently steering the system into areas of high-uncertainty. 

A key ingredient for successfully applying DRL is to learn over long time-horizons. This posed a significant challenge to SINDy-RL (and Dyna-style learning more broadly) because the learned dynamics models are not guaranteed to be stable or converge---especially under the presence of control. We address this by incorporating known constraints, such as resetting the environment if a predicted state value exits a bounding box
and projecting the state-space back onto the appropriate manifold after each step. There has been substantial work constraining dictionary dynamics models with structured priors---such as conservation laws~\cite{Loiseau2017jfm}, symmetry~\cite{otto2023unified},  stability regions~\cite{kaptanoglu2021promoting}, and other forms of domain knowledge~\cite{ahmadi2020learning,bramburger2024synthesizing}---which further offer many promising avenues for practitioners to encourage agents to stably interact over long time horizons.

Due to the combinatorial scaling of the library, dictionary learning is challenging to apply directly to high-dimensional spaces. In this work, we have demonstrated that projecting onto a low-dimensional linear subspace using the SVD can be sufficient to make this tractable. After an initial preprint of this work, SINDy-RL was applied to controlling PDEs~\citep{wolf2024interpretable} by using the SINDy autoencoder framework~\cite{Champion2019pnas} to discover \textit{nonlinear} projections of the observation and action spaces onto a low-dimensional manifold. 
Discovering a coordinate system where the dynamics are smooth and globally defined may also help address the challenges that dictionary approaches face when the model discovery is piecewise or discontinuous. Partial observability of the environment also poses challenges for this framework. We have shown that sometimes there is sufficient information available to model the reward function from observations, but in practice this will not always be the case. Recent work~\cite{bakarji2023discovering} has investigated deep delay embeddings to identify governing dynamics when there is substantial missing information, which is an opportunity for future investigation.

While we have restricted our attention in this work to using SINDy for Dyna-style MBRL by exploiting a model-free DRL algorithm, it is important to note that there are further opportunities to combine SINDy with control. Instead of DRL, a gradient-free policy optimization such as evolutionary algorithms~\cite{salimans2017evolution} could completely replace a model-free DRL optimizer. Furthermore SINDy provides an analytic, differentiable representation of the dynamics, thus SINDy can act as a differentiable physics engine for control and used for directly calculating gradients of RL objectives. For example, by modeling the dynamics and value functions as dictionary models, the differentiable structure was utilized with the Hamilton-Jacobi-Bellman equations to directly calculate the optimal control of a system~\cite{farsi2020structured}. Finally, there may be a way to bypass the use of a policy network all-together by representing the policy directly as a sparse dictionary model for policy gradient optimization.

% ------------------------------------------------
\section*{Materials, Data, and Code Availability}
All experiments, with the exception of the \airfoil environment, were performed using a single-node, Linux engineering workstation consisting of a total of 40 CPUs (Intel$^\textsf{®}$ Xeon$^\textsf{®}$ Gold 6230). The \airfoil experiments used NVIDIA A100 GPUs on JUWELS Booster and JURECA at the Jülich Supercomputing Centre (JSC) / Forschungszentrum Jülich. 
Code and training configurations are publicly available in our repository:   \href{https://github.com/nzolman/sindy-rl}{https://github.com/nzolman/sindy-rl}.
%

% ------------------------------------------------
\section*{Acknowledgments}
The authors acknowledge support from the National Science Foundation AI Institute in Dynamic Systems
(grant number 2112085).  SLB acknowledges support from the Army Research Office (W911NF-19-1-0045) and the Boeing Company. NZ acknowledges support from The Aerospace Corporation. CL acknowledges support from the German Research Foundation within the Walter Benjamin fellowships LA~5508/1-1. The authors gratefully acknowledge the Gauss Centre for Supercomputing e.V. for funding this project by providing computing time on the GCS Supercomputers. Furthermore, the authors would like to thank the \hydrogym developers---especially Samuel Ahnert---for their help and input regarding the \hydrogym examples. Finally, the authors would like to acknowledge the helpful feedback from the anonymous reviewers in strengthening the paper.

    \putbib[refs]

\onecolumn
\clearpage
\renewcommand{\appendixname}{SI}

\appendixpage
\begin{appendices}
    \renewcommand{\thesection}{\arabic{section}}
    \renewcommand{\thefigure}{SI~\arabic{figure}}
    \setcounter{figure}{0}
    \renewcommand{\thetable}{SI~\arabic{table}}
    \setcounter{table}{0}
    % \begin{bibunit}[unsrt]

\section{Extended Introduction} 
\label{appendix:intro}
Much of the success of modern technology can be attributed to our ability to control dynamical systems: designing safe biomedical implants for homeostatic regulation, gimbling rocket boosters for reusable launch vehicles, operating power plants and power grids, industrial manufacturing, among many other examples. Over the past decade, advances in machine learning and optimization have rapidly accelerated our capabilities to tackle complicatxed data-driven tasks---particularly in the fields of computer vision~\cite{szeliski2022computer} and natural language processing~\cite{khurana2023natural}. Reinforcement learning (RL) is at the intersection of both machine learning and optimal control, and the core ideas of RL date back to the infancy of both fields. By interacting with an environment and receiving feedback about its performance on a task through a reward, an RL agent iteratively improves a control policy. Deep reinforcement learning (DRL), in particular, has shown promise for uncovering control policies in complex, high-dimensional spaces
    \cite {Kober2012book,dutta2018reinforcement,recht2019tour,agarwal2019reinforcement,Brunton2022book,van2016deep,wang2016dueling,qureshi2018adversarial,cheng2018fast}.
DRL has been used to achieve super-human performance in games 
   ~\cite{mnih2015human,silver2017mastering, silver2018general, berner2019dota, vinyals2019grandmaster} 
and drone racing 
   ~\cite{kaufmann2023champion}, 
to control the plasma dynamics in a tokamak fusion reactor
   ~\cite{degrave2022magnetic}, 
to discover novel drugs 
   ~\cite{popova2018deep}, 
and for many applications in fluid mechanics
    \cite {gazzola2014reinforcement,colabrese2017flow,Pivot2017aiaa,verma2018efficient,biferale2019zermelo,novati2019controlled,fan2020reinforcement,rabault2020deep,beintema2020controlling,novati2021automating, bae2022scientific}.
However, these methods rely on neural networks and typically suffer from three major drawbacks: (1) they are infeasible to train for many applications because they require millions---or even billions~\cite{vinyals2019grandmaster}---of interactions with the environment; (2) they are challenging to deploy in resource-constrained environments (such as embedded devices and micro-robotic systems) due to the size of the networks and need for specialized software; and (3) they are ``black-box'' models that lack interpretability, making them untrustworthy to operate in safety-critical systems or high-consequence environments. In this work, we seek to create interpretable and generalizable reinforcement learning methods that are also more sample efficient via sparse dictionary learning. 

\paragraph{Sample-Efficient DRL.} 
There has been significant research into reducing the amount of experience needed to train RL policies. 
Offline RL seeks to train policies on a pre-collected dataset as opposed to collecting new experience in an environment~\cite{levine2020offline}. When the dataset consists of ``expert'' trajectories, imitation learning approaches, such as behavior cloning, have been successful in approximating the expert policy using offline RL, which has been widely used for autonomous robotic and vehicle systems~\cite{hussein2017imitation,fang2019survey,le2022survey}. Experience replay is a popular method to bridge offline and online RL, where policies are continuously trained on new and past experience using replay buffers~\cite{schaul2016prioritized,rolnick2019experience}. Hindsight experience replay takes previous low-scoring trials and alters them to be high-scoring by changing the goal, thus improving learning rates~\cite{andrychowicz2017hindsight}. 
In contrast, transfer learning takes embodied knowledge from a similar, pre-trained task and uses it to initialize a new task---typically reusing many weights of a neural network and fine-tuning on the new task~\cite{zhu2023transfer}.  To increase the likelihood of transfer learning being successful, researchers have trained DRL policies to interact with many different (but similar) tasks or environments using meta-learning~\cite{wang2016learning, finn2017model}.

Surrogate environments are perhaps the most common way to reduce the number of interactions in an environment. Whether it's a simulation of a physical environment or a low-fidelity approximation of a high-fidelity model, one can greatly reduce the number of agent interactions by training almost entirely in the surrogate. For instance, researchers first trained agents in a simulation before deploying the agent to a real tokamak~\cite{degrave2022magnetic}. Because surrogates are only approximations of the ground-truth environments, researchers have also explored ``sim2real'' approaches that transfer learned experience from a simulated environment to a real environment~\cite{hofer2021sim2real}. Direct access to the dynamics can also be exploited by model-based reinforcement learning (MBRL) which can be used to calculate optimal control using the Hamilton-Jacobi-Bellman equations or localized planning~\cite{lewis2012optimal}. For example, model-predictive control (MPC) methods are frequently used in modern control to plan and executed optimal controls over finite-time horizons~\cite{grune2017nonlinear}.

However, there are many cases where a surrogate model---such as a low-fidelity simulation---does not exist and the dynamics must be learned. Dyna-style MBRL algorithms iteratively switch between learning and improving a surrogate model of the environment and training model-free policies inside the surrogate environment by generating ``imaginary'' experience~\cite{sutton1990integrated}. For physical environments, this iterative process of learning a surrogate model is particularly important because there may be regions of the environment that are impossible to access without a sophisticated controller---such as the unstable equilibrium of the inverted pendulum that starts in the stable, downward position. Deep MBRL algorithms have shown significant promise for reducing sample complexity on benchmark environments~\cite{wang2019benchmarking} by simultaneously training neural network models of the environment. For example, model-based meta-policy optimization (MB-MPO)~\cite{clavera2018model} trains an ensemble of neural network models for the dynamics that can be used to learn policies capable of quickly adapting to different members of the ensemble. Although neural network models have recently become popular and are gaining wide adoption over traditional modeling methods, they are still overparameterized, data-inefficient, and uninterpretable. In this work, we examine MBRL using lightweight, sparse dictionary models~\cite{brunton2016discovering} that are fast to train on limited data, convenient for estimating uncertainty, and provide an interpretable symbolic representation by construction.

\paragraph{Sparse Dictionary Learning.} Sparse dictionary learning is a type of symbolic regression that learns a representation of a function as a sparse linear combination of pre-chosen candidate dictionary (or ``library") functions.  The dictionary functions, such as polynomials or trigonometric functions, are chosen based on domain expertise and are evaluated on input data. 
Sparse learning has become a popular method for discovering dynamics, as in the sparse identification of nonlinear dynamics (SINDy) algorithm~\cite{brunton2016discovering}, which models the vector field of a dynamical system as a sparse linear combination of dictionary terms. The sparse symbolic representation of the model lends itself naturally to interpretation and model analysis---especially for physical systems where the dictionary terms have physical significance~
\cite{Sorokina2016oe,addison2017dynamics,Dam2017pf,zhang2019convergence,boninsegna2018sparse,Thaler2019jcp,zhang2019learning}. 
SINDy has been extensively used to develop reduced-order models~\cite{Loiseau2017jfm,Loiseau2018jfm,callaham2021nonlinear,guan2020sparse,Loiseau2020tcfd,Deng2020JFM,deng2021galerkin,Callaham2022jfm,Callaham2022scienceadvances} and turbulence closures~\cite{zanna2020data,schmelzer2020discovery,beetham2020formulating,beetham2021sparse} in fluid mechanics.
These methods have been extended to PDEs~\cite{Rudy2017sciadv}, simultaneously discovering coordinates from high-dimensional representations~\cite{Champion2019pnas}, and promoting stability~\cite{kaptanoglu2021promoting}.
Moreover there has been significant progress in improving SINDy techniques for the high-noise, low-data limit~\cite{schaeffer2017sparse, reinbold2020using, messenger2021weak, messenger2021weakGalerikin, both2021deepmod, forootani2023robust}. In particular, constructing an ensemble of SINDy models (E-SINDy) has been shown to provide an efficient and robust way to learn models from limited and noisy data~\cite{fasel2022ensemble} with accurately  uncertainty quantification (UQ)~\cite{niven2020bayesian,gao2023convergence}. In addition to the inherent interpretability of a sparse symbolic model, quantifying uncertainty provides additional trust in the algorithm and is useful for sensitivity analyses. 

Importantly, SINDy has been extended to systems with actuation and control~\cite{brunton2016sindyc,kaiser2018sparse} and used for designing model predictive control laws~\cite{kaiser2018sparse, fasel2021sindy, lore2023time}. SINDy methods are incredibly efficient---both for model creation and deployment---making them promising for both online learning and control in resource-constrained operations. In particular, SINDy has been used for online simultaneous dynamics discovery and optimal control design~\cite{farsi2020structured, farsi2021structured}.  Finally, recent work has suggested leveraging a single SINDy model to accelerate DRL, and has demonstrated its use on simple DRL benchmarks~\cite{arora2022model}. In this work, we generalize this DRL framework to include ensembles of dictionary models to accelerate learning in the low-data limit and quantify uncertainty. 

% ----------------------------------------------
\section{Extended Background} 
\label{app:background}
In this work, we consider control systems that evolve according to a governing differential equation 
    $$\frac{d}{dt}{\vx} = f(\vx,\vu)$$ 
or discrete-time update equation 
    $$\vx_{k+1} = f(\vx_k, \vu_k)$$ 
where the state $\vx \in \mathbb{R}^m$ and control input $\vu \in \mathbb{R}^l$ are continuously defined variables.  However, reinforcement learning is a more general framework~\cite{Sutton1998book,Brunton2022book} and can be applied to a wider class of systems, such as board games with discrete action and/or observation spaces.

% ------------------------------------------------
% ------------------------------------------------
% Reinforcement Learning
% ------------------------------------------------
% ------------------------------------------------
\subsection{Reinforcement Learning} \label{background:intro-RL}
\begin{figure*}[t]
    \centering
    \includegraphics[width=0.95\textwidth]{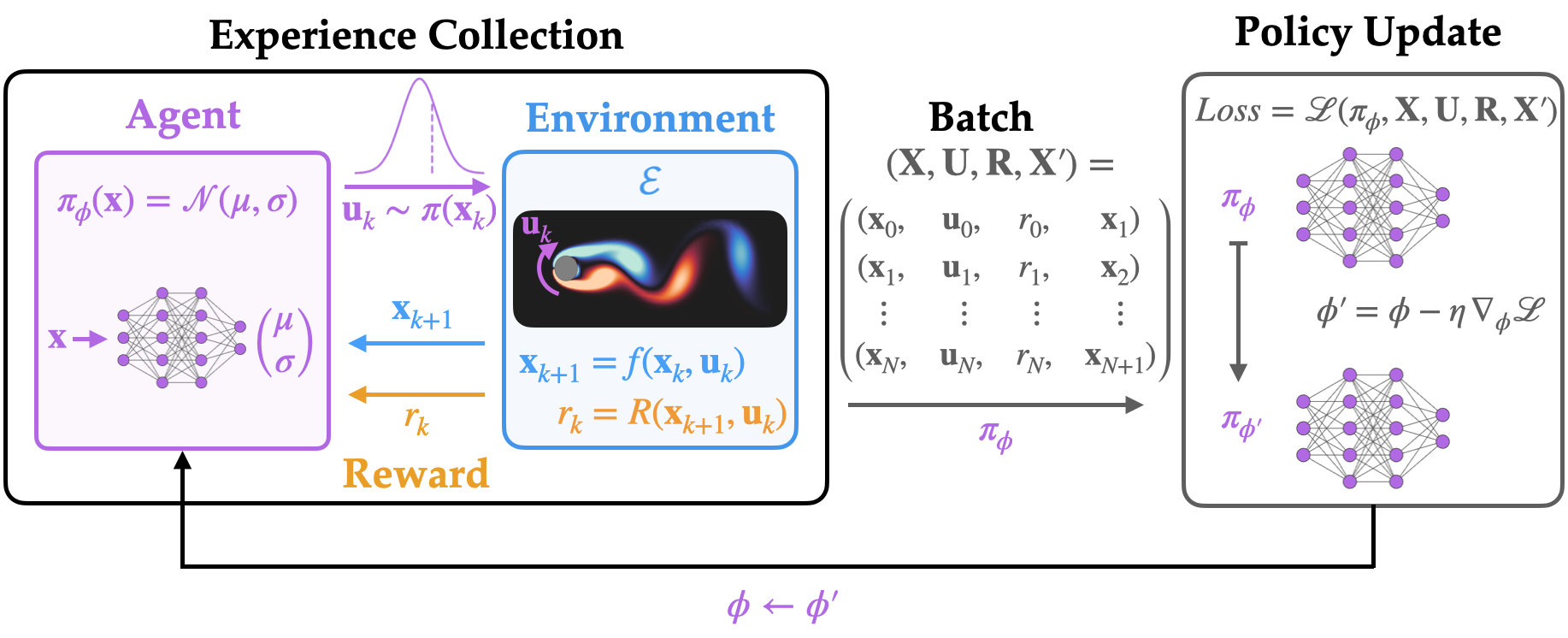}
    \caption{An overview of deep reinforcement learning. \textit{Left}: An agent collects experience by sampling an action from the policy distribution (parameterized by a deep policy network) which is executed in the environment. The environment provides a new state and reward based on the performance. \textit{Right}: A batch of experience, collected in (state, action, reward, new state) sequences, is fed to the policy optimization, where the loss is computed. The neural network parameters $\policyParams$ are updated using a gradient-based optimization.
    } 
    \label{fig:drl}
\end{figure*}

Reinforcement learning comprises a family of methods where an agent interacts in an environment, $\env$, through sequences of different states, $\RLstate_n$, by taking actions $\RLact_n$
    \footnote{
    It has become popular to refer to actions as $a_n$ and states as $s_n$ in the reinforcement literature; however, we use $u_n$ and $x_n$ to better align with the control and dynamics communities.
    }. 
The agent interacts in the environment by sampling an action from a \textit{policy} at each state  $\RLact_n \sim \pi(\RLstate_n)$ and executing it in the environment, producing a new state $\RLstate_{n+1}$ and reward $\RLrew_n$---a signal indicating how well the agent performed on that step. We define the value function, $V_\RLpolicy$ to be the expected future return for taking actions from the policy $\RLpolicy$:
\begin{align}
V_\RLpolicy(\RLstate) 
    = \mathbb{E}\left(
        \sum_{k=0}^\infty \gamma^k \RLrew_k
        \bigg \vert \RLstate_0 = \RLstate
        \right)
\end{align}
 where $0 < \gamma \leq 1$ is the \textit{discount rate} that weights earlier rewards to be worth more than later ones, and the expectation is taken when either the policy or environment are stochastic. RL methods seek to find a policy that maximizes the value function, i.e.
\begin{align}
    \RLpolicy^*(\RLstate) = \argmax_\RLpolicy V_\RLpolicy(\RLstate).
\end{align}
This can become a very challenging optimization problem, especially in the case of high-dimensional state-spaces, continuous action spaces, and nonlinear dynamics. Deep reinforcement learning (DRL) has made significant progress in addressing these problems by directly parameterizing functions as deep neural networks, such as the policy $\RLpolicy(\RLstate) \approx \RLpolicy_\policyParams(\RLstate)$, and training on collected experiences. Figure \ref{fig:drl} shows a simplified schematic of DRL training and agent interaction.

\textit{Model-free} DRL methods optimize control policies in an environment without any explicit model of how the environment will transition between states, and have gained popularity because of the ease of application and flexibility of the framework to a given problem. Model-free DRL approaches, however, generally require an enormous number of training experience. This in contrast to \textit{model-based} reinforcement learning (MBRL) methods which have shown to be substantially more sample-efficient by exploiting a model of the environment~\cite{wang2019benchmarking}. Classical model-based approaches, such as model-predictive control (MPC), exploit a model to plan optimal trajectories over a finite time-horizon. More generally, shooting algorithms~\cite{rao2009survey} are popular approaches for solving these optimization problems, and neural networks have become widely adopted for representing complicated environment dynamics~\cite{nagabandi2018neural, chua2018deep}. Other MBRL approaches optimize $V_\RLpolicy$ directly by calculating the gradients analytically with respect to the policy actions or exploiting modern auto-differentiation software to back-propagate gradients through a parameterized policy~\cite{deisenroth2011pilco,tassa2012synthesis, heess2015learning, levine2016end}. 
Finally, in order to exploit the success of model-free approaches while reducing the total number of interactions in a ground-truth environment, Dyna-style RL methods learn a surrogate model of the environment, such as linear models~\cite{sutton2012dyna}, neural networks~\cite{clavera2018model}, 
dynamic mode decomposition or Koopman models~\cite{retchin2023koopman,KoopmanRL}, and train agents to interact in this approximation of the environment.

% ------------------------------------------------
% ------------------------------------------------
% Sparse Dictionary Learning
% ------------------------------------------------
% ------------------------------------------------
\subsection{Sparse Dictionary Learning} \label{background:intro-sparse}

Sparse dictionary learning methods are a form of symbolic regression used to uncover interpretable representations of nonlinear functions from available data. Many symbolic regression algorithms utilize evolutionary or genetic algorithms to generate functions as the combinations of some candidate library functions through elementary mathematical operations (e.g. addition, multiplication, and composition)~\cite{Bongard2007pnas,schmidt2009distilling,cranmer2023interpretable}. These methods can be very expressive; however, the search space is typically vast, and since they rely on gradient-free algorithms that propose and evaluate many generations of functions, they can be prohibitively computationally expensive. In contrast, dictionary learning algorithms approximate a function purely as a linear combination of candidate ``dictionary'' functions and can be efficiently fit using traditional linear regression methods. 

The sparse identification of nonlinear dynamics (SINDy) algorithm~\cite{brunton2016discovering} uses sparse dictionary learning to discover the governing equations for dynamical systems. 
We first present a general formulation of sparse dictionary learning and then make the connection to the different dynamics discovery formulations.

\paragraph{Dictionary Learning.}
Suppose we have collected $N$ labeled data points
$
    (\vx_1, \vy_1), \dots (\vx_N, \vy_N)
$
, 
$
    \vx_k \in \mathbb{R}^m
$, 
$
    \vy_k \in \mathbb{R}^n
$
and would like to learn a function $\vy = f(\vx)$ relating these variables. To create a dictionary  model, we first compile a candidate set of $d$ dictionary (or ``library'') functions (such as polynomials, sines, cosines, etc.): 
$
    \vTheta (\vx) = (\theta_{1} (\vx), \theta_2 (\vx), \dots \theta_{d} (\vx) )
$. We assemble data  and label matrices, $\vX$ and $\vY$ respectively 
\begin{align*}
    \vX 
        &= [\vx_1, \vx_2, \dots, \vx_N]^T \in \mathbb{R}^{N \times m}  
    \\
    \vY      
        &= [\vy_1, \vy_2, \dots, \vy_N]^T \in \mathbb{R}^{N \times n},
\end{align*}
and evaluate the dictionary on the collected data $\vX$
$$
\qquad \vTheta(\vX)
                = [\theta_1 (\vX), \theta_2 (\vX), \dots, \theta_d (\vX)] \in \mathbb{R}^{N \times d}
$$
to form the dictionary model: 
\begin{equation}\label{eq:sparse-model}
    \vY = \vTheta(\vX) \vXi
\end{equation}
where $\vXi \in \mathbb{R}^{d\times n}$ are the coefficients to be fit. 
Sparse dictionary learning assumes that the desired function can be well approximated by a small subset of terms in the library, i.e. $\vXi$ is a sparse matrix, as in the LASSO from statistics~\cite{Tibshirani1996lasso}. 
For dynamics discovery, where $\vy=\dot{\vx}$, this is physically motivated by the idea of parsimony---the governing equations for most physical systems have relatively few terms. To achieve this parsimony, a sparse optimization problem is formulated:
\begin{equation} \label{app:eq:sparse-opt}
\vXi = \argmin_{\hat\vXi} 
    \Vert \vY -  \vTheta(\mathbf{X}) \hat{\vXi}\Vert_F^2 + \mathcal R(\hat{\vXi})
\end{equation}
where $||\cdot||_F$ is the Frobenius norm and  $\mathcal R(\vXi)$ is a sparsity-promoting regularization. . While the $L^0$ ``counting norm''
$\mathcal R(\vXi) = \Vert \vXi \Vert_0$ is the most appropriate to ensure sparsity, it is non-convex, non-continuous, and generally difficult to optimize when $d \gg 1$. 
For dynamics learning, sequentially thresholded least squares (STLS) has proven particularly effective for promoting sparsity by iteratively fitting the model and removing terms from the library whose coefficients are less than a certain magnitude~\cite{brunton2016discovering,zhang2019convergence}. In our work, we consider a STLS with $L^2$ regularization (STLRidge) as in~\cite{Rudy2017sciadv}.

The SINDy algorithm is a particular case of the above framework when the data $\vX$ are sampled points of the state space at a given time $\vx_k = \vx(t_k)$ and the labels $\vY$ are either the time derivatives $\dot{\vX}$ or the next state $\vX_{k+1}$ in the discrete setting; i.e. $\vy_k = \dot{\vx}(t_k)$ or $\vy_k = \vx(t_{k+1})$ respectively. These, and other dynamics discovery algorithms, are summarized with our notation in Table \ref{tab:formualations}:
the SINDy-C algorithm~\cite{brunton2016sindyc} incorporates control by augmenting the state measurements with the corresponding control inputs, $\dot{\vX} = \vTheta(\vX, \vU)\vXi$. Similarly, for discovering PDEs of the form $\psi_t = \mathcal{N}(\psi, \psi_x, \psi_{xx}, \dots)$, we can choose the data to be flattened snapshots of the solution field and any spatial derivatives, $\vX = [\vPSI, \vPSI_x, \vPSI_{xx} \dots]$ and labels $\vY = \vPSI_t := \partial \vPSI / \partial t$.

% ---------------
% E-SINDy
% ---------------
\paragraph{Ensemble Learning.} 
Ensemble-SINDy (E-SINDy)~\cite{fasel2022ensemble} introduced a way to build an ensemble of $N_e$ SINDy models by bootstrapping data samples to create new datasets $ \vX^{(k)},  \dot{\vX}^{(k)}$ and dictionaries $\vTheta^{(k)}$, $k = 1, 2, \dots N_e$ by dropping out some fraction of terms from the library. By choosing the median coefficients 
\begin{equation}\label{eq:median}
    \vXi^*_{ij} = \text{Median}_k \left[ \vXi^{(k)}_{ij} \right]    
\end{equation}
the resulting SINDy model was more robust to noise, particularly in the low-data limit. In this work, we consider a generalization of the E-SINDy ensembling methods for arbitrary labeled datasets $\vX,  \vY$ by considering ensembles: 
$$
    \vY^{(k)} 
        = \vTheta^{(k)}  (\vX^{(k)}) \vXi^{(k)}
$$
and taking the median coefficients $\vXi^*$ as in Equation \ref{eq:median}.

% ---------------
% UQ
% ---------------
\paragraph{Uncertainty Quantification.}
The linearity of dictionary models not only provides a framework for interpretable models, but it also enables us to perform uncertainty quantification. 
For the scalar-valued dictionary model $y(\vx) = \vTheta(\vx)\vxi$ and $\vxi \in \mathbb{R}^d$, it is simple to calculate the pointwise model variance under the assumption that the coefficients $\xi$ are being drawn from some arbitrary distribution. Because $\vTheta(\vx)$ are independent from $\vxi$, we can use the linearity of the dictionary model to write:
\begin{equation}\label{eq:var}
    \text{Var}_\vxi[y(\vx) | \vx] 
        = \vTheta(\vx) \text{Cov}(\vxi) \vTheta(\vx)^T
\end{equation}
where $\text{Cov}(\vxi)$ is the covariance matrix of the coefficients $\vxi$. For a dictionary model of several variables  $\vy(\vx) = (y_1(\vx),  \dots y_n(\vx))$ where $y_i(\vx) = \vTheta(\vx)\vxi_i$, the total variance of $\vy(\vx)$ is given by the trace of its covariance matrix. Explicitly: 
\begin{equation}\label{app:eq:multi-var}
    \text{Tr} \left(\text{Var}_\vXi[\vy(\vx) | \vx] \right)
        = \sum_{i=1}^n \vTheta(\vx)\text{Cov}(\vxi_i) \vTheta(\vx)^T.
\end{equation}
An ensemble of sparse models acts as an approximation of the coefficient distribution, and thus the pointwise variance can be rapidly computed just by estimating the covariance matrices of the fitted ensemble of coefficients.

% ---------------
% Control
% ---------------
\paragraph{Dictionary Learning for Control.}
By using SINDy-C~\cite{brunton2016sindyc}, it is possible to exploit the dynamics for the purposes of control. Model predictive control (MPC) is a popular family of methods for optimal control that repeatedly optimizes control over a finite time horizon. This task can be incredibly expensive because it requires many forward evaluations of the dynamics in order to determine the optimal next step, and thus becomes challenging to deploy in real-time systems. In~\cite{kaiser2018sparse}, SINDy was used to discover a lightweight dictionary model of the dynamics from limited data and used as a rapid surrogate model for MPC. This has been applied for the purposes of tokamak control~\cite{lore2023time}. SINDy has also been used to accelerate model-free DRL training using a surrogate SINDy model~\cite{arora2022model}; by leveraging random data to train a SINDy model, the authors show that it is sometimes possible to train a DRL agent without any additional data from the original environment. 

There have also been approaches for online learning. By modeling unknown dynamics using SINDy and the unknown optimal value function as a quadratic form of the dictionary functions~\cite{farsi2020structured, farsi2021structured}, it is possible to use the Hamilton-Jacobi-Bellman equations to derive an expression for optimal control inputs. Then, by iteratively deploying control into the system, better dynamics models and value functions are learned. 
Finally, E-SINDy also explored the use of ensemble statistics for active learning: by identifying regions of the state space with high variance, the accuracy of the dynamics model can be efficiently improved by sampling trajectories from initial conditions from those regions.

\begin{table*}
    \centering
    \begin{tabular}{|c|c|c|c|}
        \hline
         Algorithm &  Discovery Task & $\vX$ & $\vY$\\
         \hline
            SINDy   
            & $\dot{x} = f(x)$  
            & $\vx_k = \vx(t_k)$ 
            &  $\vy_k = \dot\vx(t_k)$ 
         \\
            Discrete SINDy     
            & $x_{n+1} = f(x_n)$ 
            &  $\vx_k = \vx(t_k)$ 
            &  $\vy_k = \vx(t_{k+1})$ 
         \\
            Discrete SINDy-C            
            & $x_{n+1} = f(x_n, u_n)$  
            &  $\vx_k = (\vx(t_k), \vu(t_k))$
            &  $\vy_k = \vx(t_{k+1})$ 
        \\
            PDE-FIND
            & $\psi_t = \mathcal{N}(\psi, \psi_x, \dots, x)$
            &  $\vx_k = (\psi, \psi_x,  \dots x)$
            &  $\vy_k = \psi_t(x, t_k)$
        \\
        \hline
    \end{tabular}
    \caption{Different formulations of data-driven dynamics discovery with the sparse dictionary learning model $\vY = \vTheta(\vX) \vXi$.}
    \label{tab:formualations}
\end{table*}
\clearpage

% ----------------------------------------------
\section{Environments} 
\label{appendix:envs}
\begin{figure*}[t]
    \centering
    \includegraphics[width=\textwidth]{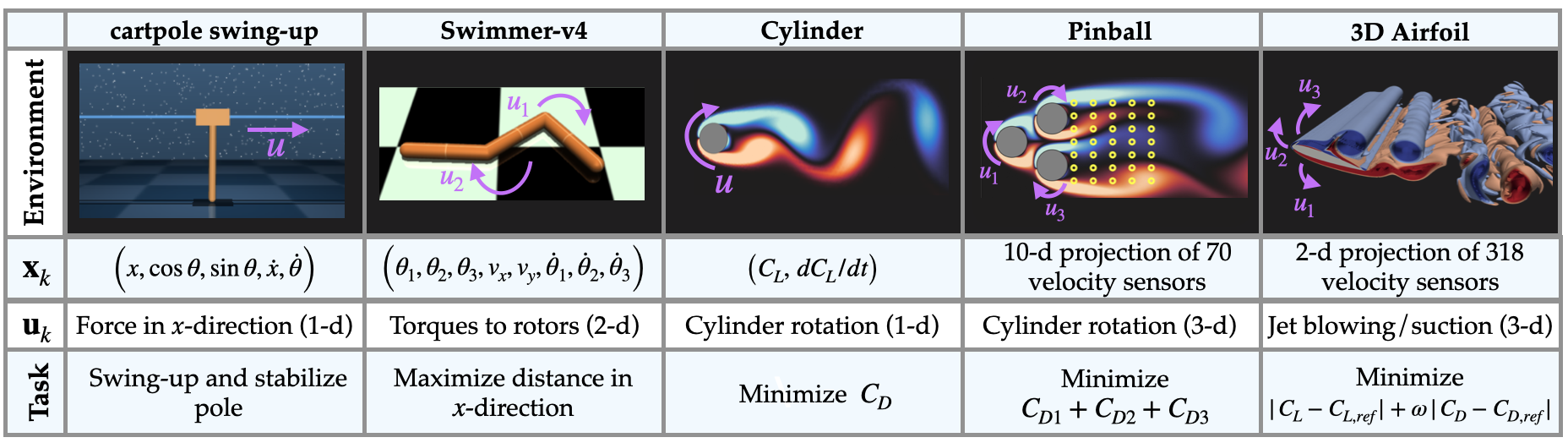}
    \cprotect \caption{The environments considered in this work, their respective observation spaces $\vx_k$, and their action spaces $\mathbf{u}_k$. 
\textit{Left:} The \swingup environment objective is to balance the pole on the cart in the upright (unstable) position starting from the downward (stable) position at rest by actuating the cart left and right.
\textit{Center Left:} The \swimmer environment objective is to travel as far in the $x$-direction as possible in fixed time horizon starting from rest by actuating two rotors. 
\textit{Center:} The Cylinder environment objective is to minimize the drag force on the cylinder surface by rotating the cylinder. 

\textit{Center Right:} The Pinball environment objective is to minimize the net drag force on the cylinder system by independently rotating the three cylinders. 
\textit{Right:} The \airfoil environment objective is to minimize fluctuations of the lift and drag forces from refernence values.
    }
     \label{fig:envs}   
\end{figure*}

As depicted in Figure \ref{fig:envs}, we evaluate our methods among several different environments: the \dmcontrol \swingup environment, the \gym \footnote{formerly known as ``OpenAI \texttt{gym} ''} \swimmer environment, the \hydrogym Cylinder, the \hydrogym Pinball environment, and the \hydrogym-GPU 3D Airfoil  environment. Below, we describe each environment in more detail.

% -----------
% Swingup
% -----------
\subsection{Swing-up.}
The \dmcontrol \swingup environment uses continuous actions to swing up and balance a pendulum on a cart, starting from the downward position, by actuating the cart forward and backward along a horizontal rail. This is a more challenging controls task than the \gym ``CartPole'' (discrete actions) and ``Inverted Pendulum'' (continuous actions) benchmark environments which start near the upright positions---thus, \swingup mimics the more practical setup where the system starts in a stable, but undesirable configuration and requires sophisticated control to reach a region of interest (e.g. an unstable fixed point). As shown in Figure \ref{fig:envs}, the states are given by the cart position $x$, pole angle $\theta$, and their respective derivatives $\dot x, \dot \theta$, 
where the angle $\theta$ has been embedded on the circle $S^1$ to ensure continuous dynamics. The presence of the cart makes the dynamics significantly more complicated than a fixed pendulum. Without the cart, the pendulum dynamics can be recovered by dictionary learning with a library consisting of trigonometric and polynomial candidates. However, the presence of the cart leads to dynamics of the form $\mathbf{P}(\vx)\dot \vx  = \mathbf{Q}(\vx)$; thus $\dot \vx$ cannot be easily expressed in terms of linear combination of elementary functions. While dynamics of this form have been addressed with dictionary learning in SINDy-PI~\cite{kaheman2020sindy} (by considering dictionary candidates for both $\mathbf{P}(\vx)$ and $\mathbf{Q}(\vx)$ and solving a more sophisticated optimization problem), we do not utilize that approach in this work. Instead, we demonstrate that even when the dynamics cannot be fully represented by our library, the imperfect representation is enough to achieve the controls task with SINDy-RL. As described in~\cite{tunyasuvunakool2020dm_control}, the reward signal is given by a product of different  functions bounded between 0 and +1
$$
r = 
\left( \frac{1 + \cos \theta}{2}\right)
\left( \frac{1 + \Psi_1(x)}{2}\right)
\\
\left( \frac{4 + \Psi_2(u)}{5}\right)
\left( \frac{1 + \Psi_3(\dot \theta )}{2}\right)
,$$
where each $\Psi_i$ are ``tolerance'' functions such that they are $+1$ near the desired behavior and decay to $0$ away from it---providing a rich, but challenging learning signal. The combination of nonlinear dynamics/rewards, frequent interactions (100Hz), and long trajectory lengths (1000 interactions) makes this environment very challenging to explore. 

% -----------
% Swimmer-v4
% -----------
%
\subsection{Swimmer-v4}
The \gym Mujoco environment \swimmer consists of a 3-segment swimmer agent immersed in a viscous fluid as depicted in Figure \ref{fig:envs}. The 8-dimensional observation space is comprised of the body orientation $\theta_1$, the joint angles of the two rotors $\theta_2$ and $\theta_3$, their respective angular velocities $\omega_k$, and the $x$- and $y$- velocities of the front segment's tip.
By providing torque to the rotors on its two joints, the agent is tasked with moving as far as it can in the $x$-direction in a given number of time steps. To encourage this objective, the agent is rewarded with the body-centered $x$-velocity. However, this quantity is not provided as part of the agent's observation space. In order to use MBRL, a surrogate reward must be provided. It has been well-documented that---even with access to the exact rewards---solving the \swimmer environment is extremely challenging because of the placement of the velocity sensors; researchers have even created their own modified versions of the environment when performing benchmarks~\cite{wang2019benchmarking}.
It has been established across different benchmarks that the PPO algorithm struggles to achieve any meaningful performance despite its popularity and promising performance on other continuous controls tasks~\cite{weng2022tianshou, huang2022cleanrl}. Others have suggested that different hyper-parameters are needed~\cite{franceschetti2022making}.

% -----------
% Cylinder
% -----------
%
\subsection{Cylinder}
The \hydrogym Cylinder environment consists of the canonical two-dimensional unsteady fluid flow past a cylinder. This example is often used as a first approximation for many engineering systems that involve a fluid wake behind a blunt object, and it has become a popular test case for fluid flow control~\cite{rabault2020deep, fan2020reinforcement, varela2022deep, wang2023dynamic}. By actuating the cylinder via rotations about its center, the flow around the surface of the cylinder changes and affects the forces it experiences. The objective is to minimize the drag, $C_D$, on the cylinder. The reward function is approximately $r = -\Delta t C_D $, where $\Delta t = 0.1$ is the is amount of time between the agent taking actions. The dynamics of the flow are governed by the Navier-Stokes equations, and thus the full state of the system is infinite dimensional. To simulate this, \hydrogym uses a finite element solver with thousands of degrees of freedom and, depending on the solver\footnote{In the version of \hydrogym used to produce the Cylinder results, an incremental pressure correction scheme was used, causing step sizes to take longer. An updated version of the code uses a semi-implicit BDF solver, reducing the cost of a computational step.}
configuration, it may take several seconds to compute the effect of a single action, making it exceptionally time-consuming to use model-free RL. In many flow-control problems, engineers do not have access to all the desired sensing information. To emulate the under-observed setting in this work, we only provide the agent with observations of the lift coefficient and an estimate of its time derivative: $(C_L, \dot C_L)$; thus, in order to use MBRL, we must learn a representation of the reward function from our observations. From a theoretical perspective, these variables are motivated from the uncontrolled setting, where previous work has shown that flow can be described by measurements evolving on the $(C_L, \dot C_L)$ state space~\cite{Loiseau2018jfm}.

Because the $C_L$ can be extremely sensitive to actuation, we first apply a low-pass filter 
$$\hat{C}_{L}(t_{k+1}) = \hat{C}_{L}(t_k) + \frac{\Delta t}{\tau} (C_{L}(t_k) - \hat{C}_{L}(t_k)$$ 
where $\tau = 0.556 = T/10$, one-tenth of the vortex shedding frequency. We estimate $\dot C_L$ with an Euler estimate $\frac{\hat{C}_L(t_{k+1}) - \hat{C}_L(t_k)}{\Delta t}$. To achieve this, the solver timestep was set to $ \Delta t =0.01$ seconds with the agent interacting every $0.1$ seconds (10Hz) and a simple first difference approximation was used.

Instead of simply setting the angular velocity of cylinder as performed in other fluid control simulators, the angular velocity of the cylinders are governed by the equation
$$
\dot \omega = (\tau u - \omega) \frac{1}{\tau I_{CM}},
$$
where $u$ is the control input (torque),  $\tau$ is the same time constant, and $I_{CM} = 0.0115846$ is the moment of inertia. For fixed control, the angular velocity will trend to $\tau u$. These dynamics act as a dampening caused by internal friction or external viscous dampening; thus providing a more realistic and complicated control scenario. The variable $\omega$ also acts as a hidden state variable since it is not given to the agent during training. In addition to the default actuation dampening in the environment provided every $\Delta t$, we provide a first-order hold in the intermediate timesteps between agent interactions in order to provide a smoother control signal into the environment. 

For the initial off-policy data collection, we simulate the environment until the flow instability leads to periodic vortex shedding. Then we use 5 trajectories driven with sinusoidal control for $50$ seconds followed by $10$ seconds of no control. The $k$-th trajectory's sinusoid was parametrically defined by $u(t) = A \sin(\frac{2 \pi t}{k T} - \phi) + B$, where $T=5.56$ is the vortex shedding period of the single cylinder at Re=100---i.e. the period of each sine wave is an integer multiple of the shedding period. The parameters $A, \phi, B$ were sampled uniformly from the box: $[0.25, 1.0] \times [0, \pi] \times [-\frac{\pi}{2}+1, \frac{\pi}{2}-1]$ and chosen so that the maximal control of the environment $(\pm \pi/2)$ should never be exceeded. 

% -----------
% Pinball
% -----------

\begin{figure*}[h]
    \centering
    \includegraphics[width=0.9\textwidth]{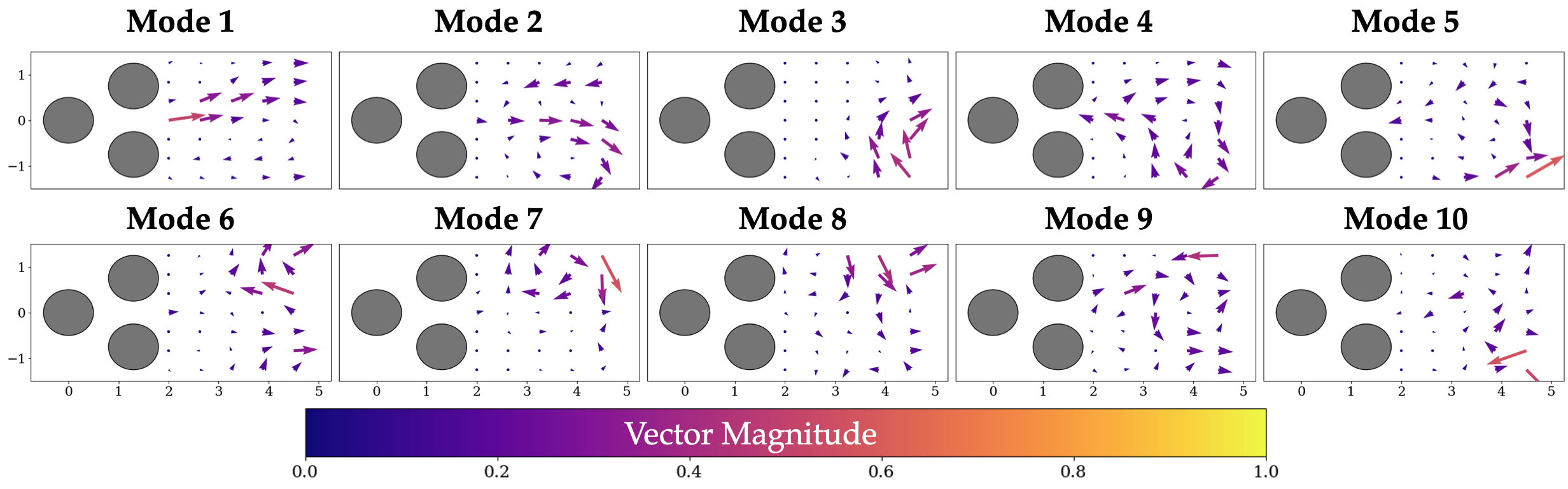}
    \cprotect \caption{\textbf{Pinball SVD Modes.} An illustration of the SVD modes obtained from snapshots at $Re=100$ plotted as a vector-field at the sensor locations. Color indicates the magnitudes of the components at the given sensor. 
    }
     \label{fig:app-pin-pod}   
\end{figure*}

\subsection{Pinball}
The ``fluidic pinball'' introduced Noack and Moryzński \cite{noack2017fluidic} consists of a 2-dimensional flow past three cylinders arranged in an equilateral triangle, as shown in Figure \ref{fig:envs}. As the Reynold's number, $Re$, increases, the uncontrolled system dynamics experience a series of bifurcations. In particular, at $Re=100$, the system exhibits quasi-periodic vortex shedding and at $Re\approx115$, the dynamics become chaotic ~\cite{deng2020low}. The objective of the environment is to minimize the total net drag $C_{D1} + C_{D2} + C_{D3}$ by independently rotating the three cylinders. In this version of the environment, observations are generated through a $7 \times 5$ grid of $x-$ and $y-$ velocity sensors in the wake as pictured in Figure \ref{fig:envs}, for a total observation size of $70$. 

The initial condition was generated by solving for the steady state at $Re=100$ and taking a sequence of 100 random actions to excite the instability in the flow and was simulated for an additional 200 seconds without control with $dt = 0.01$ to reach the quasi-steady shedding. During this $200s$ period, snapshots of the observations were taken and used to obtain a low-dimensional SVD basis. To reduce the dimension of the observation, observations were projected onto the leading 10 SVD modes. Figure \ref{fig:app-pin-pod} illustrates the modes as a vector field at the sensor locations. Many of the modes appear to be continuous vector fields highlighting vortex features in the wake. 

Whereas the Cylinder was actuated at $dt=0.1$, the Pinball is actuated at $dt=0.01$---interacting with the environment 10 times as often. Just like the Cylinder, the actuation mechanism is applying torque to the cylinders until it saturates to some maximal angular velocity. In the version of \hydrogym used for these results\footnote{commit hash 0a19230a24a9b671893923c88a2da34899d79032}, though, the angular velocity is simplified to

$$
    \dot \omega  = \tau (u-\omega)
$$

\noindent where $\tau = 0.05$.

Just as the Cylinder, the initial off-policy data collection was produced from this uncontrolled initial condition and simulating forward with independently generated sinusoidal control on each of the three cylinders for $30$ seconds followed by $5$ seconds of no control (at $dt = 0.01$). The $k$-th trajectory's sinusoid was parametrically defined by $u(t) = A \sin(\frac{2 \pi t}{k T} - \phi) + B$, where $T=5.56$ is the vortex shedding period of the single cylinder at $Re=100$---i.e. the period of each sine wave is an integer multiple of the shedding period. The parameters $A, \phi, B$ were sampled uniformly from the box: $[0.25, 0.75] \times [-\pi, \pi] \times [-0.25, 0.25]$ and chosen so that the maximal control of the environment $\pm 1$ should never be exceeded.

\bigskip

\subsection{3D Airfoil}
\subsubsection{Numerical Setup}
The direct numerical simulations are conducted using a lattice Boltzmann method (LBM) with a cumulant-based collision operator implemented in the m-AIA (multiphysics - Aerodynamisches Institut Aachen) solver framework \cite{lintermann2020zonal}, which provides highly efficient and scalable computational capabilities. The framework benefits from hybrid parallelization based on MPI and shared memory models, enabling hardware-agnostic implementation on both CPU and GPU architectures with favorable strong and weak scaling on modern HPC systems \cite{HydroGym-GPU}. The hierarchical unstructured Cartesian grids are generated using a massively parallel grid generator \cite{lintermann2014massively} integrated within m-AIA.

To enable efficient communication between the LB solver and reinforcement learning agents, m-AIA's MPI interface is extended using Multiple Program Multiple Data (MPMD) mode. This approach facilitates complex distributed computations across heterogeneous hardware while supporting different programming languages within the same application, enabling seamless integration between the CFD environment and Python-based RL libraries \cite{HydroGym-GPU}. The MPMD interface enables frequent data exchange and coordination between different computational tasks with minimal computational overhead, supporting multi-agent and multi-environment training protocols in a straightforward fashion.

The LBM solver employs a D3Q27 velocity discretization scheme on hierarchical unstructured Cartesian grids. The lattice Boltzmann equation is formulated as:
\begin{equation}
f_i(\mathbf{x} + \boldsymbol{\xi}_i\delta t, t + \delta t) - f_i(\mathbf{x}, t) = -\omega(f_i(\mathbf{x}, t) - f_i^{eq}(\mathbf{x}, t))
\end{equation}
where $f_i$ represents the particle probability distribution functions (PPDFs) at neighboring fluid cells, $\boldsymbol{\xi}_i = (\xi_{i1}, \xi_{i2}, \xi_{i3})^T$ denotes the discrete molecular velocity vector, and $\omega$ is the collision frequency \cite{he1997theory}. The macroscopic flow quantities are obtained from moments of the distribution functions:
\begin{align}
\rho(\mathbf{x},t) &= \sum_i f_i(\mathbf{x},t)\\
\mathbf{u}(\mathbf{x},t) &= \frac{1}{\rho} \sum_i \boldsymbol{\xi}_i f_i(\mathbf{x},t)
\end{align}

Following the cumulant approach of Geier et al. \cite{geier2015cumulant}, instead of relaxing the momentum distribution function towards equilibrium in momentum space, countable cumulants are relaxed in cumulant space:
\begin{equation}
c_\alpha^* = c_\alpha + \omega_\alpha[c_\alpha^{eq} - c_\alpha]
\end{equation}
where $c_\alpha^{eq}$ represents the Maxwellian equilibrium in cumulant space, and $\omega_\alpha$ denotes the relaxation frequency. The cumulants are observable quantities that are Galilean invariant and statistically independent of each other \cite{geier2015cumulant}. Following standard practice, all relaxation rates except $\omega_1$ are set to unity ($\omega_1 = \omega_{BGK}$, $\omega_\alpha = 1$ for $\alpha \neq 1$), where $\omega_{BGK}$ represents the BGK relaxation frequency.

The computational domain spans \textbf{[$32 \times$ 16 $\times 4$]} chord lengths $c=1.0$, with the NACA 0012 airfoil positioned \textbf{[8c]} from the inflow boundary. The domain extent is designed to minimize boundary reflections while maintaining computational efficiency. Non-reflecting characteristic boundary conditions are applied at the inflow and outflow boundaries to prevent pressure reflections \cite{izquierdo2008characteristic}. The airfoil surface is treated with a no-slip boundary condition implemented through an interpolated bounce-back scheme. Periodic boundary conditions are imposed in the spanwise direction to simulate a nominally two-dimensional configuration while capturing three-dimensional flow instabilities. The domain height is set to 16 chord lengths to ensure far-field boundary effects do not influence the near-airfoil flow physics.

\subsubsection{Grid Resolution and Refinement Strategy}
The simulations employ DNS where all relevant scales are directly resolved without the use of subgrid-scale models. The grid resolution is chosen to ensure that the smallest scales (Kolmogorov scales) are adequately captured, particularly in the separated flow regions downstream of the airfoil at $\alpha = 20^\circ$. This approach provides the highest fidelity representation of the complex unsteady flow physics, including vortex shedding, flow separation, and turbulent mixing processes that are critical for accurate gust-airfoil interaction modeling.

\begin{figure}[h!]
    \centering
    \includegraphics[width=0.5\linewidth]{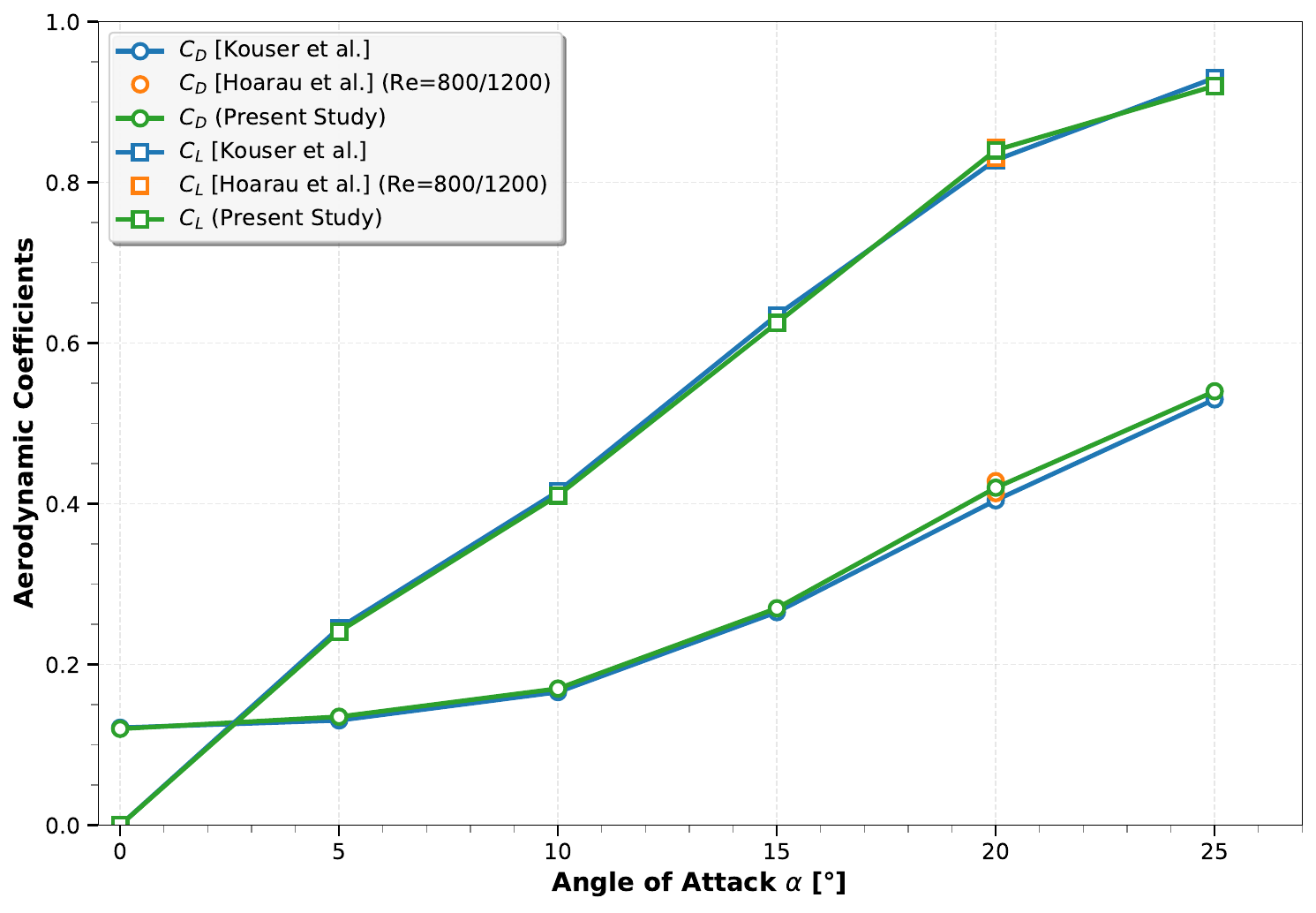}
    \caption{Validation of numerical setup against benchmark data. Comparison of time-averaged lift ($C_L$) and drag ($C_D$) coefficients for NACA 0012 airfoil across various angles of attack ($0^\circ < \alpha < 25^\circ$) at $Re = 1000$. Present study results show good agreement with literature data (Kouser et al.~\cite{kouser2021direct} / Hoarau et al.~\cite{hoarau2006first}).}
    \label{fig:naca-validation}
\end{figure}

The computational mesh employs an adaptive refinement approach with 14 refinement levels to efficiently resolve the disparate length scales present in the problem. Local grid refinement is implemented using hierarchical Cartesian grids with refinement patches positioned to capture critical flow regions. The spacing of the Cartesian grid on refinement level $r$ is given by $\Delta x_r = L/2^r$, where $\Delta x_0$ represents the bounding box of the computational domain. The finest grid spacing near the airfoil surface is $1.953125 \times 10^{-3}$ ensuring adequate resolution of the boundary layer and wake structures for DNS requirements. The numerical setup is validated against benchmark data from literature \cite{kouser2021direct,hoarau2006first} showing good agreement in time-averaged lift and drag coefficients across various angles of attack $0^\circ<\alpha<25^\circ$ (see Fig. \ref{fig:naca-validation}).

\subsubsection{Flow Conditions and Gust Implementation}

The baseline flow conditions correspond to a chord-based Reynolds number of $Re_c = 1000$ at a Mach number of $M = 0.2$. The airfoil is positioned at a fixed angle of attack of $\alpha = 20^\circ$, placing the system well into the post-stall regime where traditional linear aerodynamic models are invalid.

The transverse gust is implemented through a prescribed velocity perturbation imposed at the inflow boundary. The gust profile follows a 1-cosine approach with a gust factor $G = 2.0$, representing the ratio of peak gust velocity to freestream velocity. The gust temporal development is controlled to achieve the desired interaction dynamics with the airfoil.

The control system leverages the HydroGym-GPU platform \cite{HydroGym-GPU}, which provides a standardized interface between high-fidelity CFD simulations and state-of-the-art reinforcement learning algorithms. This platform extends the original HydroGym framework by incorporating GPU-accelerated three-dimensional flow environments with the m-AIA solver, enabling efficient training of RL agents on complex fluid flow problems with grid sizes on the order of $10^8$ cells.

Communication between the CFD environment and RL agents is facilitated through the MPMD interface, enabling real-time data exchange with minimal computational overhead \cite{HydroGym-GPU}. This architecture supports multi-agent training protocols and allows for efficient exploration of the control parameter space during the learning process. All communications between environments and agents as well as inter-environment communication are handled by the MPMD interface, while relevant inter-agent communication is realized using existing deep learning libraries.

The airfoil-gust interaction environment is formulated as a discrete-time Markov Decision Process.
Here, this discretization is inherent in the numerical simulation of fluid dynamics, which proceeds in discrete time steps. Starting from a given state (flow field), applying a control action results in a transition to a new state after a fixed interval of 640 time steps. 

\begin{figure}[t!]
    \centering
    \includegraphics[width=\linewidth]{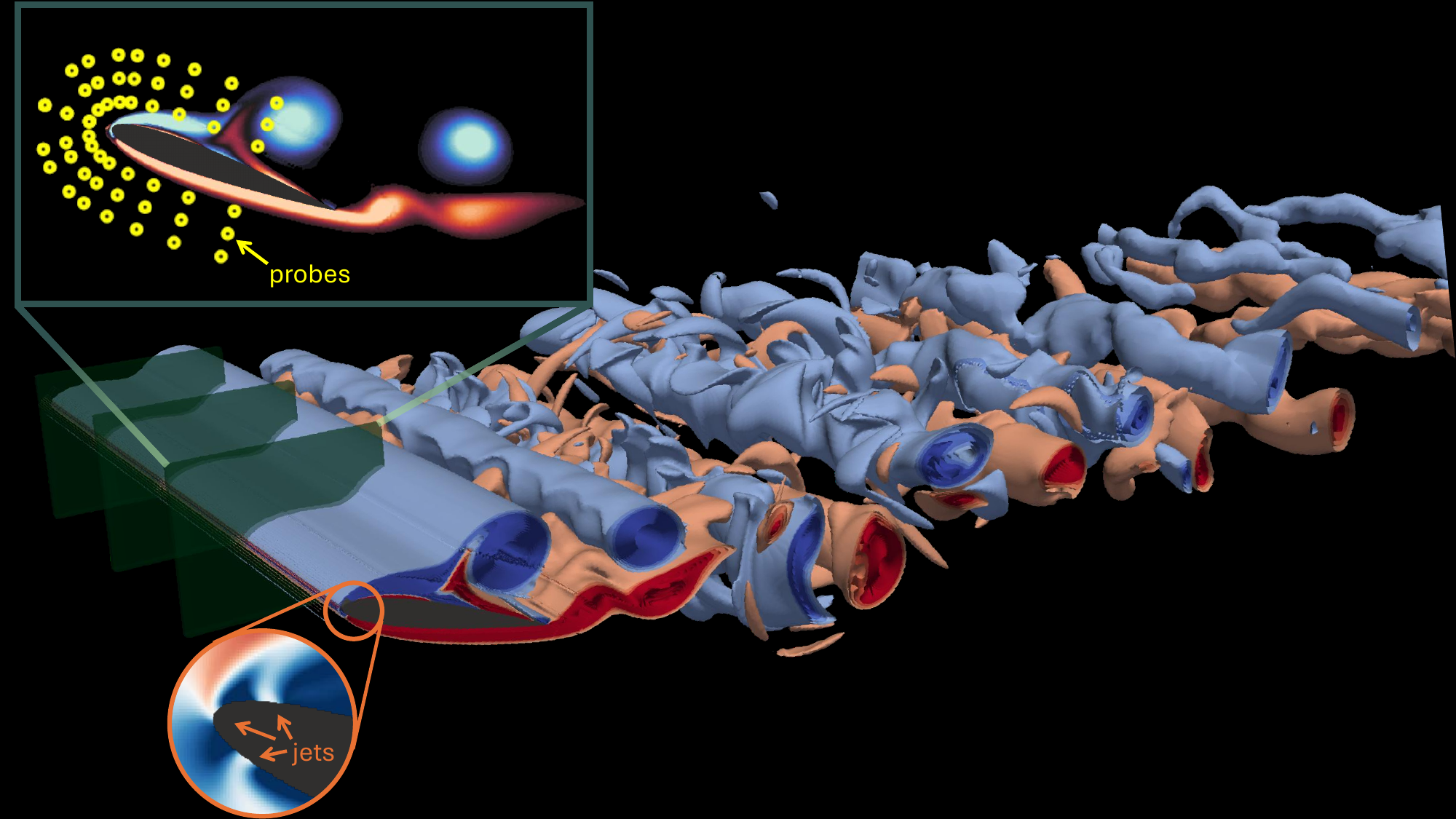}
    \caption{Computational domain setup, sensor probe distribution and actuation strategy: Three-dimensional view showing the probe distribution around the NACA 0012 airfoil with vortical structures visualized through Q-criterion isosurfaces. The yellow dots in upper zoom-in highlight sensor probe locations in a single spanwise planes. Active flow control is performed using three independent jet actuators positioned along the leading edge of the NACA 0012 airfoil, each covering $3\%$ of chord length and extending across the full span (see lower zoom-in).}
    \label{fig:naca-actuation}
\end{figure}

The state space encompasses 53 probes around the airfoil collecting horizontal and vertical velocity information across three different planes which are located at $z_p=[-1.0, 0.0, 1.0]$ (see Fig. \ref{fig:naca-actuation}). The collected probes values are normalized by the inflow velocity $U_\infty$. To facilitate interactions between the RL agent and the flow environment, three jet actuators are distributed across the leading edge of the airfoil (see Fig. \ref{fig:naca-actuation}). Each actuator can be controlled independently, covers $3\%$ of the chord length, and extends over the entire spanwise location. The action spaces are continuous and normalized in the range $[-1.0, 1.0]$. To simulate physical damping in the actuator and prevent numerical instabilities from high-frequency inputs, we model the actuators as leaky integrators. 

The reward function is formulated to minimize gust-induced force fluctuations while maintaining aerodynamic efficiency:
\begin{equation}
R(t) = -|C_L(t)-\overline{C_{L,ref}}| - \omega ~|C_D(t)-\overline{C_{D,ref}}|
\end{equation}
where $C_L(t)$ and $C_D(t)$ are the instantaneous drag and lift coefficients while $\overline{C_{L,ref}}$ and $\overline{C_{D,ref}}$ represent time-averaged coefficients of the unperturbed reference case. The weighting coefficient $\omega=0.25$ is tuned to balance the competing objectives of gust mitigation and drag reduction.

To compute the observations for both the SINDy-RL and baseline DRL experiments, the 318-dim measurements (= 2 velocities $\times$ 3 planes $\times$ 53 probes) were projected onto a 2-dim space. The space was obtained by collecting 10 episodes worth of data from random actuation of the flow. The mean sensor values were subtracted, and we project onto the leading 2 SVD modes from the resulting data matrix.

\clearpage

% ----------------------------------------------
\section{DRL Training and Benchmarking}
\label{appendix:benchmarking}
Table \ref{tab:mbrl-params} provides the hyperparameters for each environment used in the Dyna-style SINDy-RL (\appendixAlgoRef). We collect experience from the full-order environments in the form of a data store $\mathcal{D}$ consisting as a union of the off-policy data store $\mathcal{D}_{\text{off}}$ and on-policy data store $\mathcal{D}_{\text{on}}$.  In an effort to not bias the dynamics too heavily towards the learned control policy, $\mathcal{D}_{\text{on}}$ employed a queue that removed old on-policy experience as new experience was collected. Other approaches to balance the two datasets, such as data augmentation or prioritized replay buffers \cite{schaul2016prioritized} that strategically sample from the set of all collected data could also be employed, but were not considered in this work. To generate initial observations for the surrogate SINDy dynamics, we use the same sampling function for the full-order environment initial conditions. In the case of the \hydrogym environments (where full-flow field checkpoints are needed to generate the initial conditions), we simply sample elements of $\mathcal D$ to initialize an episode of surrogate experience. 

\begin{table}[h]
    \centering
    \begin{tabular}{|c|c|c|c|}
        \hline
        Environment &  $N_{\text{collect}}$ & $N_{\text{off}}$  & $n_{\text{batch}}$ \\
        \hline
         \swingup   &  1000          & 8000       & 40\\
         \swimmer   &  1000          & 12000      & 5 \\
         Cylinder   &  200           & 3000       & 25\\
         Pinball    &  1000          & 17500      & 25\\
         3D Airfoil &   250 &   10000   &   25 \\ 
         \hline
    \end{tabular}
    \caption{\appendixAlgoRef Hyperparameters}
    \label{tab:mbrl-params}
\end{table}

For all of the DRL experiments involving the \swingup, \swimmer, Cylinder, and Pinball, we utilize RLlib version 2.6.3. For the \airfoil environment, we implement PPO directly in PyTorch to avoid issues with the HPC cluster and GPU m-AIA code.

\paragraph{RLlib.} For each of the RLlib experiments, 20 different instantiations (e.g. random seed initializations) were used to characterize the variability of agents. Unless otherwise specified, we use RLlib's default parameters. For training the Baseline, Dyna-NN, and SINDy-RL algorithms, we use RLlib's \cite{pmlr-v80-liang18b} default implementation of PPO \cite{schulman2017proximal} and draw hyperparameters most similarly from CleanRL's exhaustive benchmarks \cite{huang2022cleanrl}. All hyperparameters explicitly given to RLlib to override the default are provided in  Table \ref{tab:ppo-hyper}. Just like CleanRL's benchmarks, we use annealing for the learning rate---decreasing linearly between an initial learning rate: $3.0 \times 10^{-4}$ and $3.0 \times 10^{-9}$. Each neural network policy was a fully-connected neural network with two hidden layers consisting of 64 units each and hyperbolic tangent activation functions. For the Baseline PPO, Dyna-NN, and all SINDy-RL experiments (across all environments), policy updates are applied after 4000 interactions of experience from data collected in the (surrogate or full-order) environment. 

\begin{table}[h]
    \centering
    \begin{tabular}{|c|c|c|c|c|c|c|c|c|}
        \hline
             $\gamma$ 
            &$\lambda$  
            & VF Loss Coeff 
            & PPO Clip
            & VF Clip 
            & Gradient Clip 
        \\ \hline
              0.99 % gamma
            & 0.95 % lambda
            & 0.5  % VF Coeff
            & 0.2  % PPO Clip
            & 0.2  % VF Clip
            & 0.5  % Grad Clip
        \\ \hline
    \end{tabular}
    \caption{PPO Hyperparameters used in experiments.}
    \label{tab:ppo-hyper}
\end{table}

\paragraph{Differences for 3D Airfoil}
For the \airfoil, we use a custom implementation of PPO using PyTorch for both the baseline PPO and the policy algorithm $\mathcal A$ in \appendixAlgoRef. Hyperparameters were chosen to match RLlib and those in Table \ref{tab:ppo-hyper}. Due to computational constraints, only 4 independent instantiations of a run were deployed for the \airfoil in comparison to 20 instantiations for all other environments/algorithms. However, there was little performance variability between the agents.

% ------------------------
% MBRL
% ------------------------
\subsection{MBRL Comparison---Swing-Up}
\label{appendix:benchmarking-mbrl}
For the SINDy-RL (linear and quadratic) and Dyna-NN experiments presented in 
the benchmark results, 
the cosine and sine components of the observation are projected back onto the unit circle after each interaction and the surrogate environments are reset if certain thresholds are exceeded (as described in \TheAppendix Section \ref{appendix:dynamics}). For training the Dyna-NN dynamics, we use an ensemble of 5 neural network models (each a 2-hidden layer MLP with 64 hidden units per layer). The data is randomly split into an 80-20 training and validation split and trained for 50 epochs using PyTorch's Adam optimizer with a batch size of 500. Early stopping was used if the validation loss began to increase.

For the MB-MBPO comparison, we try to employ hyperparameters from the original paper (and RLlib implementation) for the \texttt{half-cheetah} environment because of its comparable state-dimensions. The MB-MPO policy is the same architecture, but with 32 nodes per hidden layer and trained with a fixed learning rate of size $3 \times 10^{-4}$. The MB-MPO dynamics were represented by an ensemble of 5 fully-connected neural network models with 2 hidden layers of size 512. Each dynamics model was trained for 500 epochs with early stopping if a running mean of the validation loss began to increase. For a fair comparison, the MB-MPO implementation collects $8000$ interactions of random experience at the beginning of training and $1000$ for every on-policy update.

\begin{figure*}[t]
    \centering
    \includegraphics[width=\textwidth]{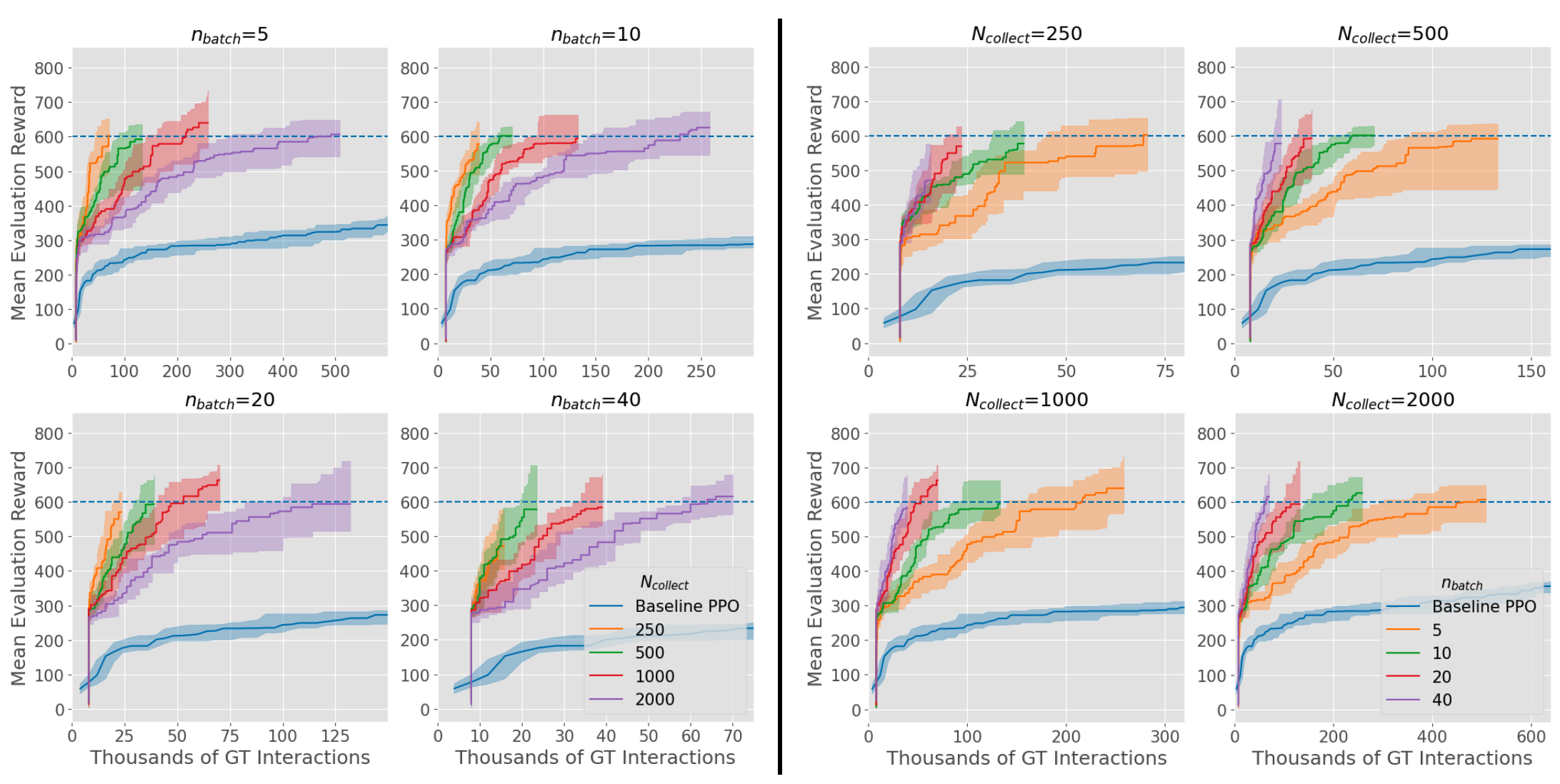}
    \cprotect \caption{A comparison of the number of policy updates, $n_{\text{batch}}$,  and the number of on-policy experience, $N_{\text{collect}}$, collected from the \swingup environment before refitting the dynamics. The dashed blue line corresponds to the median best performance achieved in the baseline comparison.
        \textit{Left}: Each plot compares $(n_{\text{batch}}, N_{\text{collect}})$ pairs where $n_{\text{batch}}$ is fixed and $N_{\text{collect}}$ varies. 
        \textit{Right}: Each plot compares $(n_{\text{batch}}, N_{\text{collect}})$ pairs where $N_{\text{collect}}$ is fixed and $n_{\text{batch}}$ varies. 
    }
     \label{fig:cart-sweep_2_4x4}   
\end{figure*}

\begin{figure*}[t]
    \centering
    \includegraphics[width=\textwidth]{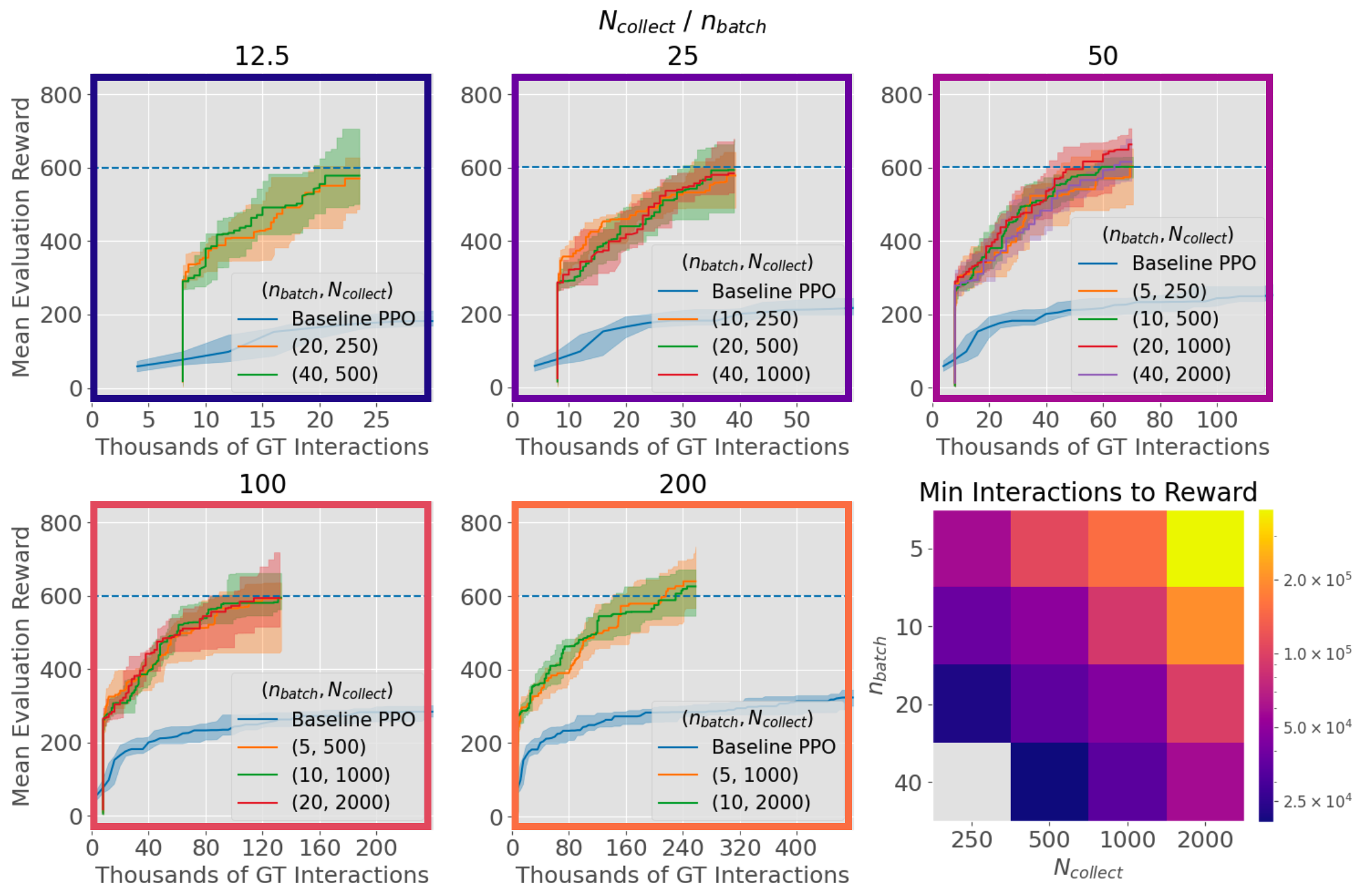}
    \cprotect \caption{A comparison of the number of policy updates, $n_{\text{batch}}$,  and the number of on-policy experience, $N_{\text{collect}}$, collected from the \swingup environment before refitting the dynamics. The dashed blue line corresponds to the median best performance achieved in the baseline comparison. Each of the five timeseries plots are organized into combinations of ($n_{\text{batch}}, N_{\text{collect})}$ with fixed rations. The bottom right heatmap indicates the minimal number of interactions for the median agent to have achieved a reward of 570. The diagonal level sets of the heatmap verify similar performance for the ratios. The (250, 40) combination remains empty because this combination's median best performance never reached the desired threshold. 
    }
     \label{fig:cart-sweep_ratio}   
\end{figure*}

\paragraph{On-Policy Collection Study.} With SINDy-RL, we seek to minimize the number of interactions in the full-order environment. To this end, we investigate effect that the number of policy updates per the dynamics update, $n_{\text{batch}}$, and the amount of on-policy data collected before each dynamics update, $N_{\text{collect}}$. We perform a grid search using $N_{\text{collect}} \in \{250, 500, 1000, 2000\}$ and $n_{\text{batch}} \in \{5, 10, 20, 40\}$ with the \swingup environment and train each environment for $1250$ policy updates (5M samples of surrogate environment updates) using an initial off-policy collection of $8000$ random steps and a maximum on-policy queue of equal size. Just as in the benchmarks from the main text, we evaluate performance on 20 different random seeds and track each policy's best performance. Figure \ref{fig:cart-sweep_2_4x4} demonstrates two different views of the data. Remarkably, the final performance is approximately the same for all $(n_{\text{batch}}, N_{\text{collect}})$ combinations, indicating that for the most part, selecting higher values of $n_{\text{batch}}$ and lower values of $N_{\text{collect}}$ will provide the most rapid policy improvement while minimizing the number of interactions in $\env$. However, the $(n_{\text{batch}}, N_{\text{collect}}) = (40, 250)$ combination does have the minimal performance, indicating that there is an optimal sampling frequency/size. In Figure \ref{fig:cart-sweep_ratio}, we separate experiments by the ratio $N_{\text{collection}}/n_{\text{batch}}$. We clearly see that members from each ratio perform comparably for a fixed budget of $\env$ interactions, indicating that the ratio is a more important quantity than which combination results in the ratio. However, we believe that for extremely large values of $n_{\text{batch}}$, the neural network will overfit to the imperfect dynamics much faster than if it fit more frequently with smaller samples.

\clearpage
% ----------------------------
% Swimmer
% ----------------------------
\subsection{Population-Based Training---Swimmer-v4}
\label{appendix:benchmarking-pbt}
\begin{figure}[t]
    \centering
    \includegraphics[width=0.95\linewidth]{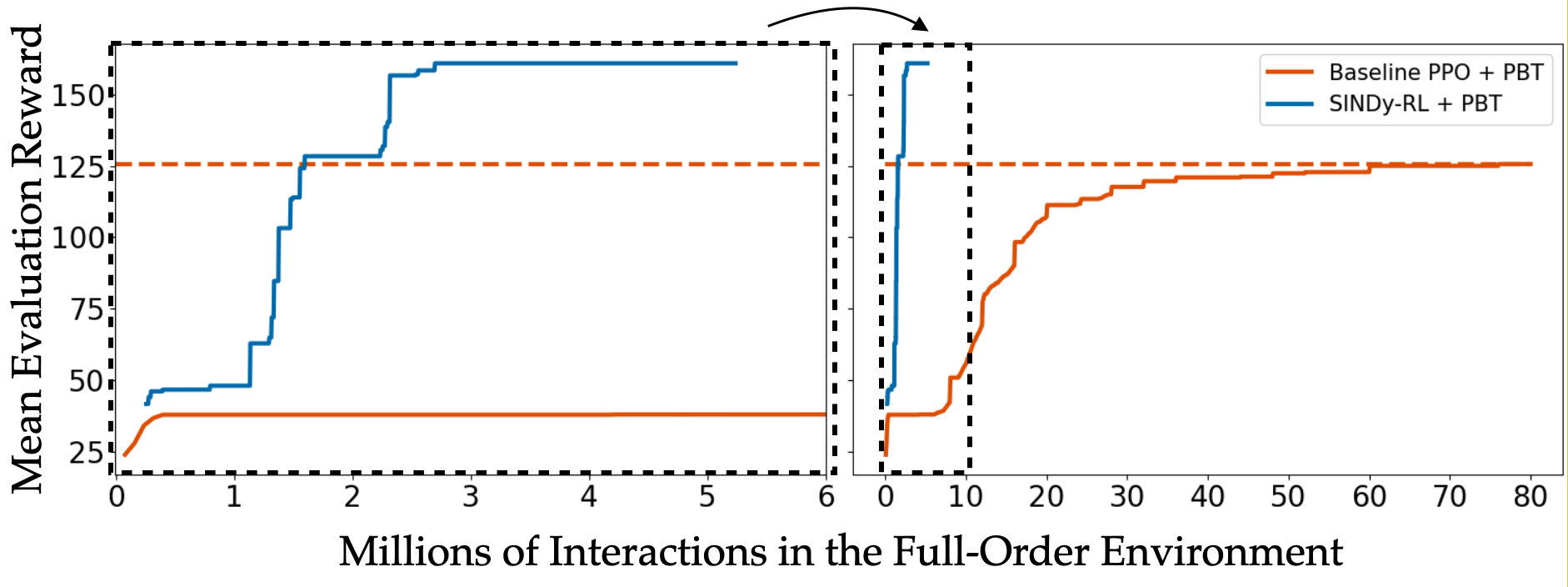}
    \cprotect 
    \caption{\textbf{Swimmer Sample Efficiency}.
    Comparison of the total number of interactions in the full-order \swimmer environment using population-based training (PBT) for PPO hyper-parameter tuning using a population of 20 experiment seeds. Each experiment consists of a single policy and is periodically evaluated over 5 episodes. The population's best evaluation reward is tracked, recorded, and updated. The dashed red line shows the baseline PPO+PBT best final reward.
    }
     \label{fig:swimmer-baseline-best}   
\end{figure}

In population-based training (PBT), a population of candidate policies are trained in parallel using different random seeds and DRL hyperparameters. Periodically the members of the population are evaluated: the bottom performers are stopped and resampled from the top performers with some corresponding mutation (such as changing a hyperparameter). PBT and other population-based frameworks have become a popular method of hyperparameter tuning and avoiding local equilibria that neural networks a extremely prone to. However, PBT for DRL struggles from a practical perspective because these methods must evaluate---not just one, but---many policies in the full-order environment.

We employ the Ray Tune \cite{liaw2018tune} and RLlib implementation for PBT to tune PPO hyperparameters. Evaluation and resampling occurs synchronously every 50 policy updates (200k experience collected from either the surrogate environment for the SINDy-RL implementation or the full-order environment for the baseline PPO). For each experiment, we use a population size of 20 different policies. When evaluating and resampling, the performers from the bottom quartile are stopped and restarted using the policies from the upper quartile. New hyperparameters are either chosen from the original distribution at random or seeded and mutated from the top performer that was copied. Although each hyperparameter can be drawn from a continuous range, we only consider a discrete set of combinations. Explicitly, we consider the following hyperparameter combinations: 
\begin{align*}
    \text{LR} &\in 
        [1 \times 10^{-6} , 5 \times 10^{-6}, 1 \times 10^{-5}, 5 \times 10^{-5}, 1 \times 10^{-4}, 5 \times 10^{-4}, 1 \times 10^{-3}, 5 \times 10^{-3}]
        \\
    \lambda &\in 
        [0.9, 0.95, 0.99, 0.999, 0.9999, 1.0]
        \\
    \gamma &\in 
        [0.9, 0.95, 0.99, 0.999, 0.9999, 1.0]
\end{align*}
all other hyperparameters are fixed and the same as used in the other PPO experiments. The results can be seen in Figure \ref{fig:swimmer-baseline-best}.

\clearpage
\subsection{Cylinder Benchmark}
\label{appendix:benchmarking-cyl}
\begin{figure*}[t]
    \centering
    \includegraphics[width=0.75\textwidth]{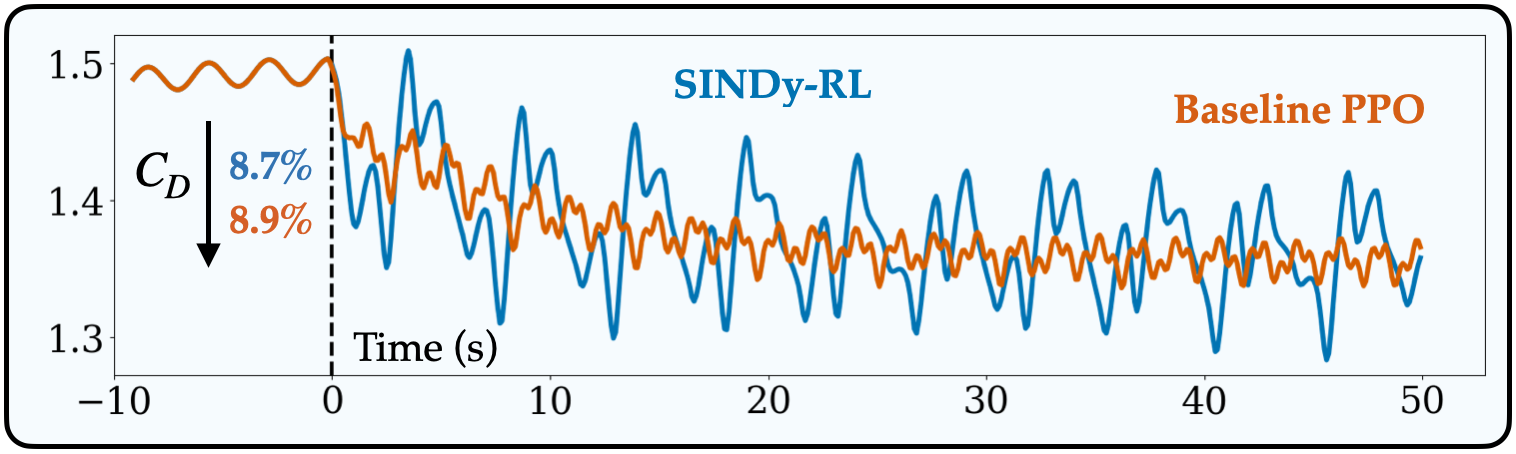}
    \cprotect \caption{\textbf{Cylinder Medium Mesh Comparison}. 
    Evaluation of the best performing SINDy-RL and baseline agents using the fine mesh  after 4,600 and 100,000 full-order interactions respectively. Control starts at $t=0$ marked by the black dashed line. For ease of visualization, the 1-second window-averaged $C_D$ value is plotted for both agents.
    }
     \label{fig:app-bench-cyl_1}   
\end{figure*}

\begin{figure*}[t]
    \centering
    \includegraphics[width=\textwidth]{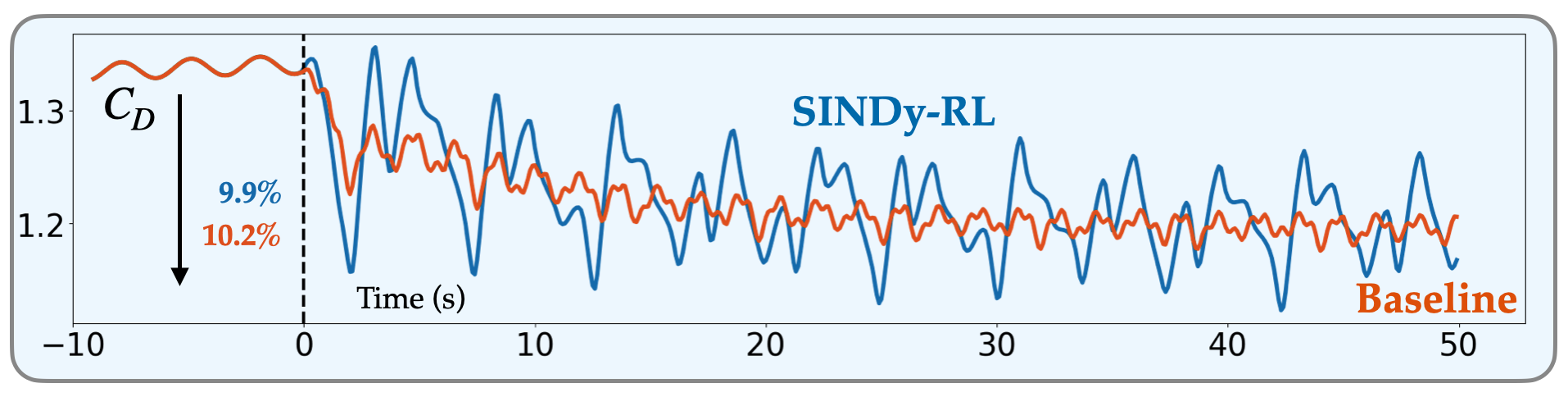}
    \cprotect \caption{\textbf{Cylinder Fine Mesh Comparison}. 
    Evaluation of the best performing SINDy-RL and baseline agents using the fine mesh  after 4,600 and 100,000 full-order interactions respectively. Control starts at $t=0$ marked by the black dashed line. For ease of visualization, the 1-second window-averaged $C_D$ value is plotted for both agents.
    }
     \label{fig:app-bench-cyl}   
\end{figure*}

\hydrogym also supports a fine mesh consisting of hundreds of thousands of finite elements. Because of its massive size, it's not recommended for training on the Cylinder. However, our agents only receive information about the lift coefficients and are thus mesh-agnostic. While we never collect data from the fine mesh, we can still deploy our trained agents to it. Figures \ref{fig:app-bench-cyl_1} and \ref{fig:app-bench-cyl} depict the agents from the benchmarks in the main text being applied to the medium and fine mesh, validating that our approach translates to unseen, higher-fidelity environments.

\clearpage
\subsection{Pinball Benchmark}
\label{appendix:benchmarking-pinball}

\begin{figure*}[h]
    \centering
    \includegraphics[width=\textwidth]{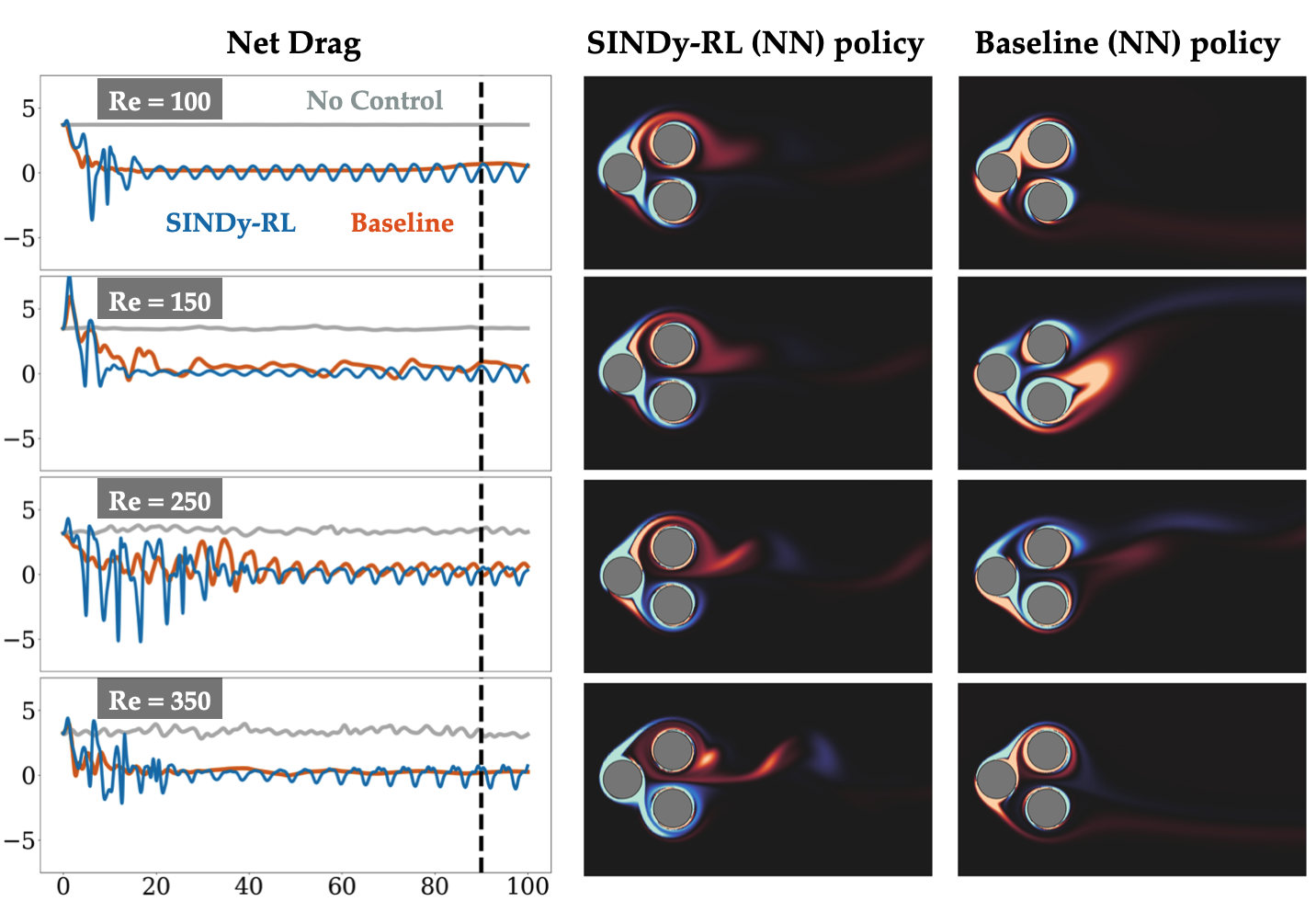}
    \cprotect \caption{\textbf{Pinball $Re$ evaluation}. 
    Evaluation of the best performing SINDy-RL and baseline agents across $Re$. \textit{Left:} Net drag experienced on the system for the SINDy-RL agents (blue), baseline agents(red), and without control (gray). Negative values indicate thrust generation. Dashed line at $t=90$ corresponds to the snapshots on the right. \textit{Right:} Vorticity snapshots at $t=90$.}
     \label{fig:app-bench-pin}   
\end{figure*}

In Figure \ref{fig:app-bench-pin}, we plot the comparison of the net drag experienced on the system with and without control for the SINDy-RL and baseline approaches for various $Re$ when only being trained at $Re=100$. For each $Re$, the two approaches provide similar values for net drag with the SINDy-RL agent oscillating near zero, and veering more into the negative (higher rewards) corresponding to larger thrust production. 

\clearpage

\subsection{3D Airfoil Benchmark}
\label{appendix:benchmarking-airfoil}

\begin{figure*}[h]
    \centering
    \includegraphics[width=\textwidth]{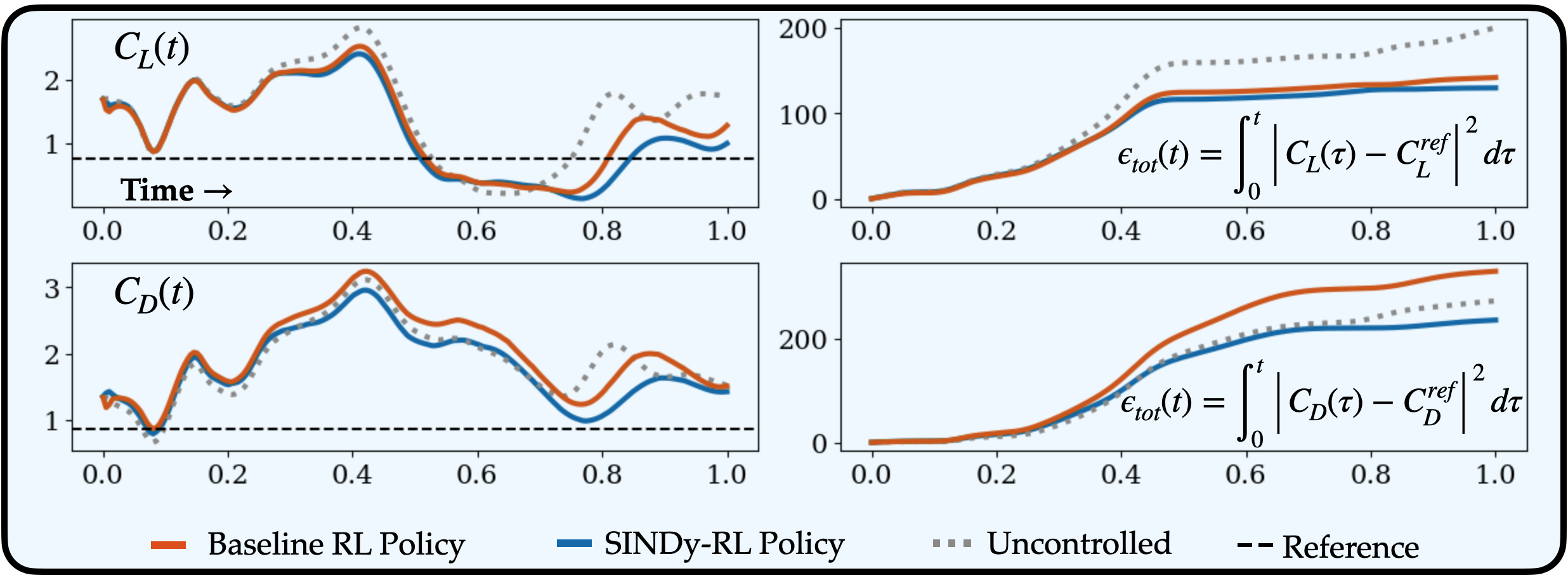}
    \cprotect \caption{\textbf{\airfoil Evaluation}. 
    Evaluation of the best performing SINDy-RL and baseline agents. \textit{Left:} Net lift and drag experienced on the system for the SINDy-RL agents (blue), baseline agents(red), and without control (gray dotted). \textit{Right:} Cumulative error from the desired reference values.}
     \label{fig:app-bench-naca}   
\end{figure*}

In Figure \ref{fig:app-bench-naca} we compare the effect of the SINDy-RL policies and the baseline PPO policy. Both agents are able to drive the $C_L$ force closer to the desired reference value much sooner than the uncontrolled case, with the SINDy-RL agent performing slightly better. The SINDy-RL agent also does a better job of mitigating the gust for the $C_D$ value, but to a lesser extent; however, the PPO baseline agent actually increases the amount of drag on the airfoil.

\begin{table}
    \centering
    \begin{tabular}{|c|c|c|c|c|c|}
    \hline
            Env    
                &Full step    
                &Surrogate step       
                &Surrogate policy training
                &Evaluation      
                &E-SINDy Fit \\
            \hline
            Cylinder  
                & $6.86$s      
                & $1.55 \times 10^{-3}$s 
                & $3.46 \times 10^2$s
                & $\mathbf{1.37 \times 10^{3}}$s 
                & $5.9 \times 10^{-2}$s\\
            Pinball  
                & $1.78$s       
                & $3.77 \times 10^{-3}$s 
                & $4.59 \times 10^2$s 
                & $\mathbf{1.56 \times 10^{3}}$s 
                & $13.5$s \\
            3D Airfoil  
                & 45s       
                & $3.49\times 10^{-4}$s
                & $3.20 \times 10^{2}$s
                & $\mathbf{1.13 \times 10^3}$s
                & $5.0 \times 10^{-2}$s \\
         \hline
    \end{tabular}
    \caption{\textbf{Fluid Environment clock-times}. A comparison of different clock-times for SINDy-RL on the \hydrogym environments. ``Full step'' and ``Surrogate step'' measure the amount of time used to take a single step in the full-order and surrogate environments---for the Cylinder and Pinball, this was derived from Ray-RLlib ``mean\_env\_wait\_ms'' metric; for the 3D Airfoil all values were estimated manually from training logs.
    ``Surrogate policy training'' refers to the amount of time used to train the policy in the surrogate environment before updating the surrogate models (for $25$ policy updates with $4000$ interactions with the surrogate per step). ``Evaluation'' refers to the amount of time for evaluating the environment (i.e. collecting $N_{\text{collect}}$ worth of experience and updating the surrogate environments). ``E-SINDy Fit'' refers to the amount of time to fit the 20-member ensemble of E-SINDy models in serial using the \textit{maximal} amount of data allowed by the datastore buffer.
    \\
    \textit{
    Note: the actuation frequency used for the Cylinder environment was 10 times smaller than the Pinball (10Hz compared to 100Hz), resulting in a much larger simulation time in between actions.}
    }
    \label{tab:clock_times}
\end{table}

\clearpage

% ----------------------------------------------
\section{Dictionary Dynamics}
\label{appendix:dynamics}
We learn the surrogate dictionary dynamics and reward functions for all environments, with the exception of the \swingup task where we only learn the dynamics. For the results presented in this work, we restrict ourselves to learning discrete control-affine dynamics
\begin{equation} \label{eq:affine}
    x_{n+1} = f(x_n) + g(x_n)u_n
\end{equation}
where $f$ and $g$ are polynomials of some maximal degree, though a much broader class of libraries and SINDy-algorithms can be used, such as SINDy-PI \cite{kaheman2020sindy}. For dictionary learning, we use \pysindy's \cite{desilva2020pysindy, Kaptanoglu2022pysindy} E-SINDy implementation with an ensemble of $N_e = 20$ models using STLRidge. Table \ref{tab:SINDy} summarizes the libraries and hyperparameters for each model. During the Dyna-style training in \appendixAlgoRef from the main text, trajectories are terminated and the surrogate environments are reset if the observations exceed environment specific thresholds, which are detailed below. 

\begin{table*}[h]
    \centering
    \begin{tabular}{|c|c|c|c|c|c|c|c|c|c|}
        \hline
         Environment &$f(x)$                        &$g(x)$         &Thresh             &$\alpha$           &Ensemble\\
         \hline
         \swingup    &Quadratic                     &Quadratic      &$7 \times 10^{-3}$ &$5 \times 10^{-5}$ & Median  \\
         \swimmer    &Quadratic (no cross)   &Quadratic (no cross)  &$2 \times 10^{-2}$ &$5 \times 10^{-1}$ & Median \\
         Cylinder    &Cubic                         &$1$            &$1 \times 10^{-3}$ &$1 \times 10^{-5}$ & Median \\
         Pinball     &Quadratic                      & $1$          &$1 \times 10^{-3}$ &$1 \times 10^{-5}$ & Median\\
         \airfoil     &Quadratic (no cross)                     & Quadratic (no cross)         &$1 \times 10^{-2}$ &$1 \times 10^{-5}$ & Median\\
         \hline
    \end{tabular}
    \caption{For each environment, the library types for $f,g$ (Eq. \ref{eq:affine}) are listed with the sparsity threshold and ridge regression weight, $\alpha$. ``Ensemble'' refers to whether the median or mean coefficients were used. No $f$ libraries include a bias term. ``no cross'' implies no cross terms in the quadratics. }
    \label{tab:SINDy}
\end{table*}

\subsection{Swing-up}
\label{appendix:dynamics-swingup}

\begin{figure}[t]
    \centering
    \includegraphics[width=0.95\linewidth]{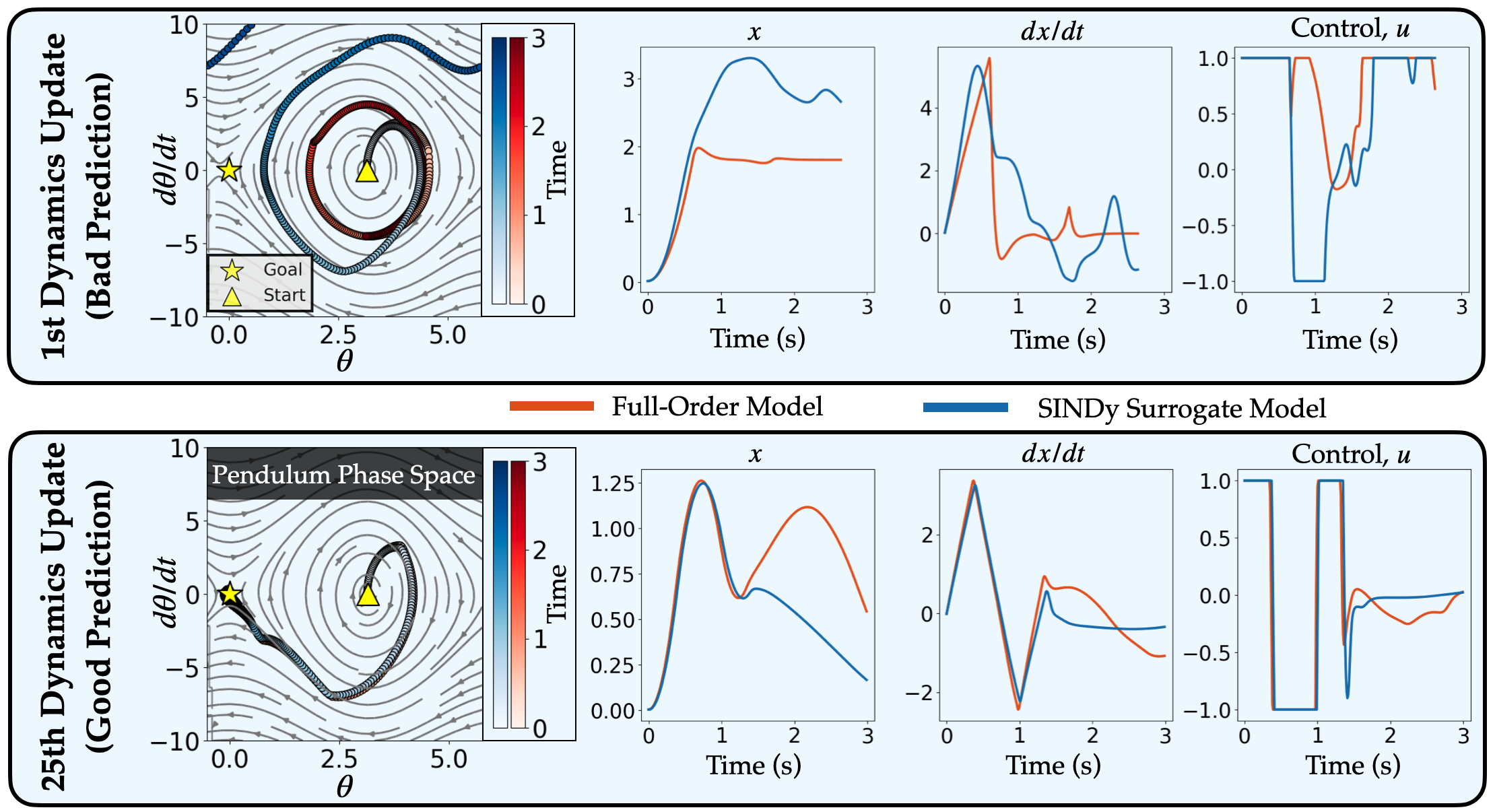}
    
    \cprotect \caption{\textbf{Swing-up Dynamics.} The learned dynamics and control of the \dmcontrol \swingup environment by rolling out the DRL control policy in the high fidelity and surrogate environments after the 1st and 25th dynamics updates. 
    \textit{Left:}
    The phase portrait for the angle, $\theta$ (evaluated at $x=\dot{x} = u =0$ for ease of visualization), and the corresponding DRL-agent trajectories of the full-order and surrogate models. 
    \textit{Right:} The DRL-agent trajectories of the full-order and surrogate models for the $x, \dot{x}$ components of the state and the corresponding control.
    }
     \label{fig:cartpole-dyn}   
\end{figure}

\begin{figure*}[h!]
    \centering
    \includegraphics[width=0.95\textwidth]{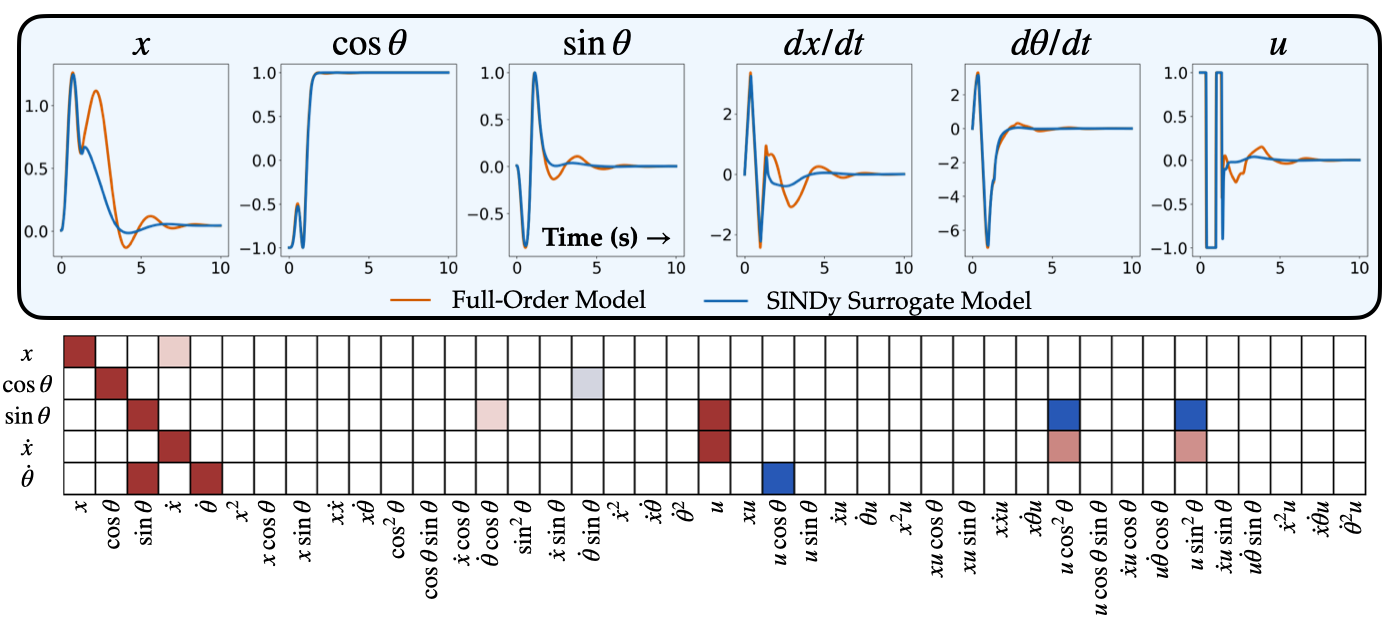}
    \cprotect \caption{\textbf{Swing-Up Dynamics.}
    \textit{Top:} Rollouts of the SINDy surrogate (blue) and full-order (red) dynamics under the influence of feedback control of the learned neural network  policy in each respective environment after being trained with \appendixAlgoRef. 
    \textit{Bottom:} A heatmap of the learned dictionary coefficients, $\vXi$. For ease of visualizing smaller coefficients, each row's coefficients are normalized by the maximum value and the color map clips between  small positive (red) and negative (blue) value of equal magnitude to demonstrate the sign.
    }
     \label{fig:app-dyn-cart}   
\end{figure*}

As shown in Figure \ref{fig:app-dyn-cart}, we use the affine library as in Equation \ref{eq:affine} where $f$ and $g$ are polynomials of maximal degree 2. Note that because our state-space includes $\sin(\theta)$ and $\cos(\theta)$ terms, there are  redundant combinations of library functions due to the trigonometric identity: $\cos(\theta)^2 + \sin(\theta)^2 = 1$. To accommodate for this redundancy, we do not incorporate a constant offset into the library. The final learned dynamics are given by the equation below, highlighting the Euler-like update, which gives rise to the diagonal entries in Figure \ref{fig:app-dyn-cart}. 
\begin{align*}
    x_{k+1} 
        &= 1.000 x_k + 0.010 \dot{x}_k 
    \\
    \cos(\theta_{k+1}) 
        &= 1.000 \cos(\theta_k)  -0.010 \sin(\theta_k) \dot{\theta}_k
    \\
    \sin(\theta_{k+1}) 
        &= 0.999 \sin(\theta_k) + 0.010 \cos(\theta_k) \dot{\theta}_k
    \\
    \dot{x}_{k+1} 
    &= 0.998 \dot{x}_k + 0.063 u_k + 0.032 \cos(\theta_k)^2 u_k + 0.030 \sin(\theta_k)^2 u_k
    \\
    \dot{\theta}_{k+1} 
        &= 1.000 \dot{\theta}_k + 0.148 \sin(\theta_k) +  0.005 x_k u_k \\ 
        & \quad -0.142 \cos(\theta_k) u_k  -0.007 x_k^2 u_k + 0.004 x_k \cos(\theta_k) u_k
\end{align*}
To fit the SINDy model, we first collect $N_{\text{off}} = 8000$ interactions (80 seconds) of random experience to fit the dynamics. For rolling out the surrogate dynamics, we assume prior knowledge that the angle is embedded on the circle $(\cos(\theta), \sin(\theta))$ and renormalize after every step taken in the surrogate dynamics. 
During training, we collect $N_{\text{collect}} = 1000$ interactions (10 seconds) worth of on-policy experience for every dynamics update, and $\mathcal{D}_{\text{on}}$ employed a queue of size $8000$. 

A trajectory is terminated and the surrogate environment is reset if the observations exceed the following thresholds:
$$
    |x| \leq 5, \qquad 
    |\cos\theta| \leq 1.1, \qquad 
    |\sin\theta| \leq 1.1, \qquad
    |\dot{x}| \leq 10, \qquad
    |\dot \theta | \leq 10
$$
For the linear dynamics models used in the benchmark comparison, we use a linear-affine library (with bias) and set the threshold and regularization coefficients to zero.

\subsection{Swimmer-v4}
\label{appendix:dynamics-swimmer}
\begin{figure*}[t!]
    \centering
    \includegraphics[width=0.95\textwidth]{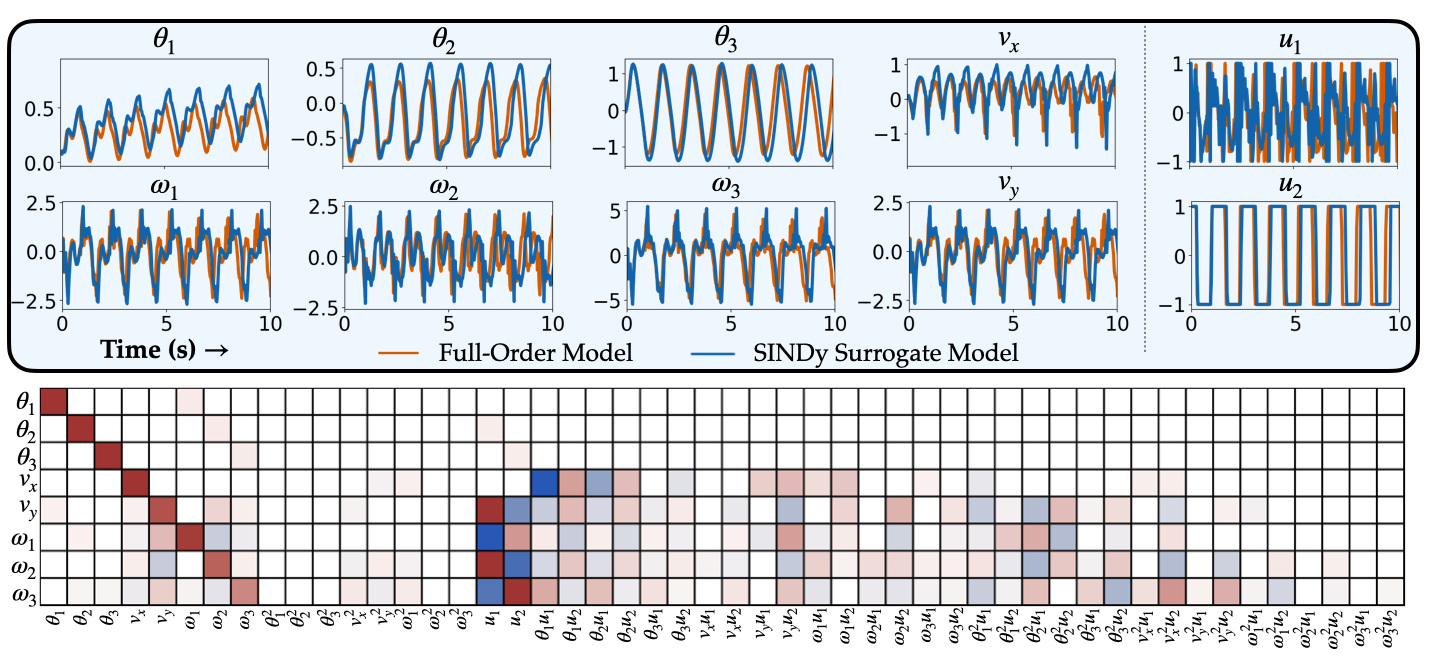}
    \cprotect \caption{\textbf{Swimmer-v4 Dynamics.}
    \textit{Top:} Rollouts of the SINDy surrogate (blue) and full-order (red) dynamics under the influence of feedback control of the learned neural network  policy in each respective environment after being trained with \appendixAlgoRef. 
    \textit{Bottom:} A heatmap of the learned dictionary coefficients, $\vXi$. For ease of visualizing smaller coefficients, each row's coefficients are normalized by the maximum value and the color map clips between  small positive (red) and negative (blue) value of equal magnitude to demonstrate the sign.
    }
     \label{fig:app-dyn-swim}   
\end{figure*}

Just as in the \swingup example, we use \pysindy to fit the dynamics (along with the reward function---see \TheAppendix \ref{appendix:rewards} for more details).  
As shown in Figure \ref{fig:app-dyn-swim}, we use a quadratic dynamics library in terms of the observation components without the interacting terms; the learned dynamics match well with the full-order environment under the influence of the control policy. Unlike the $\swingup$ environment, we make no assumptions on the manifold that the state-space live when rolling out the surrogate dynamics. However, we do include bounds:
$$
    |\theta_1 | \leq \pi, \qquad 
    |\theta_2|, |\theta_3| \leq 1.7453, \qquad
    |v_x|, |v_y|, |\dot \theta_k| \leq 10
$$
We bound $\theta_2$ and $\theta_3$ so that we do not reach the maximal angle for the swimmer robot. Physically, it is reasonable to assume knowledge of this ahead of time in order to not damage the physical asset. This also avoids the piecewise dynamics needed to describe the environment. If there is prior knowledge about how the dynamics behave near the boundary these could also be incorporated (e.g. clipping the state and zero-ing out the velocity terms) but was not explored in this work.

\begin{figure*}[t]
    \centering
    \includegraphics[width=0.95\textwidth]{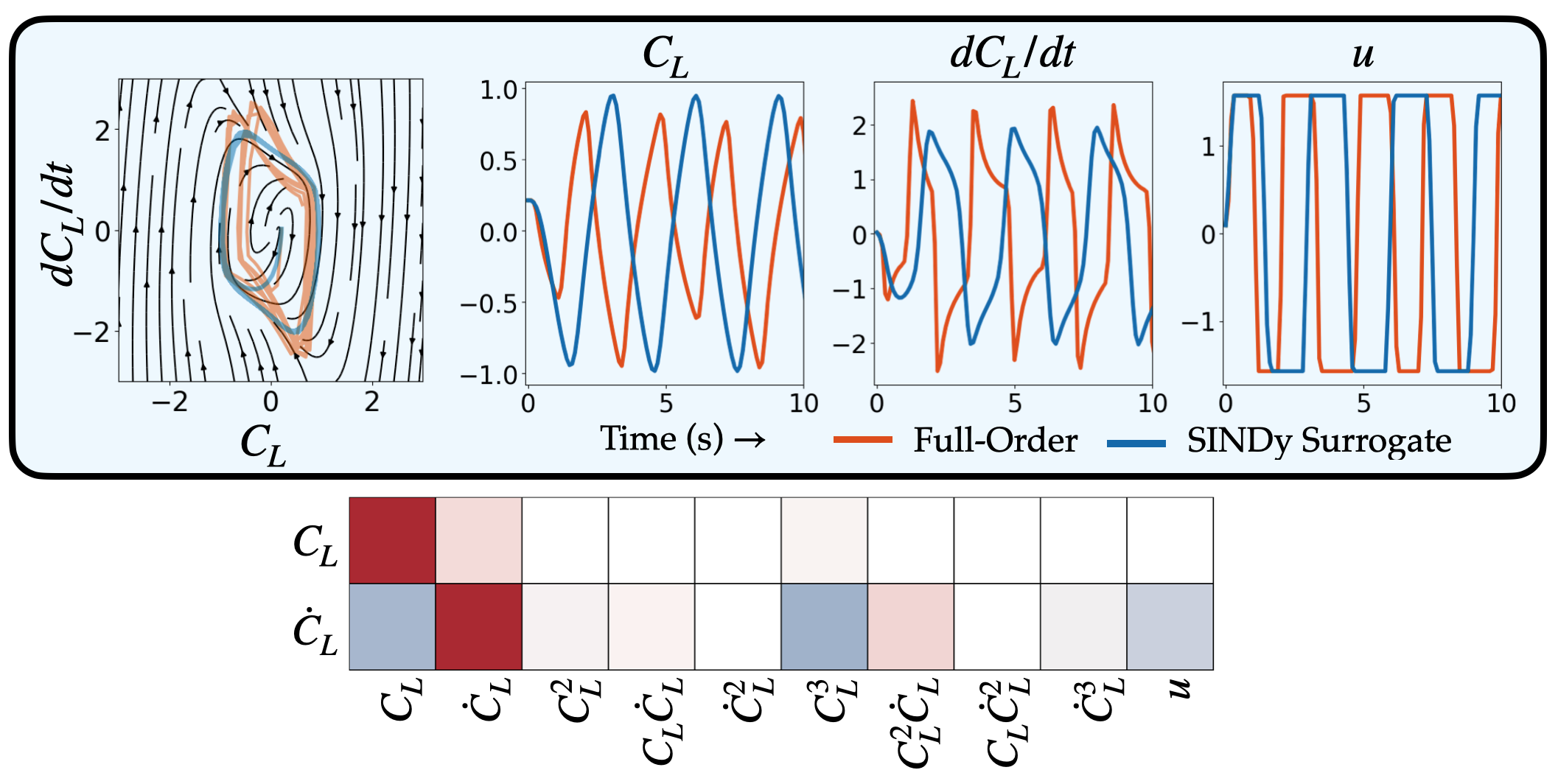}
    \cprotect \caption{\textbf{Cylinder Dynamics.}
    \textit{Top:} Rollouts of the SINDy surrogate (blue) and full-order (red) dynamics under the influence of feedback control of the learned neural network  policy in each respective environment after being trained with Algorithm \appendixAlgoRef. A vector field of the learned dynamics with feedback control is plotted in phase space on the left. 
    \textit{Bottom:} Heatmap of the learned dictionary coefficients, $\vXi$. For ease of visualizing smaller coefficients, each row's coefficients are normalized by the maximum value and the color map clips between  small positive (red) and negative (blue) value of equal magnitude to demonstrate the sign.
    }
     \label{fig:app-dyn-cyl}   
\end{figure*}

\subsection{Cylinder}
\label{appendix:dynamics-cylinder}
As shown in Figure \ref{fig:app-dyn-cyl}, the Cylinder dynamics were fit using a 3rd degree polynomial (without interaction terms) in the state variables $(C_L, \dot C_L)$ and a completely isolated control term. A conservative set of bounds were used on the observation space: 

$$
|C_L | \leq 10, \qquad |\dot C_L| \leq 200
$$

Under the influence of feedback control, a stable limit cycle emerges in the surrogate dynamics. However, as indicated by the data plotted in red, the true dynamics vary slightly and leave the cycle.

\subsection{Pinball}
\label{appendix:dynamics-pinball}
\begin{figure*}[t]
    \centering
    \includegraphics[width=0.95\textwidth]{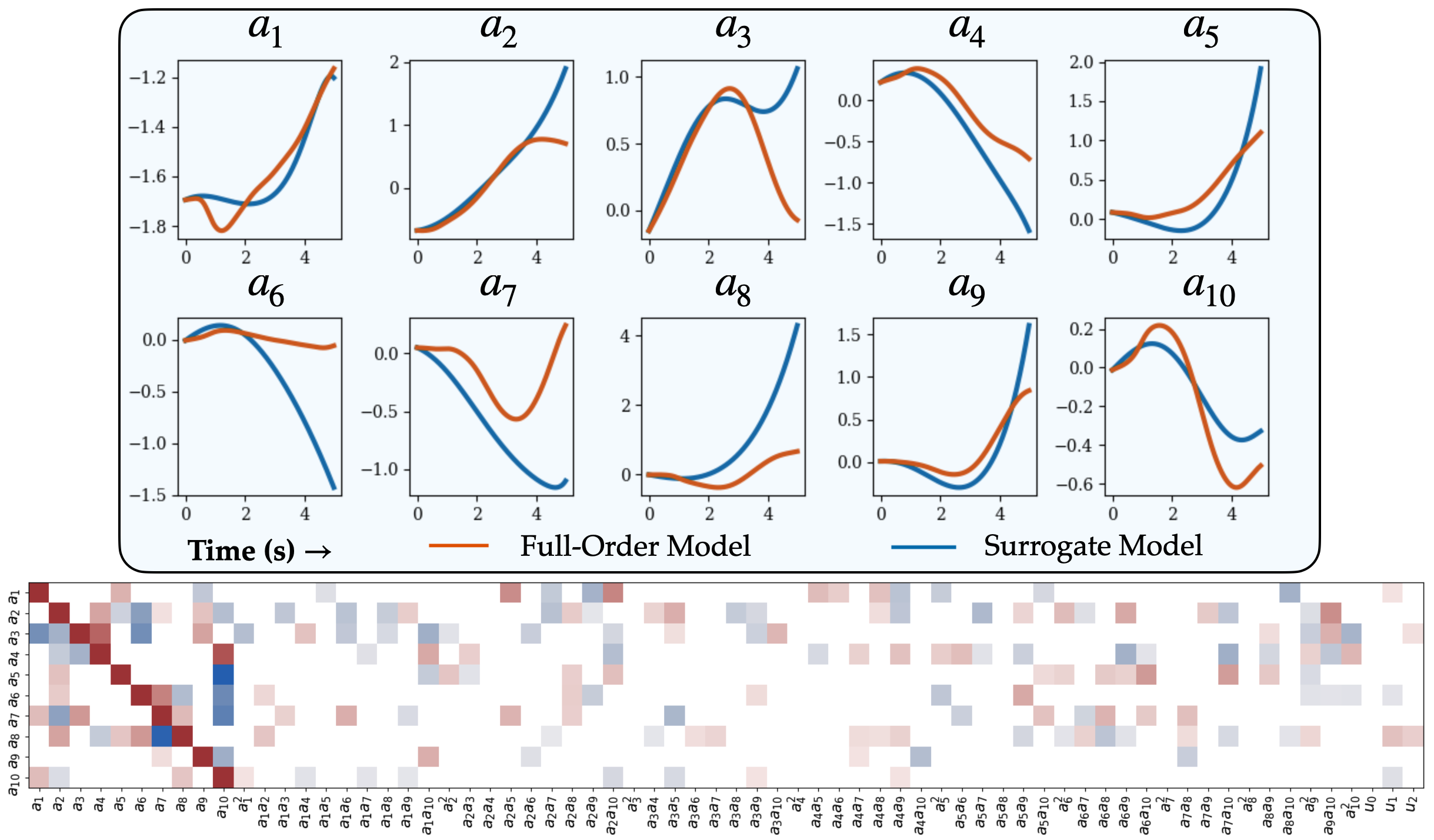}
    \cprotect \caption{\textbf{Pinball Dynamics.}
    \textit{Top:} Rollouts of the SINDy surrogate (blue) and full-order (red) dynamics under the influence of feedback control of the learned neural network  policy in each respective environment after being trained with \appendixAlgoRef.
    \textit{Bottom:} Heatmap of the learned dictionary coefficients, $\vXi$. For ease of visualizing smaller coefficients, each row's coefficients are normalized by the maximum value and the color map clips between  small positive (red) and negative (blue) value of equal magnitude to demonstrate the sign.
    }
     \label{fig:app-dyn-pin}   
\end{figure*}

As shown in Figure \ref{fig:app-dyn-pin}, the Pinball dynamics were fit using a 2nd degree polynomial in the SVD coefficients $a_i$  and completely isolated control terms, $u_i$. A conservative set of bounds were used on the observation space: 

$$
    |a_i | \leq 20
$$

\clearpage

\subsection{3D Airfoil}
\label{appendix:dynamics-airfoil}

\begin{figure*}[t]
    \centering
    \includegraphics[width=0.95\textwidth]{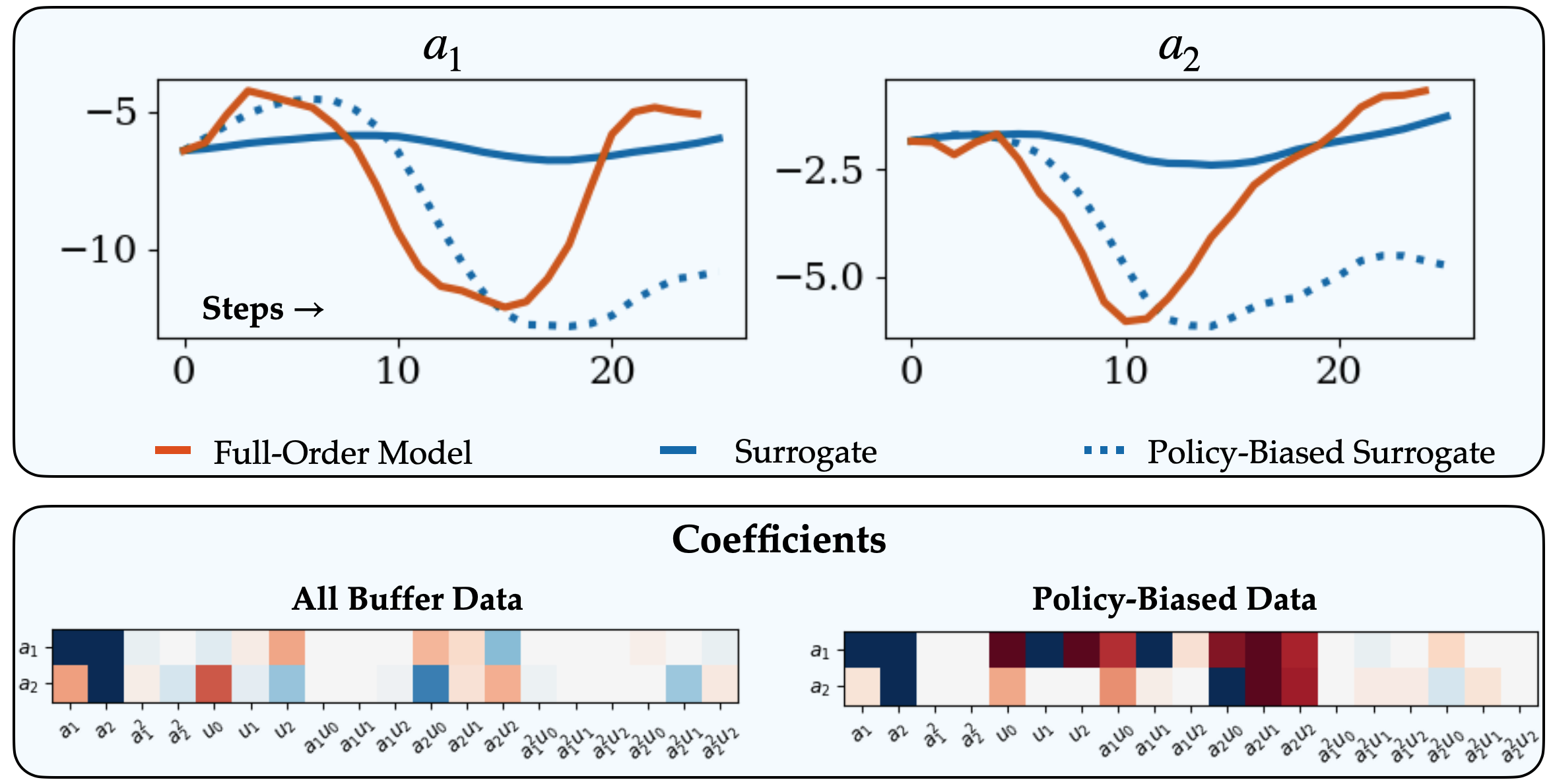}
    \cprotect \caption{\textbf{\airfoil Dynamics.}
    \textit{Top:} Rollouts of the SINDy surrogate (blue) and full-order (red) dynamics under the influence of feedback control of the learned neural network  policy in each respective environment after being trained with \appendixAlgoRef. Solid blue line corresponds to the learned dynamics model during training that used all 10 of the initially collected trajectories and all subsequent evaluations. Dotted blue line corresponds to a biased model that was only trained on 3 of the collected data and all subsequent evaluations (Note: biased model is only included as a demonstration; it was not used during training). 
    \textit{Bottom:} Heatmap of the learned dictionary coefficients, $\vXi$ for the learned dynamics model during training (left) and the model that was biased on the data collected from training the policy (right). For ease of visualizing smaller coefficients, each row's coefficients are normalized by the maximum value and the color map clips between  small positive (red) and negative (blue) value of equal magnitude to demonstrate the sign.
    }
     \label{fig:app-dyn-naca}   
\end{figure*}

As shown in Figure \ref{fig:app-dyn-naca}, the \airfoil dynamics were fit using a 2nd degree polynomial in the SVD coefficients $a_i$ and control terms $u_i$. A conservative set of bounds were used on the observation space: 

$$
    |a_i | \leq 1000
$$

For training our SINDy-RL agents,  used the \textit{entirety} of the off-policy buffer (consisting of the initial random trajectories) was used and was sufficient to train a meaningful control policy. However if the dynamics for the randomly controlled agent are substantially different than the controlled agent, then it may be favorable to prune those trajectories later during training. Figure \ref{fig:app-dyn-naca} demonstrates the difference in rolling out the final surrogate dynamics model that used all the available data vs. a model that only used 30\% of the initial random trajectories (i.e. one that is biased towards the on-policy collected data). While the biased-dynamics struggle to forecast the final behavior of the trajectory, the dynamics agree reasonably well for the first half of the trajectory. In contrast, when using the full dataset, the influence of control is greatly tempered. This can be verified by examining the coefficients; the biased model exhibit a much stronger dependence on the control terms.

\clearpage

% ----------------------------------------------
\section{Dictionary Rewards}
\label{appendix:rewards}
Just as with the dynamics, we use \pysindy's E-SINDy implementation with $N_e = 20$ models and STLRidge. All models use the median coefficients for predicting rewards. A summary of libraries and hyperparameters can be found in Table \ref{tab:rewards}.

\begin{table*}[h]
    \centering
    \begin{tabular}{|c|c|c|c|c|c|c|c|}
        \hline
         Environment &Libray         &Thresh             &$\alpha$           \\
         \hline
         \swimmer    &Quadratic$^*$  &$5 \times 10^{-2}$ &$5 \times 10^{-5}$  \\
         Cylinder    &Quadratic      &$1 \times 10^{-4}$ &$5 \times 10^{-5}$  \\
         Pinball     &Quadratic      &$5 \times 10^{-1}$ &$1 \times 10^{-5}$ \\
         \airfoil     &Cubic $^*$      &$6 \times 10^{-4}$ &$1 \times 10^{-5}$ \\
         \hline
    \end{tabular}
    \caption{For each environment, the library types for the reward function are listed along with the STLS threshold and the ridge regression weight, $\alpha$.  $^*:$ The swimmer and \airfoil libraries did not  contain cross terms. The swimmer did not contain a bias, nor terms involving the control variables.}
    \label{tab:rewards}
\end{table*}

\subsection{Swing-up}
\label{appendix:rewards-swingup}
We do not learn the dictionary rewards as we have access to an analytic equation, as described in \TheAppendix \ref{appendix:envs}. We also expect that it would be challenging to learn such a reward function because it is a product of four functions, each with an approximately quadratic Taylor Series---indicating that the rewards would only be well described near the equilibrium by an 8th order polynomial and would have trouble decaying away from the equilibrium. While a library with better properties could be examined, it was not investigated in this work.

\subsection{Swimmer-v4}
\label{appendix:rewards-swimmer}
\begin{figure*}[h]
    \centering
    \includegraphics[width=0.95\textwidth]{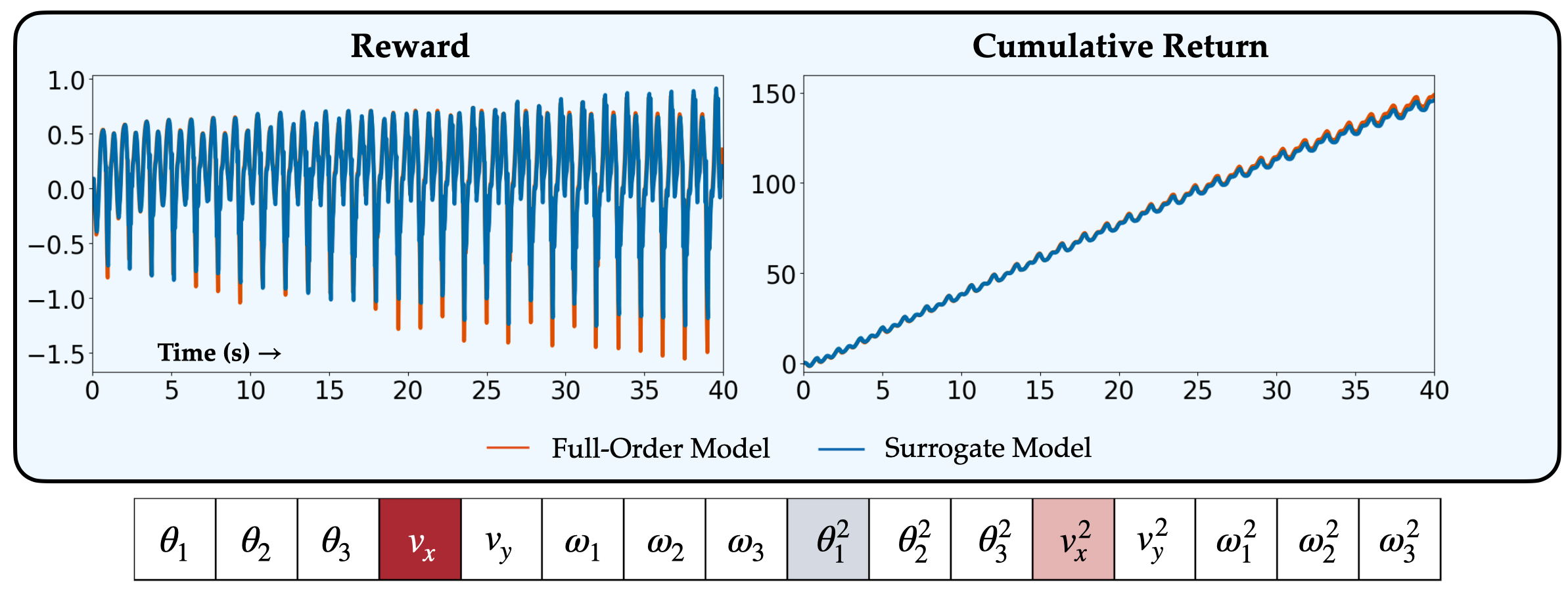}
    \cprotect \caption{\textbf{Swimmer-v4 Reward.}
    \textit{Top:} Surrogate dictionary reward predictions evaluated on trajectories produced from the full-order environment.    
    \textit{Bottom:} A heatmap of the learned dictionary coefficients, $\vXi$. For ease of visualizing smaller coefficients, the row's coefficients are normalized by the maximum value and the color map clips between  small positive (red) and negative (blue) value of equal magnitude to demonstrate the sign.
    }
     \label{fig:app-rew-swim}   
\end{figure*}

As illustrated in Figure \ref{fig:app-rew-swim}, we use a quadratic dictionary on the observation variables with no interaction terms to model the \swimmer surrogate reward. Using median coefficients, the final learned reward was:
$$
    \hat{r} = 0.917 v_x  - 0.086 \theta_1^2 + 0.141 v_x^2, 
$$

\noindent indicating that the  $x$-velocity of the leading segment $v_x$ is a good proxy for the body-center velocity that is not present. Figure \ref{fig:app-rew-swim} depicts the predictions of the reward during a full-rollout of the controlled full-order environment, and indeed the rewards are well-correlated---especially for the cumulative return used to estimate the value function.

\subsection{Cylinder}
\label{appendix:rewards-cylinder}
\begin{figure*}[h]
    \centering
    \includegraphics[width=0.95\textwidth]{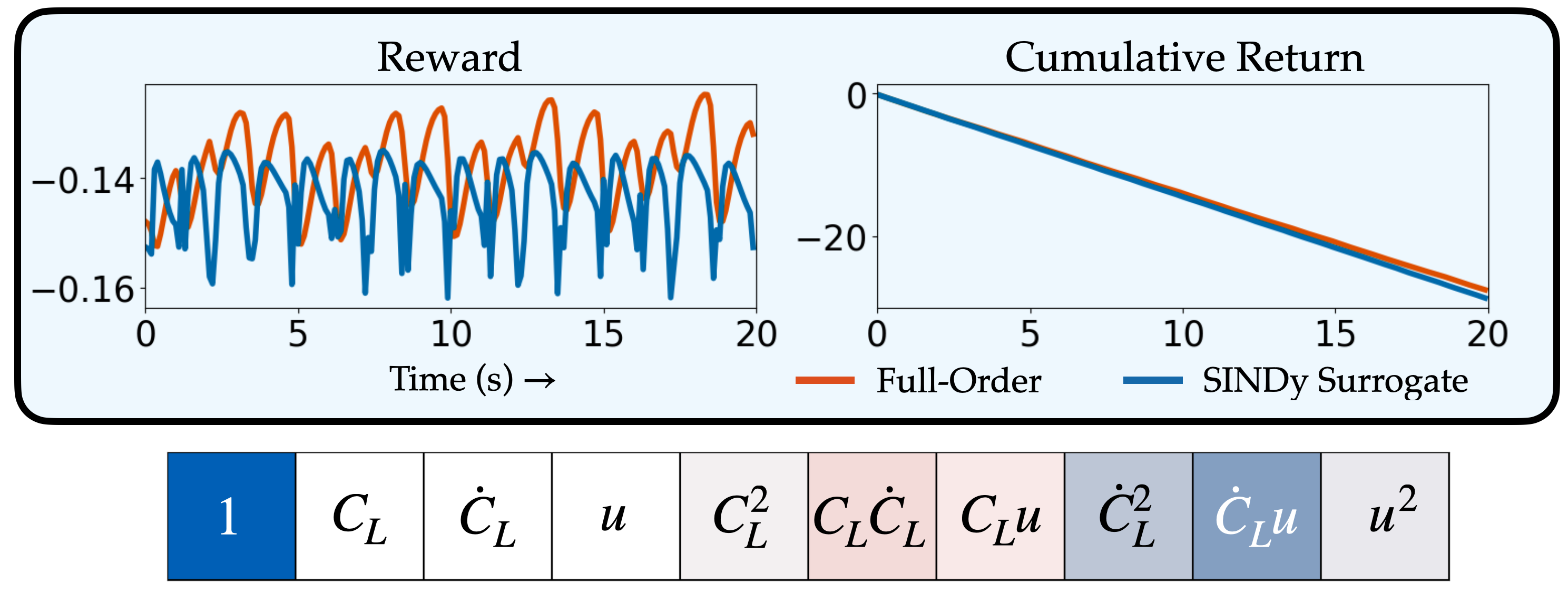}
    \cprotect \caption{\textbf{Cylinder Reward.}
    \textit{Top:} Surrogate dictionary reward predictions evaluated on trajectories produced from the full-order environment. 
    \textit{Bottom:} A heatmap of the learned dictionary coefficients, $\vXi$. For ease of visualizing smaller coefficients, the row's coefficients are normalized by the maximum value and the color map clips between  small positive (red) and negative (blue) value of equal magnitude to demonstrate the sign.
    }
     \label{fig:app-rew-cyl}   
\end{figure*}

As illustrated in Figure \ref{fig:app-rew-cyl}, a quadratic library with mixing terms was used to fit the surrogate reward function. The resulting model was found to be

\begin{align*}
\hat r = 
-( 1.52 \times 10^{-1}) \quad 
&-  (8.21 \times 10^{-4})C_L 
&- (9.85 \times 10 ^{-4}) \dot C_L 
&+ (6.22 \times 10^{-3}) C_L \dot C_L  
&- (2.72 \times 10^{-3} ) C_L^2
\\
-(5.94 \times 10^{-4} ) u 
&+ (3.34 \times 10^{-3} ) C_L u  
&- (1.34 \times 10^{-2}) \dot C_L^2  
&- ( 2.05 \times 10^{-2})  \dot C_L u 
&- (5.06 \times 10^{-3} ) u^2 
\end{align*}

In Figure \ref{fig:app-rew-cyl}, the \textit{raw} surrogate reward function is shown to be more sensitive than the full-order environment rewards, and does not provide very accurate immediate predictions. Despite this, the reward provided a suggestive enough learning signal to train SINDy-RL agents to successfully reduce the drag. 

\subsection{Pinball}
\label{appendix:rewards-pinball}
In Figure \ref{fig:app-bench-pin}, the median surrogate reward function is shown to be more ``optimistic'' than the ground truth. While reasonably correlated, the median coefficients often predicts higher rewards, resulting in a net cumulative reward that is significantly larger. The individual members of the ensemble, however, have magnitudes closer to the ground truth. While not explored, this indicates that the \textit{mean} coefficients might be better suited for the purposes of training, whereas the median are better for interpretability and compression. Through the sparsity promoting optimization, only 24 out of the 105 median coefficients are nonzero. Because the observation only contains information from sensors in the wake, the reward relies significantly on the control input, which directly affects the local field and fluid forces on the system. The result includes non-trivial coupling between the sensor modes and the actuation.

\begin{figure*}[h]
    \centering
    \includegraphics[width=0.95\textwidth]{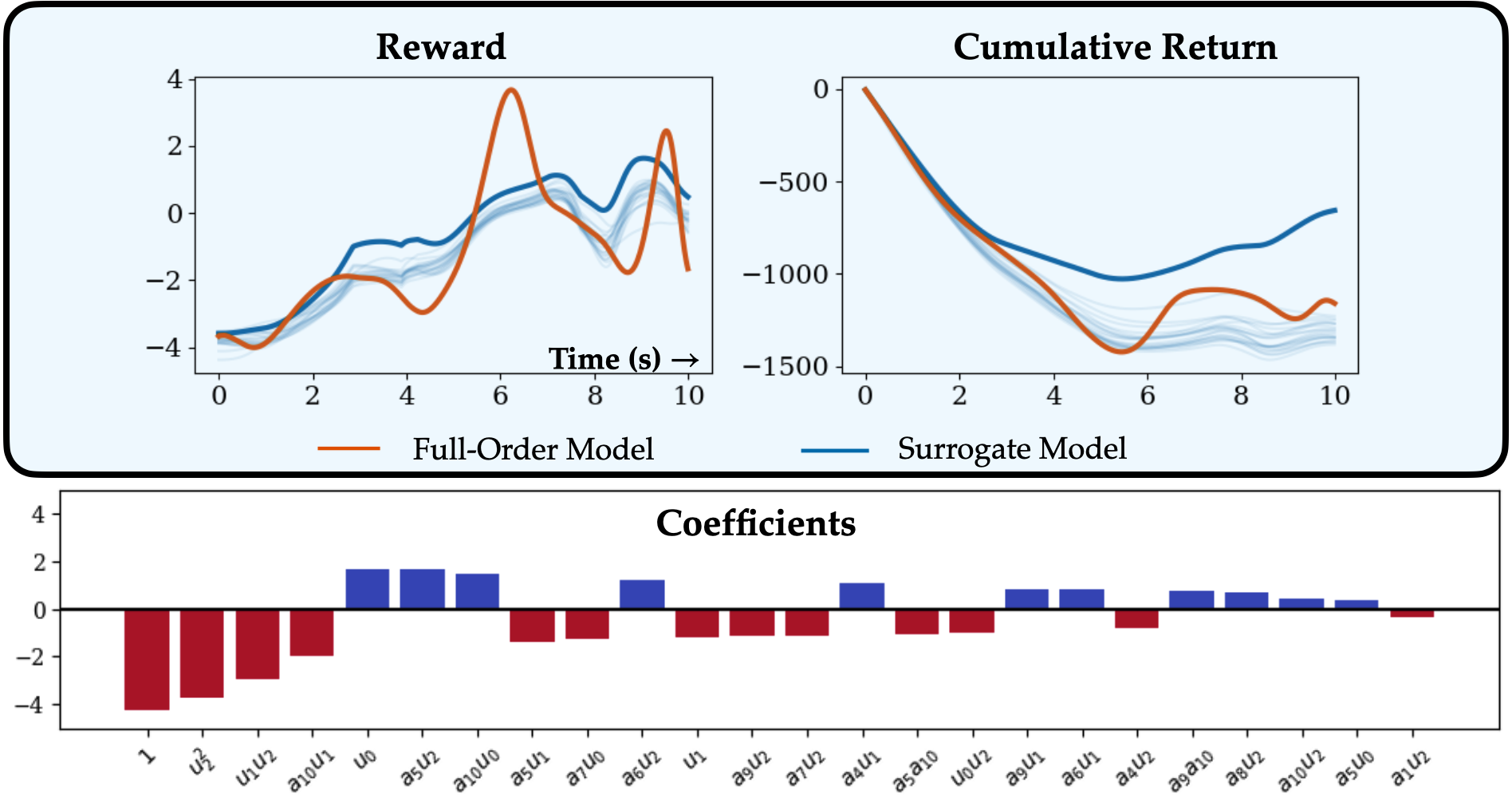}
    \cprotect \caption{\textbf{Pinball Reward.}
    \textit{Top:} Surrogate dictionary reward predictions evaluated on a trajectory produced from the full-order environment. The blue bold line indicates the predictions with the surrogate model with \textit{median} coefficients. The faint blue lines are individual predictions from members of the ensemble.
    \textit{Bottom:} A comparison of the learned, non-zero dictionary coefficients, $\vXi$, ordered by magnitude.
    }
     \label{fig:app-rew-pin}   
\end{figure*}

\clearpage

\subsection{3D Airfoil}
\label{appendix:rewards-airfoil}

The final airfoil reward was fit as a cubic polynomial of the projection onto the SVD modes $a_i$ and the control inputs $u_i$ (with no cross-terms forming the interaction). It's clear that the control terms dominate the learned reward. 

\begin{figure*}[h]
    \centering
    \includegraphics[width=0.95\textwidth]{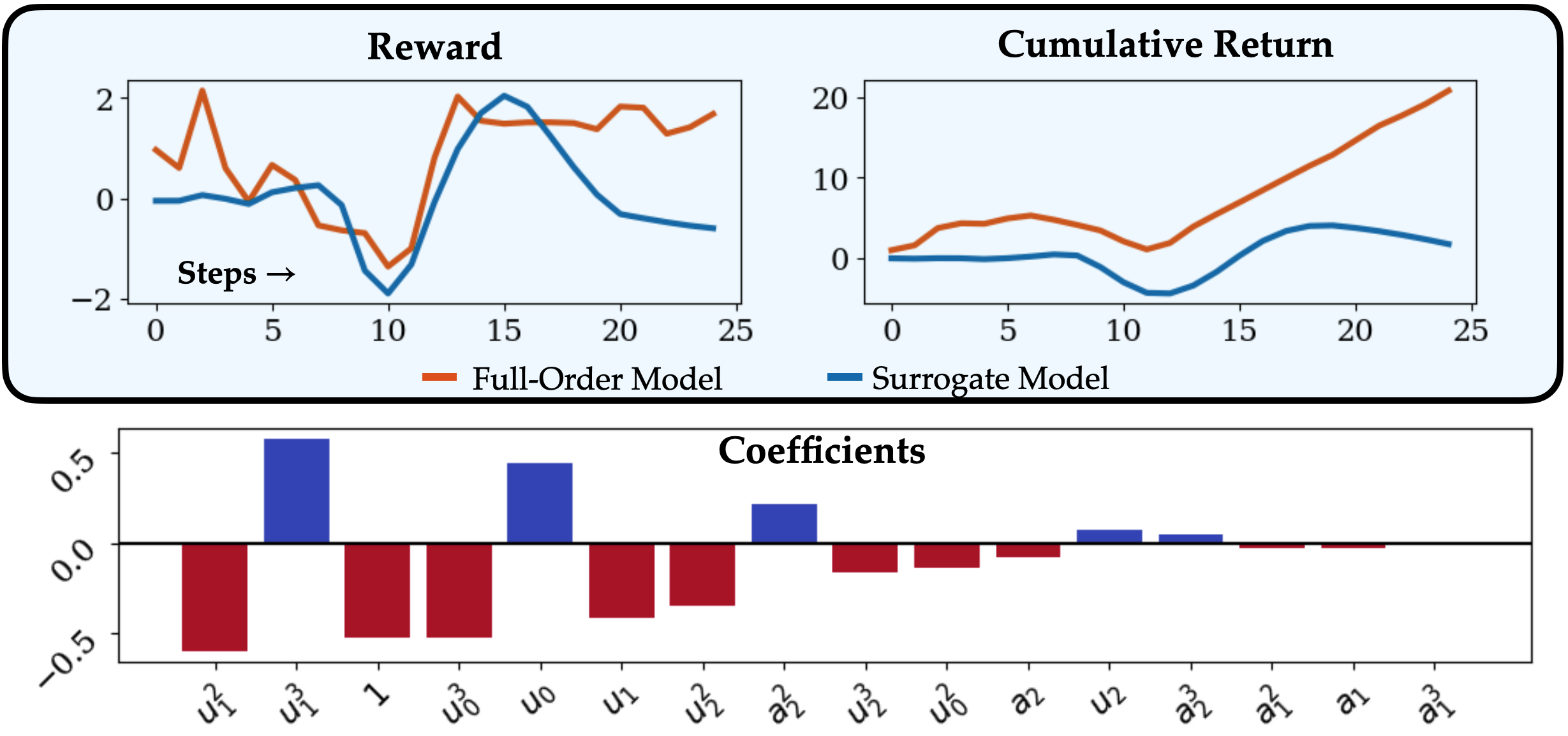}
    \cprotect \caption{\textbf{\airfoil Reward.}
    \textit{Top:} Surrogate dictionary reward predictions evaluated on a trajectory produced from the full-order environment. 
    \textit{Bottom:} A comparison of the learned, non-zero dictionary coefficients, $\vXi$, ordered by magnitude.
    }
     \label{fig:app-rew-naca}   
\end{figure*}

In Figure \ref{fig:app-rew-naca}, we evaluate the (median) E-SINDy reward function on data from the final evaluation episode. The reward matches closely for most of the episode, but underpredicts the final values. Despite this, the model was still able to learn a competitive policy to the DRL benchmark.

\clearpage

% ----------------------------------------------
\section{Dictionary Policy}
\label{appendix:policy}
We fit the surrogate policy by compiling input observations, $\vX$, evaluating them at points $\vY = \E{\pi_\policyParams(\vX)}$ and fitting an ensemble of cubic polynomial dictionary models $\hat{\pi}^{(k)}(\vx) = \vTheta^{(k)}(\vx) \vXi^{k}$ for $k = 1,2, \dots N_e = 20$. Below, in \TheAppendix \ref{appendix:policy-sampling} we discuss different sampling strategies for collecting the data $\vX$. For all of the results presented in this work, we acquire initial conditions and use our accompanying surrogate learned dynamics models with feedback control from the neural network model to bootstrap new cheap experience without needing to access the full-order model. Just as in training \appendixAlgoRef, we reset the surrogate environment if our trajectories begin to diverge. We apply a small amount of  noise to the trajectories by sampling twice from a Gaussian distribution before querying the neural network policy. Finally, we sweep over the STLRidge hyper-parameters on a test set. Table \ref{tab:appendix-policy} summarizes the parameters for each of the environments. Note that while rolling out the policy in the real and surrogate environments, we clip the controls to be between a minimal and maximal value. However, to learn the dictionary policies, it is sometimes more effective to loosen this bound when creating the data $\vY$ to allow for smoother predictions. In all of our experiments, we clip between -5 and 5. While one could use completely unclipped values, it can risk adding unnecessary complexity to the resulting polynomial. Although sparsity was encouraged during the STLRidge sweep, the policies did not end up being truly sparse, and we found the mean coefficients to provide a more robust prediction than the median.

\begin{table*}
    \centering
    \begin{tabular}{|c|c|c|c|c|c|}
    \hline
         Env&  Initial Conditions&  Traj. Length&  Noise& $N_{\text{dyn}}$\\
    \hline
         \swingup   
            & $\theta \sim \mathcal N(0,0.1), x=\dot x = \dot \theta \sim N(0, 0.25)$  
            &  500 % Traj length
            &  $0.1$ % Noise
            & 5,000 % N_dyn
            \\
        \hline
         \swimmer  
         &  $x_i \sim N(0,0.1)$
         &  200 % Traj Length
         &  $0.1$ % Noise
         & 10,000 % N_dyn
         \\
         \hline
         Cylinder   
         &  $x \sim \mathcal D_{\text{on}} + N(0, 0.1)$
         &  10 % Traj Len
         &  $0.1$
         & 10,000
         \\
         \hline
         Pinball
         &  $x \sim \mathcal D_{\text{on}} + N(0, 0.1)$
         & 100
         &  $0.1$
         & 10,000
         \\
        \hline
         \airfoil
         &  $x \sim \mathcal D_{\text{on}} + N(0, 0.2)$
         & 5
         &  $0.2$
         & 13,800
         \\
    \hline
    \end{tabular}
    \caption{Summary of the parameters uses to fit each surrogate policy. ``Initial condition'' refers to how the initial conditions were chosen. ``Traj. Length'' is the maximal length of the surrogate trajectories. ``Noise'' refers to the scale of the zero-mean added Gaussian noise after collecting the trajectories. $N_{\text{dyn}}$ refers to the total number of points collected using the surrogate dynamics. 
    }
    \label{tab:appendix-policy}
\end{table*}

\subsection{Policy Sampling Study}
\label{appendix:policy-sampling}
For the \swingup task, we consider six different ways of collecting the data for querying the neural network policy:
\begin{itemize}
    \item Sampling from an ambient mesh in $\mathbb{R}^m$ based off the collected data.
    \item Projecting the ambient space onto a manifold of with known constraints (i.e. the circle $S^1$)
    \item Sampling random trajectories from our learned dynamics and controller.
    \item Injecting noise onto sampled trajectories from the learned dynamics and controller.
    \item Sampling from previous full-order data examples.
    \item Injecting noise onto previous full-order data examples.
\end{itemize}
Once collected, we query the policy's expected value for each point in the state space: $\vu = \mathbb{E}[\pi_\policyParams(\vx)]$ to assemble data $(\vX, \vU)$ and split into the training and validation sets. For each sampling strategy, we use cubic polynomial dictionary and an ensemble of sparse dictionary models using STLRidge using bragging and library bagging. We sweep through the threshold and $L^2$ penalty hyper-parameters and choose the ensemble with minimal validation error. In Figure 9, we show the result of these different sampling strategies by evaluating the learned policy on $15$ random initial conditions from the \swingup task. By exploiting the surrogate dynamics and learned neural network to produce trajectories, we can greatly increase the desired performance without ever needing to utilize the full-order environment after training. Interestingly, by injecting a small amount of Gaussian noise onto the trajectory before querying the neural network, we do not degrade the performance at all.

\begin{figure*}[t]
    \centering
    \includegraphics[width=0.95\textwidth]{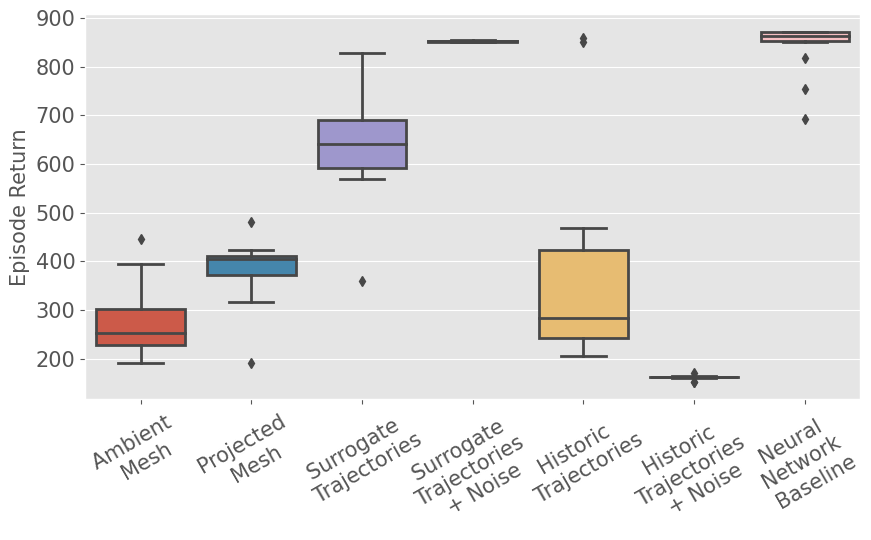}
    \cprotect \caption{Comparison of the six different data collection strategies and the performance of the resulting dictionary model deployed to the full-order environment.}
     \label{fig:policy-sample}   
\end{figure*}

\subsection{Swing-up}
\label{appendix:policy-swingup}
\begin{figure*}[t]
    \centering
    \includegraphics[width=0.5\linewidth]{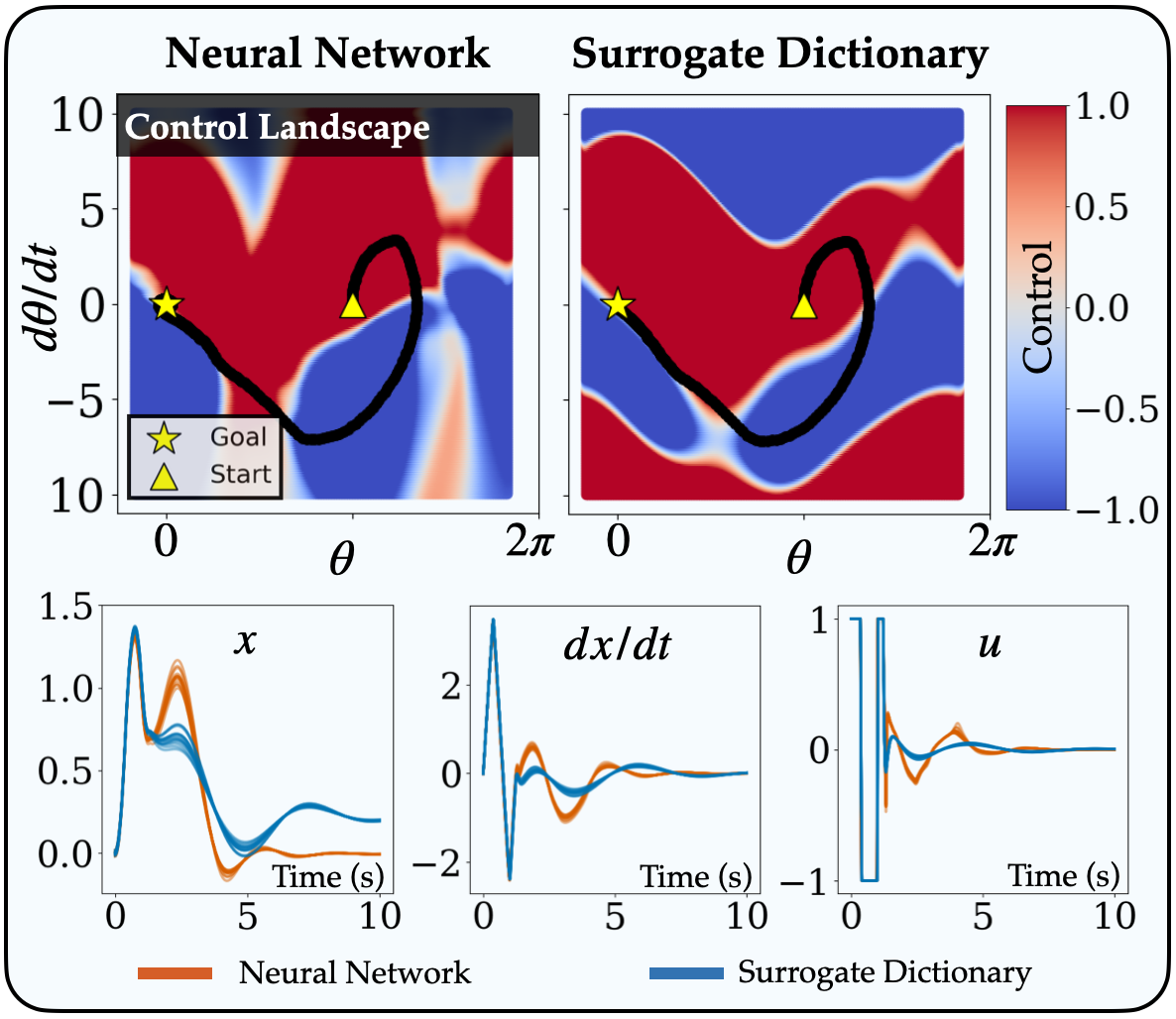}
    \cprotect \caption{%
        \textbf{Swing-up Policies.}
        Comparison of 15 trajectories from slightly different initial conditions in the full-order environment using the neural network policy (red) and surrogate dictionary approximation (blue).
        \textit{Top}: The control landscape of the two polices for different $(\theta, d\theta/dt)$ combinations
        (evaluated at $x=dx/dt = 0 $ for ease of visualization) and the corresponding trajectories under the influence of the policies. 
        \textit{Bottom}:
        The corresponding $x, dx/dt$, and $u$ for the 15 trajectories.
        }  
     \label{fig:policy-swingup}   
\end{figure*}

\begin{figure*}[h]
    \centering
    \includegraphics[width=0.95\textwidth]{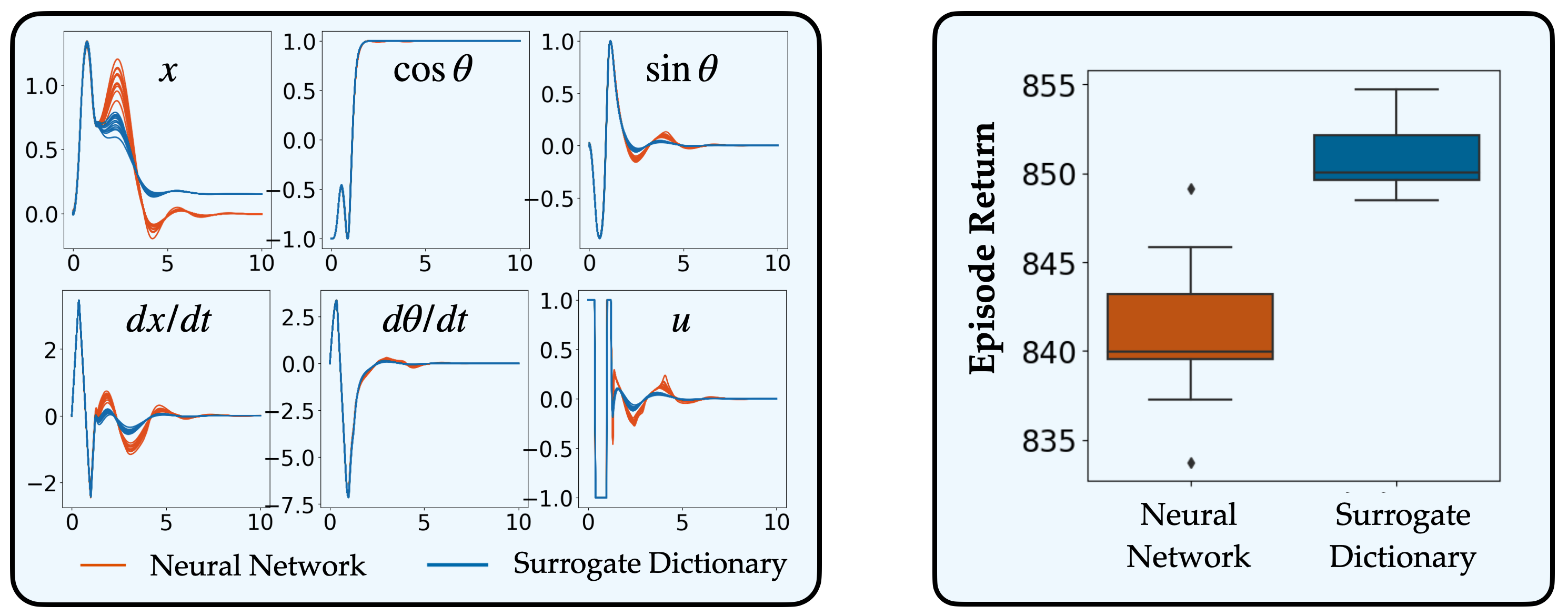}
    \cprotect \caption{\textbf{Swing-Up Policy: Trajectories.}
    The evaluation of 15 slightly different initial conditions using the surrogate dictionary (blue) and neural network policies discussed in the main text. \textit{Left:} Timeseries of the trajectories and control. 
    \textit{Right:} Distribution of the total cumulative returns from the 15 trajectories. 
    }
     \label{fig:app-policy-cart_rollout}   
\end{figure*}

\begin{figure*}[h]
    \centering
    \includegraphics[width=0.95\textwidth]{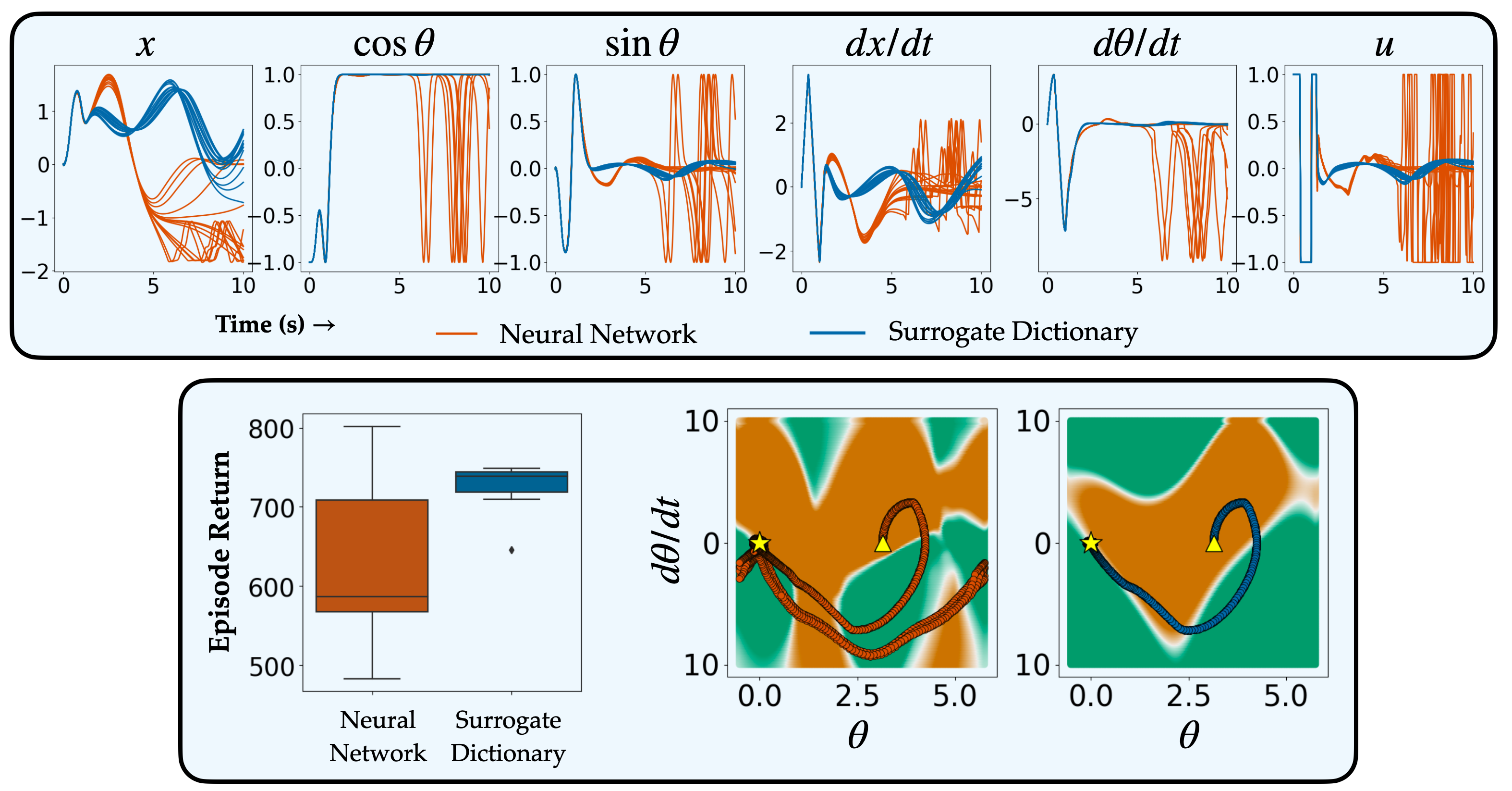}
    \cprotect \caption{\textbf{Swing-Up Policy: Neural Network Overfitting.}
    The evaluation of 15 slightly different initial conditions using the surrogate dictionary (blue) and neural network policies. \textit{Top:} Timeseries of the trajectories and control. \textit{Bottom Left:} Distribution of the total cumulative returns from the 15 trajectories.
    \textit{Bottom Right:} The control landscape evaluated on a mesh of $(\theta, \dot \theta)$ values. (For ease of visualization, $x=\dot x =0$). 
    
    }
     \label{fig:app-policy-cart-overfit}   
\end{figure*}

Figures \ref{fig:policy-swingup} and \ref{fig:app-policy-cart_rollout} demonstrate the effectiveness of the surrogate policy.
While the episode returns are larger for the surrogate dictionary, it's important to note that the performance is very comparable; with the median performance only being about 1\% different. In Figure \ref{fig:app-policy-cart-overfit}, we highlight a case where we continued to train the \swingup policy using \appendixAlgoRef beyond what was presented in Figures \ref{fig:policy-swingup} and \ref{fig:app-policy-cart_rollout}. At some point, the neural network starts to perform worse on the full-order environment, leading to an unstable control policy near the equilibrium. However, when using this neural network to create a surrogate dictionary, the resulting dictionary is a stable, consistent control policy. For both the neural network and the policy, the returns are significantly diminished compared to what's shown in Figure \ref{fig:app-policy-cart_rollout}, and this time the surrogate dictionary's median performance is about 20\% better.

\subsection{Swimmer-v4}
\label{appendix:policy-swimmer}
\begin{figure*}[h]
    \centering
    \includegraphics[width=0.95\textwidth]{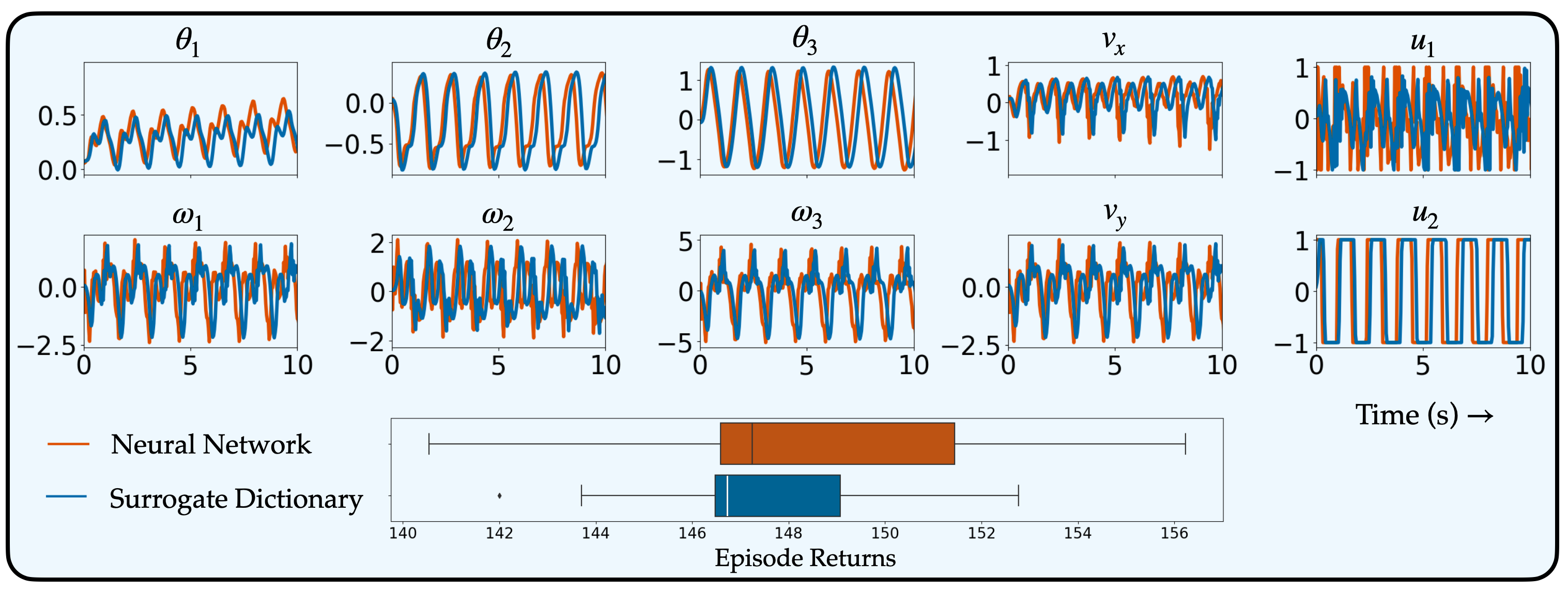}
    \cprotect \caption{\textbf{Swimmer-v4 Policy Trajectories.}
    Example of one of 15 slightly different initial conditions using the surrogate dictionary (blue) and neural network policies. 
    \textit{Bottom:} Distribution of the total cumulative returns from the 15 trajectories. 
    }
     \label{fig:app-policy-swim-rollout}   
\end{figure*}

As with the \swingup environment, we get a comparable surrogate policy using a cubic dictionary compared to the original neural network, with only a slight difference in median performance. The results can be seen in Figure \ref{fig:app-policy-swim-rollout}.

\subsection{Cylinder}
\label{appendix:policy-cylinder}
\begin{figure*}[t]
    \centering
    \includegraphics[width=0.99\textwidth]{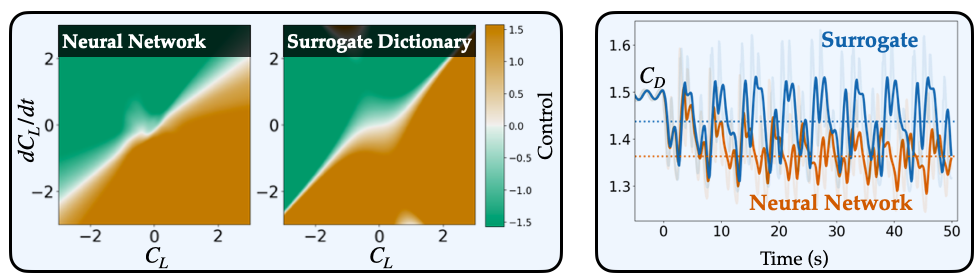}
    \cprotect \caption{%
        \textbf{Cylinder Policies.}
        \textit{Left}:
        The control landscapes for the neural network and surrogate policies
        \textit{Right}:
        The corresponding full-order $C_D$ for the neural network and surrogate policy evaluation, with control starting at $t=0$. A 1-second window moving average is plotted in bold. Dotted lines indicate the average $C_D$ over the last 30 seconds. 
        }  
     \label{fig:policy-cylinder}   
\end{figure*}

Unlike the other examples, the neural network Cylinder policy is much more difficult to approximate with a dictionary. As shown in Figure \ref{fig:policy-cylinder}, when clipping the policies between the maximal control range ($\pm \pi/2$), the neural network policy has essentially learned a ``bang-bang'' controller. The very thin, nonlinear region of zero-control acts as a discontinuity in the control landscape, which is extremely challenging to approximate with a low-dimensional polynomial. The surrogate dictionary comes close, but has too large of a region where the policy provides small control, leading to the degraded performance. 

\begin{figure*}[t]
    \centering
    \includegraphics[width=0.95\textwidth]{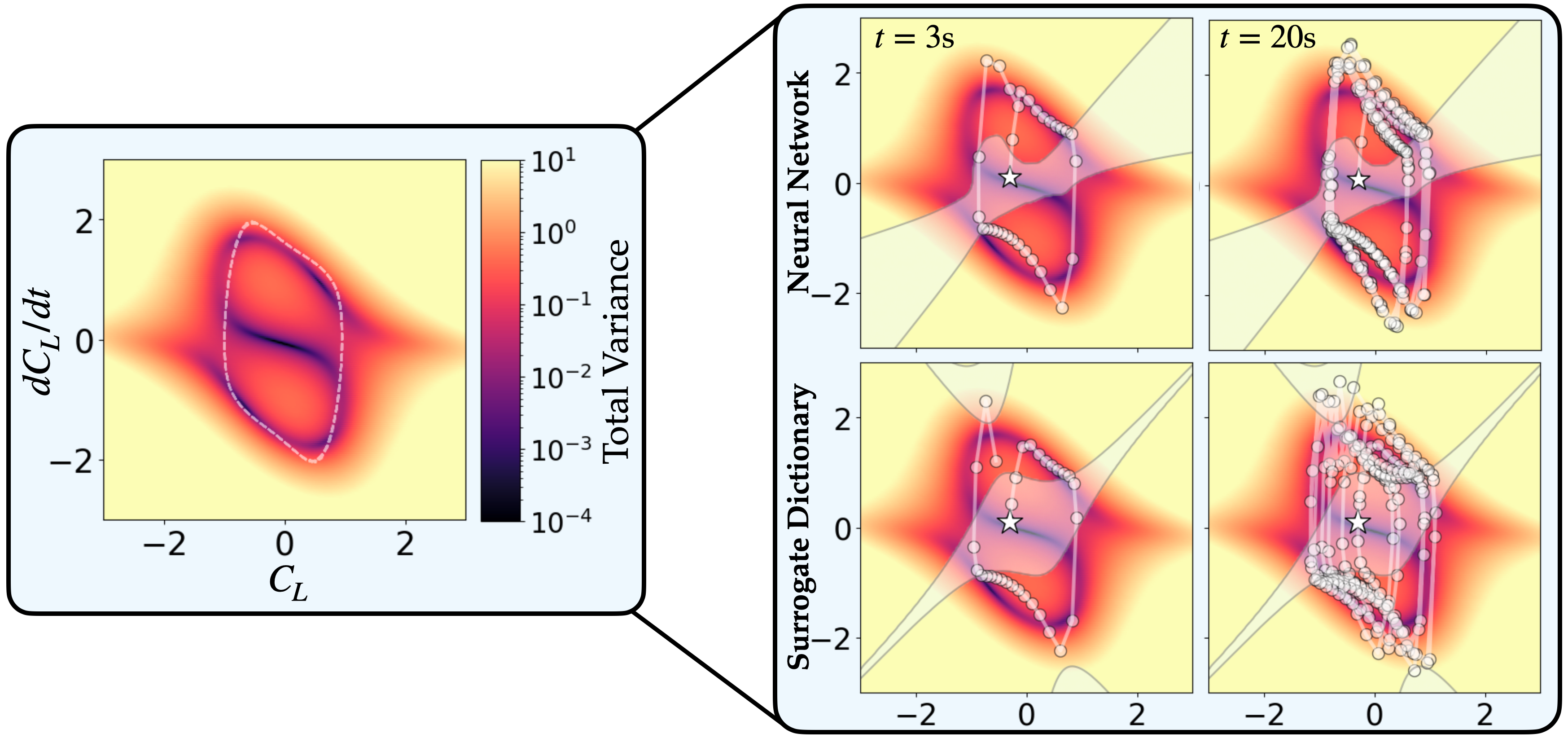}
    \cprotect \caption{\textbf{Cylinder Policy Variance.}
    \textit{Left:} The variance landscape with the learned limit cycle  induced by feedback control in the surrogate environment (dashed white line). \textit{Right:} Snapshots of trajectories from the full order environment produce from the neural network (top) and surrogate policies (bottom) plotted on top of the variance landscape. The star in the center indicate the initial condition, and regions where $|u| < u_{\text{max}} = \pi/2$ (i.e. the light-colored regions from Figure \ref{fig:policy-cylinder}) are plotted in white for policies. 
    }
     \label{fig:cyl-policy-uq}   
\end{figure*}

In Figure \ref{fig:cyl-policy-uq}, we inspect the learned policy for the Cylinder agents. There is a low-variance, ``s''-shaped basin which is well-aligned with the limit cycle arising from the neural network policy. The data used to fit the surrogate policy was primarily sampled from within the convex hull of the limit cycle; however, when we examine trajectories with feedback control from both policies in the full-order environment, we discover that trajectories leave this limit cycle in both cases. In particular, the dictionary surrogate reaches regions of large variance; at about $t=3$, it reaches an uncertain region that receives small control input at the top of the basin, bringing it back into a relatively large region of uncertainty on the interior of the limit cycle. The surrogate policy then struggles to reach the limit cycle as it is stuck in large regions of uncertainty. This uncertainty-based analysis hints that additional sampling strategies could reduce the variance of the learned surrogate policy and better imitate the neural network policy.

\subsection{Pinball}
\label{appendix:policy-pinball}
\begin{figure*}[t]
    \centering
    \includegraphics[width=0.95\textwidth]{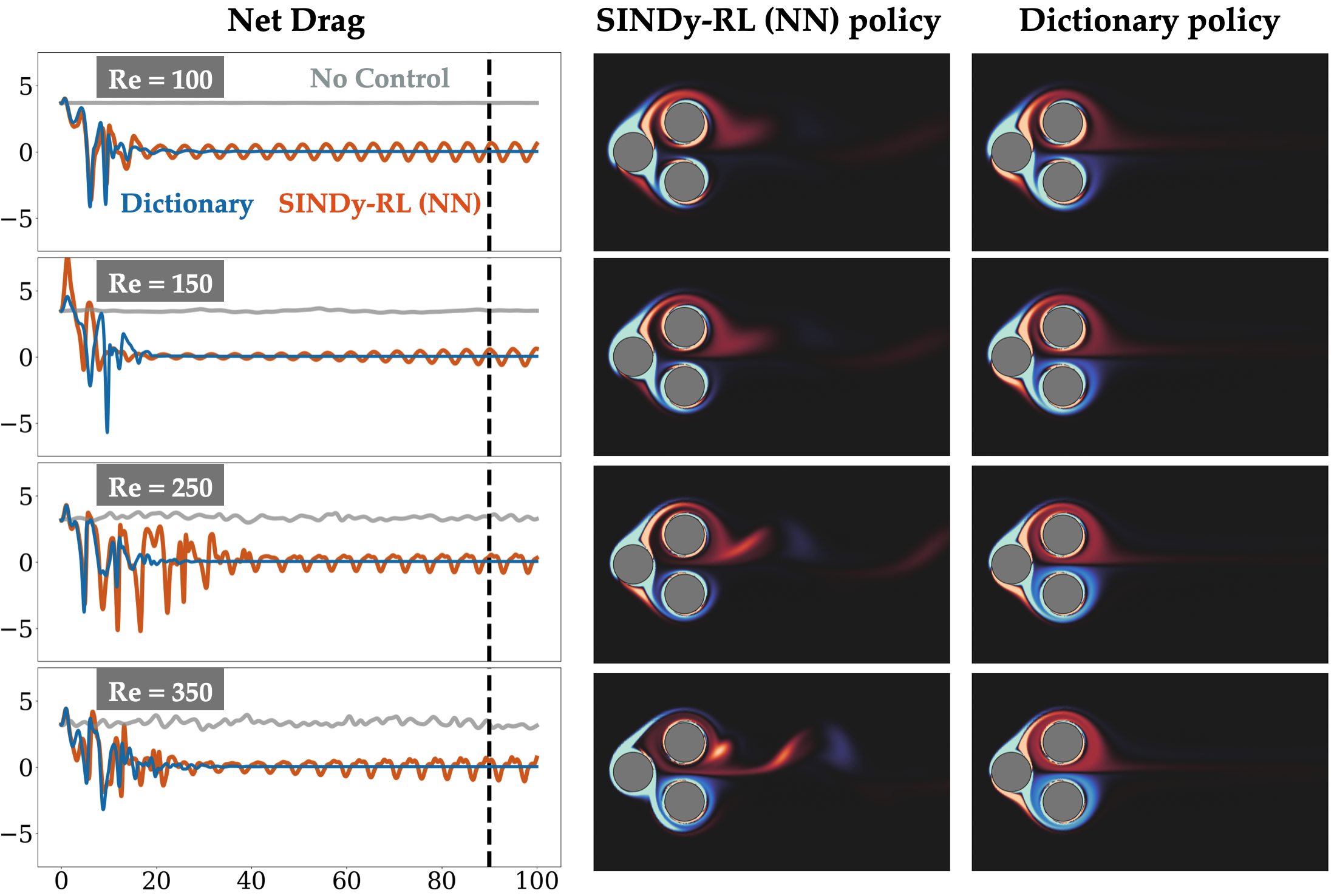}
    \cprotect \caption{\textbf{Pinball policy $Re$ evaluation}. 
    Evaluation of the best performing SINDy-RL NN and corresponding dictionary policy across $Re$. \textit{Left:} Net drag experienced on the system for the dictionary agents (blue), neural network agents (red), and without control (gray). Negative values indicate thrust generation. Dashed line at $t=90$ corresponds to the snapshots on the right. \textit{Right:} Vorticity snapshots at $t=90$.}
     \label{fig:app-policy-pin}   
\end{figure*}

In Figure \ref{fig:app-policy-pin} we compare the effect of the learned neural network policy obtained from SINDy-RL and the corresponding dictionary policy. Both policies experience an initial transient before settling into stable configuration. Whereas the neural network policy produces small oscillations around zero and small amounts of vortex shedding for large $Re$, the dictionary policy converges to a fixed drag and stabilizes the wake across all values of $Re$ that we evaluated. In addition to \swingup results presented in Figure \ref{fig:app-policy-cart-overfit}, this provides further evidence that dictionary policy might improve the generalization beyond the dynamics it was trained on.

\clearpage

\subsection{3D Airfoil}
\label{appendix:policy-airfoil}

To approximate the neural network policy obtained by SINDy-RL, a quadratic dictionary was fit: 

\begin{align}
u_0 &= 0.501 -0.022 a_1 -0.108 a_2 -0.003 a_1^2 + 0.002 a_1 a_2  -0.012 a_2^2  \\
u_1 &= 0.011 -0.076 a_1 + 0.267 a_2 -0.001 a_1^2 + 0.001 a_1 a_2 + 0.021 a_2^2  \\ 
u_2 &= 0.069 -0.010 a_1 + 0.170 a_2   \qquad \qquad \;\; + 0.003 a_1 a_2 + 0.009 a_2^2 \\
\end{align}

\begin{figure*}[t]
    \centering
    \includegraphics[width=0.95\textwidth]{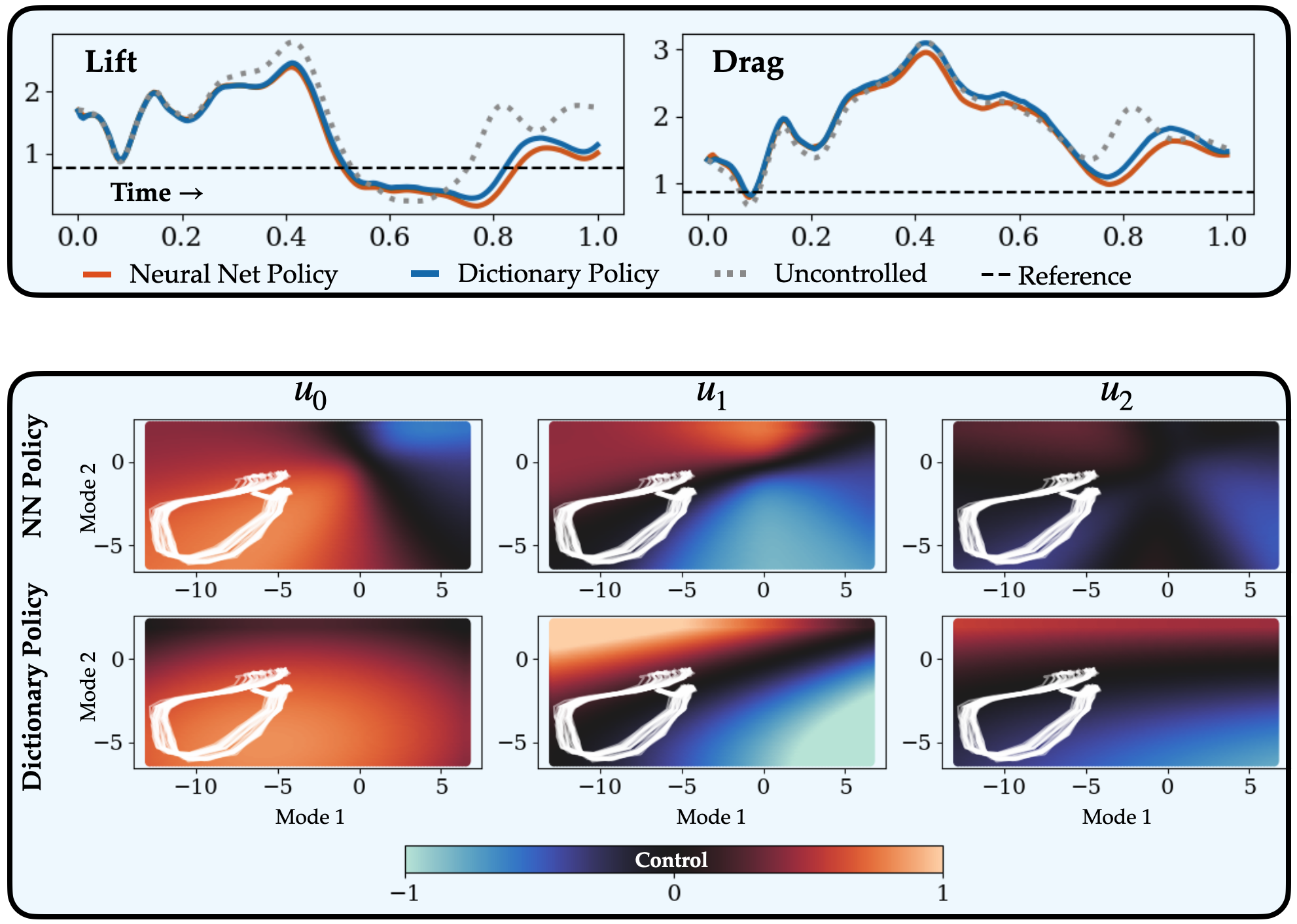}
    \cprotect \caption{\textbf{\airfoil policy}. 
    \textit{Top:} Evaluation of the final SINDy-RL NN and corresponding dictionary policy on the lift and drag forces. The gray dotted line corresponds to the uncontrolled case and the black dashed line corresponds to the desired reference value. 
    \textit{Bottom:} Control landscapes for the neural network and dictionary policies. Trajectories used for training the policy are plotted in white. 
    }
     \label{fig:app-policy-naca}   
\end{figure*}

Despite its simple form, the mean E-SINDy policy acts as a great approximation of its corresponding neural network policy. The resulting dictionary policy achieved a total return of 24.88, compared to the 25.81 from the neural network. Despite the slightly degraded performance, Figure \ref{fig:app-policy-naca} demonstrates the comparable effect of the policy on the lift and drag forces on the airfoil. It can also be qualitatively seen that the dictionary policy finds a slightly smoother control inputs around the on-policy training data.

        % \putbib[refs]
    % \end{bibunit}
\end{appendices}

\end{bibunit}
\end{document}